\documentclass{article}
\usepackage[main,final]{neurips_2026}

\usepackage{amsmath,amsfonts,bm}

\def\eqref#1{equation~\ref{#1}}

\def\1{\bm{1}}

\DeclareMathAlphabet{\mathsfit}{\encodingdefault}{\sfdefault}{m}{sl}
\SetMathAlphabet{\mathsfit}{bold}{\encodingdefault}{\sfdefault}{bx}{n}

\usepackage[utf8]{inputenc} 
\usepackage[T1]{fontenc}    
\usepackage{hyperref}       
\usepackage{url}            
\usepackage{booktabs}       
\usepackage{amsfonts}       
\usepackage{nicefrac}       
\usepackage{microtype}      
\usepackage{xcolor}         
\usepackage{makecell}    
\usepackage{amssymb}     
\usepackage{graphicx} 
\usepackage{caption}
\usepackage{wrapfig}
\usepackage{tabularx}
\usepackage{array}     
\usepackage{listings}
\usepackage{xcolor}
\usepackage{tcolorbox}
\usepackage{amsmath}
\usepackage{pifont}
\usepackage{svg}
\usepackage{algorithm}
\usepackage{algpseudocode}
\usepackage{multirow}   
\usepackage{textcomp} 
\UseRawInputEncoding

\definecolor{mygreen}{RGB}{34,139,34}
\definecolor{myred}{RGB}{220,20,60}
\definecolor{toolbg}{RGB}{248,249,250}
\definecolor{toolframe}{RGB}{220,223,228}
\definecolor{toolkey}{RGB}{0,92,175}
\definecolor{toolstring}{RGB}{163,21,21}
\definecolor{toolcomment}{RGB}{0,128,0}
\definecolor{toolnum}{RGB}{128,0,128}
\definecolor{toolrule}{RGB}{150,150,150}

\lstdefinelanguage{json}{
    basicstyle=\ttfamily\small,
    showstringspaces=false,
    breaklines=true,
    frame=single,
    backgroundcolor=\color{toolbg},
    rulecolor=\color{toolframe},
    frameround=tttt,
    numbers=left,
    numberstyle=\scriptsize\color{toolrule},
    stepnumber=1,
    numbersep=8pt,
    tabsize=2,
    columns=fullflexible,
    keepspaces=true,
    upquote=true,
    stringstyle=\color{toolstring},
    commentstyle=\color{toolcomment},
    literate=
     *{0}{{{\color{toolnum}0}}}{1}
      {1}{{{\color{toolnum}1}}}{1}
      {2}{{{\color{toolnum}2}}}{1}
      {3}{{{\color{toolnum}3}}}{1}
      {4}{{{\color{toolnum}4}}}{1}
      {5}{{{\color{toolnum}5}}}{1}
      {6}{{{\color{toolnum}6}}}{1}
      {7}{{{\color{toolnum}7}}}{1}
      {8}{{{\color{toolnum}8}}}{1}
      {9}{{{\color{toolnum}9}}}{1}
      {:}{{{\color{black}{:}}}}{1}
      {,}{{{\color{black}{,}}}}{1}
      {\{}{{{\color{black}{\{}}}}{1}
      {\}}{{{\color{black}{\}}}}}{1}
      {[}{{{\color{black}{[}}}}{1}
      {]}{{{\color{black}{]}}}}{1},
}

\lstdefinestyle{toolstyle}{
    language=json,
    basicstyle=\ttfamily\small,
    backgroundcolor=\color{toolbg},
    frame=single,
    rulecolor=\color{toolframe},
    framesep=3pt,
    xleftmargin=1em,
    xrightmargin=0.5em,
    breaklines=true,
    breakatwhitespace=false,
    showstringspaces=false,
    keepspaces=true,
    columns=fullflexible,
    numbers=left,
    numberstyle=\scriptsize\color{toolrule},
    numbersep=8pt,
    tabsize=2,
    captionpos=b,
    numbers=none
}

\title{EgoBench: An Interactive Egocentric Multimodal Benchmark for Tool-Using Agents}

\author{
  Yunqi Liu\thanks{Equal contribution.}\\
  Ant Group\\
  \texttt{liuyunqi.lyq@antgroup.com} \\
  \And
  Tong Niu$^*$ \\
  Ant Group \\
  \texttt{niutong.niu@antgroup.com} \\
  \And
  Zitong Wang \\
  Ant Group \\
  \texttt{yesi.wzt@antgroup.com} \\
  \And
  Zhenlong Dai \\
  Ant Group \\
  \texttt{daizhenlong.dzl@antgroup.com} \\
  \And
  Yuqi Qing \\
  Ant Group \\
  \texttt{qingyuqi.qyq@antgroup.com} \\
  \And
  Weiqiang Wang \\
  Ant Group \\
  \texttt{weiqiang.wwq@antgroup.com} \\
  \And
  Jian Liu\thanks{Corresponding author.} \\
  Ant Group \\
  \texttt{rex.lj@antgroup.com} \\
}

\begin{document}

\maketitle

\begin{abstract}
As AI agents increasingly operate in open, real-world environments, they require a deep synergy of multimodal perception, tool invocation with multi-hop reasoning, and dynamic interaction with users. However, existing benchmarks fail to jointly evaluate these capabilities due to challenges in designing strictly coupled multi-capability tasks, simulating natural and task-constrained user feedback, and ensuring objective evaluation of dynamic interaction. To bridge this gap, we introduce EgoBench, the first interactive multimodal benchmark for tool-using agents. EgoBench comprises 1,590 egocentric-video-grounded tasks covering five daily scenarios, along with a user-agent-tool interactive environment for evaluation. We implement a three-stage synergistic pipeline through which each task is designed to enforce the joint application of visual perception and tool-augmented multi-hop reasoning. We additionally develop a multi-agent simulated user to evaluate agents' interaction capabilities, which generates high-fidelity, task-aligned responses to agents. Furthermore, we establish a deterministic joint validation framework that guarantees objective assessment through process-based and result-based equivalence. Benchmarking eight video-MLLM agents on EgoBench reveals a severe performance ceiling: the best-performing model achieves only a 34.95\% average Joint Success Rate across the three interaction modes. Finally, we conduct a multi-dimensional error analysis to disentangle failure modes, exposing capability bottlenecks for advancing future AI agents.\footnote{Our code and dataset are available at \url{https://github.com/Wind-Flipped/EgoBench}.}

\end{abstract}

\section{Introduction}
\label{Introduction}

Recent advances in AI agents have enabled effective automation of structured workflows such as desktop task assistance, and code generation (~\cite{llmagenttool,AgentBench,codeagentsurvey,AgencyBench}). Most existing agents rely on unimodal inputs and operate within predefined task boundaries. As agents are increasingly deployed on ubiquitous devices such as wearables and integrated into daily life, they must operate in open, real-world environments grounded in users' situated context and perception. This shift requires agents to possess three key capabilities for effectively understanding user intent and coordinating tool use in complex, user-initiated tasks: (1) multimodal contextual understanding to resolve user references, (2) multi-hop reasoning and execution to process information with complex tool-based actions, and (3) interaction to adapt to users' dynamic goals and proactively obtain information (~\cite{tau2-bench,sun2025multimodal}).

However, existing benchmarks for evaluating models fail to comprehensively assess these three capabilities. For example, ~\cite{tau-bench,tau2-bench,ToolSandbox} lack the multimodal understanding assessment; ~\cite{VisualWebArena,MMs,MLLM-Tool,GTA} lack support for dynamic interaction with users; and ~\cite{MathVista,MMMU} overlook the integration of tool use and action execution. Meanwhile, constructing a benchmark that jointly assesses all three capabilities remains challenging for three main reasons: (1) Ensuring capability coupling: it is difficult to design tasks that genuinely require the joint use of multimodal understanding, logic reasoning, and tool invocation, as models often exploit shortcuts or rely on a single capability rather than integrating all (\cite{M3-bench,Beyond-Seeing}). (2) Simulating realistic user-agent interactions: It is challenging to generate user feedback that is both natural and aligned with task goals and constraints, given the unpredictability of the evaluated agent's behavior (\cite{UserLM-R1}). (3) Establishing reliable and objective evaluation criteria: Extracting quantifiable metrics from dynamic and subjective interactions is challenging yet indispensable, as it serves as the cornerstone of stable, objective, and reproducible assessment (\cite{Beyond-Single-Turn-survey}).


\begin{table}[t]
\centering
\caption{Comparisons to tool-using benchmarks. \textit{Multi-modal}: Presence of multimodal information. 
\textit{Dynamic Inter.}: Involvement of dynamic user interaction.
\textit{Multi-hop Reason.}: Requirement for multi-step reasoning.
\textit{Objective Eval.}: Use of objective metrics instead of LLM judges. 
\textit{Exec. Result Eval.}: Evaluation on database state.
\textit{State-dep. Tools}: Stateful tool dependency (\cite{ToolSandbox}).}
\label{tab:benchmark_comparison}
\scriptsize
\setlength{\tabcolsep}{2.5pt}
\renewcommand{\arraystretch}{1.0}
\begin{tabular*}{\columnwidth}{@{\extracolsep{\fill}}lcccccc@{}}
\toprule
\textbf{Benchmark} & \textbf{\makecell[c]{Multi-\\modal}} & \textbf{\makecell[c]{Dynamic\\Inter.}} & \textbf{\makecell[c]{Multi-hop\\Reason.}} & \textbf{\makecell[c]{Objective\\Eval.}} & \textbf{\makecell[c]{Exec. Result\\Eval.}} & \textbf{\makecell[c]{State-dep.\\Tools}} \\
\midrule
BFCL (\cite{BFCL}) & \textcolor{myred}{\ding{55}} & \textcolor{myred}{\ding{55}} & \textcolor{mygreen}{\checkmark} & \textcolor{mygreen}{\checkmark} & \textcolor{myred}{\ding{55}} & \textcolor{mygreen}{\checkmark} \\
ToolBench (\cite{ToolLLM}) & \textcolor{myred}{\ding{55}} & \textcolor{myred}{\ding{55}} & \textcolor{mygreen}{\checkmark} & \textcolor{myred}{\ding{55}} & \textcolor{myred}{\ding{55}} & \textcolor{myred}{\ding{55}} \\
M$^3$-bench (\cite{M3-bench}) & \textcolor{mygreen}{\checkmark} & \textcolor{myred}{\ding{55}} & \textcolor{mygreen}{\checkmark} & \textcolor{myred}{\ding{55}} & \textcolor{myred}{\ding{55}} & \textcolor{myred}{\ding{55}} \\
tau$^2$-bench (\cite{tau2-bench}) & \textcolor{myred}{\ding{55}} & \textcolor{mygreen}{\checkmark} & \textcolor{mygreen}{\checkmark} & \textcolor{mygreen}{\checkmark} & \textcolor{mygreen}{\checkmark} & \textcolor{mygreen}{\checkmark} \\
ToolSandbox (\cite{ToolSandbox}) & \textcolor{myred}{\ding{55}} & \textcolor{mygreen}{\checkmark} & \textcolor{mygreen}{\checkmark} & \textcolor{mygreen}{\checkmark} & \textcolor{mygreen}{\checkmark} & \textcolor{mygreen}{\checkmark} \\
ComplexFuncBench (\cite{ComplexFuncBench}) & \textcolor{myred}{\ding{55}} & \textcolor{myred}{\ding{55}} & \textcolor{mygreen}{\checkmark} & \textcolor{myred}{\ding{55}} & \textcolor{mygreen}{\checkmark} & \textcolor{myred}{\ding{55}} \\
ACEBench (\cite{ACEBench}) & \textcolor{myred}{\ding{55}} & \textcolor{mygreen}{\checkmark} & \textcolor{myred}{\ding{55}} & \textcolor{mygreen}{\checkmark} & \textcolor{myred}{\ding{55}} & \textcolor{myred}{\ding{55}} \\
VisualToolBench (\cite{Beyond-Seeing}) & \textcolor{mygreen}{\checkmark} & \textcolor{myred}{\ding{55}} & \textcolor{myred}{\ding{55}} & \textcolor{myred}{\ding{55}} & \textcolor{myred}{\ding{55}} & \textcolor{myred}{\ding{55}} \\
GTA (\cite{GTA}) & \textcolor{mygreen}{\checkmark} & \textcolor{myred}{\ding{55}} & \textcolor{mygreen}{\checkmark} & \textcolor{mygreen}{\checkmark} & \textcolor{myred}{\ding{55}} & \textcolor{myred}{\ding{55}} \\
\textbf{EgoBench (Ours)} & \textcolor{mygreen}{\checkmark} & \textcolor{mygreen}{\checkmark} & \textcolor{mygreen}{\checkmark} & \textcolor{mygreen}{\checkmark} & \textcolor{mygreen}{\checkmark} & \textcolor{mygreen}{\checkmark} \\
\bottomrule
\end{tabular*}
\end{table}



\begin{figure}[htbp]
    \centering
    \includegraphics[width=\textwidth]{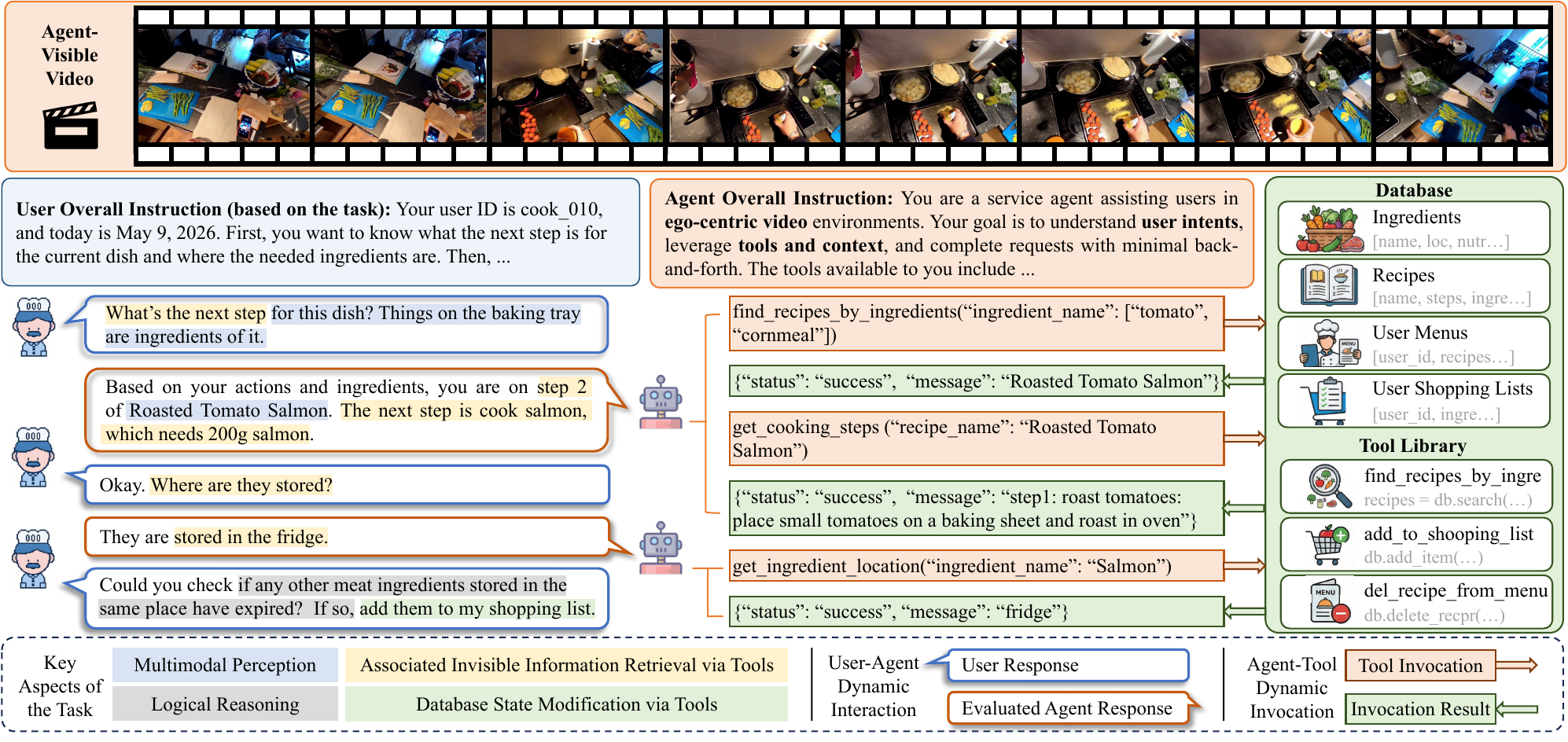}
    \caption{An illustrative interaction process for a sample EgoBench task. It demonstrates the dynamic interaction between the evaluated agent and the simulated user, including tool invocations. Different background colors in the interaction content highlight distinct aspects of the task design, illustrating how our benchmark comprehensively evaluates multimodal perception, multi-step reasoning, tool use, and interactive capabilities.}
    \label{fig:first_figure}
\end{figure}

To bridge the aforementioned evaluation gap, we introduce \textbf{EgoBench}, the first benchmark for tool-using agents grounded in egocentric video, which jointly evaluates multimodal perception, dynamic interaction, and multi-hop reasoning (Table~\ref{tab:benchmark_comparison}). To address the above challenges, EgoBench is built around the following designs: 
First, we implement a \textbf{three-stage synergistic task generation pipeline} to generate evaluation tasks and tool-using environment that assess models' integrated abilities of multimodal perception, tool use, and reasoning. Specifically, in Stage 1, we collect egocentric videos rich in explicit spatiotemporal cues to serve as visual anchors. In stage 2, we construct a comprehensive tool library and a database integrating both visible video content and invisible but contextually relevant supplementary information, introducing a visual-data information gap. In Stage 3, we design tasks that require the evaluated agent to complete four different aspects to succeed: multimodal perception, retrieval of associated invisible information via tools, logical reasoning, and database state update via tools. In total, Egobench comprises 1,590 tasks spanning five real-world scenarios: Restaurant, Kitchen, Retail, Warehouse and Household. Second, we develop a \textbf{multi-agent simulated user} with an Actor-Evaluator-Summarizer structure that creates a robust user-agent interaction environment by iteratively generating and refining user responses via four binary evaluation criteria. It further supports three interaction modes (Dynamic Easy, Dynamic Hard, and Static) to evaluate models under varying interactive complexities. Fig.~\ref{fig:first_figure} shows an example task and interaction. Third, we propose a \textbf{joint validation framework} that quantifies agent performance by simultaneously assessing process-based tool-call coverage and result-based database state equivalence. Unlike subjective LLM-based evaluations, this deterministic approach ensures objective precision, leveraging our task design which inherently requires tool invocations and state modifications. Furthermore, we benchmark eight video-MLLMs on EgoBench, and the top-performing model achieves only 36.21\%, 27.03\%, and 41.60\% Joint Success Rate in Dynamic Easy, Dynamic Hard, and Static modes, respectively, averaged over five scenarios using the benchmark task-count weights. To diagnose these failures, we implement a \textbf{multi-dimensional error analysis} that disentangles failures into five categories. Our main contributions are summarized as follows:

\begin{itemize}
    
    \item We develop EgoBench, the first benchmark to jointly evaluate tool-using agents on multimodal perception, multi-hop reasoning, and dynamic user interaction, which comprises 1,590 tasks across five daily scenarios within a dynamic environment.

    \item We implement a three-stage synergistic task generation pipeline to ensure that every generated task jointly evaluates multimodal perception and tool-augmented logical reasoning. Concurrently, we design an Actor-Evaluator-Summarizer simulated user to generate high-fidelity, goal-aligned responses to unpredictable agent behaviors, thereby providing a robust environment for evaluating interactive capabilities.

    \item We establish a deterministic joint validation framework in EgoBench that enables objective and precise assessment through process-and-result equivalence, benchmark eight video-MLLMs, and conduct a multi-dimensional error analysis to diagnose failure modes and inform future model development.

\end{itemize}




\section{Related Work}

\subsection{Tool-Using Benchmarks}
Early research on tool use primarily focuses on evaluating API comprehension and parameter generation. Gorilla (~\cite{Gorilla}) shows that domain-specific fine-tuning can outperform larger closed-source models, while ToolBench (~\cite{ToolLLM}) and API-Bank (~\cite{API-Bank}) further scale up instruction tuning and introduce multi-stage evaluation frameworks. Subsequent benchmarks such as BFCL (~\cite{BFCL}) standardize function calling across diverse programming environments, while AgentBench (~\cite{AgentBench}) advances an agent-centric paradigm by situating LLMs in terminal and database environments. More recent efforts move toward standardized agentic protocols and extended context windows: MCP-Bench (~\cite{MCP-Bench}) and LiveMCP-101 (~\cite{LiveMCP-101}) evaluate agents under the Model Context Protocol, while AgencyBench (~\cite{AgencyBench}) studies long-horizon planning under million-token context windows. Despite this progress, these benchmarks still predominantly operate in text-only, static, or highly structured environments.

\subsection{User Simulation for Conversational Agents}
LLM-based user simulators largely replace rigid goal templates, enabling more natural and context-aware dialogue evaluation (~\cite{user_simulation_1, user_simulation_2}). This paradigm extends to tool-augmented settings, where simulated users actively interact with external APIs, as exemplified by $\tau$-bench (~\cite{tau-bench}) and its dual-control extension $\tau^2$-Bench (~\cite{tau2-bench}). To enhance behavioral fidelity, recent work incorporates long-term personalization (~\cite{PersonaLens}), stateful execution environments (~\cite{ToolSandbox}), and real-time turn-taking dynamics (~\cite{Full-Duplex-Bench}). While these simulators significantly improve conversational realism, they remain predominantly text-centric and operate under predefined interaction flows. Consequently, they struggle to capture the multimodal grounded, and dynamically evolving intent shifts that characterize real-world human-agent collaboration.

\subsection{Multimodal Benchmarks}
Multimodal evaluation now extends beyond passive perception to active, tool-mediated problem solving. VisualWebArena (~\cite{VisualWebArena}) pioneers agent navigation in visually complex web interfaces, while m\&m's (~\cite{MMs}) introduces multi-step computational planning that requires coordinated tool orchestration across modalities. To address intent ambiguity, MLLM-Tool (~\cite{MLLM-Tool}) and GTA (~\cite{GTA}) evaluate context-aware tool selection with deployed APIs and diverse visual inputs, such as screenshots and handwriting. M$^3$-Bench (~\cite{M3-bench}) further standardizes multi-hop, multi-threaded evaluation via the MCP, and Beyond Seeing (~\cite{Beyond-Seeing}) highlights the necessity of active visual preprocessing under imperfect inputs. Although these benchmarks advance multimodal reasoning, they typically rely on static instructions and isolated evaluations, neglecting the user-driven feedback loops and process-level validation.


\section{EgoBench}
\label{EgoBench}

\begin{figure}[htbp]
    \centering
    \includegraphics[width=\textwidth]{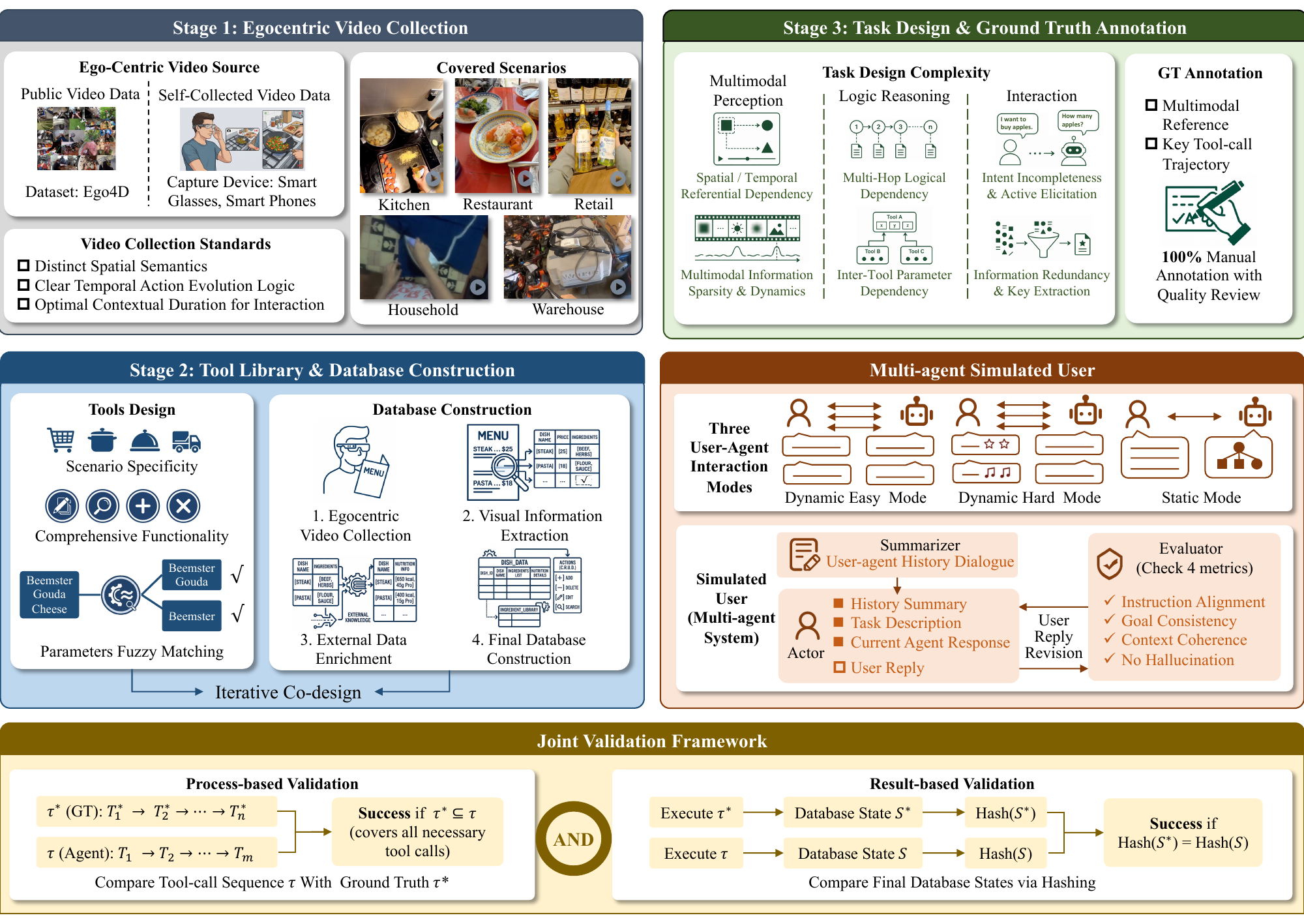}
    \caption{Overview of EgoBench construction and evaluation.}
    \label{fig:pipeline}
\end{figure}

EgoBench is a multimodal environment for evaluating interactive tool-using agents in egocentric settings.
Following the standard formulation of interactive decision-making, we model each task in EgoBench using a POMDP-style framework, denoted by \((S, A, O, T, V)\). Here, \(S\) denotes the state of the simulated user and database in the environment, \(A\) the action space, and \(O\) the observation space. The environment dynamics are driven by tool execution on the database and user-agent interaction, which are formalized as \(T: S \times A \rightarrow \Delta(S)\). In addition, \(V\) denotes the first-person video space, representing the egocentric visual input available to the agent. At each step, the evaluated agent receives the observation \(o \in O\), consisting of the user feedback and tool outputs, together with an egocentric video \(v \in V\) representing the current first-person scene. 
Based on \((o,v)\), the agent selects an action \(a \in A\), which may be either a natural-language response to the user or tool calls over the database.
The environment then implement the tool execution \(T\), updates the dataset state, and returns the next observation. 
We next describe the construction of the benchmark in detail(Fig.~\ref{fig:pipeline}).

\subsection{Egocentric Video Collection}
\label{ego_video}

We collect first-person videos from common real-world daily scenarios (Fig.~\ref{fig:pipeline}, Stage 1). To be included in the benchmark, videos \(V\) must satisfy the following three conditions: (1) they contain objects with clearly distinguishable features, such as unique colors, shapes, or spatial positions, thereby enabling evaluation of the agent's spatial understanding; (2) they exhibit explicit temporal order, such as different user actions occurring sequentially, thereby enabling evaluation of the agent's temporal understanding; and (3) the duration is limited to 10--30s to reflect the typical length of everyday videos captured on wearable devices. The benchmark includes both videos that we personally collected and recorded in real-world settings using smart glasses and mobile phones, as well as videos drawn from the public Ego4D dataset(~\cite{Ego4D}).
These videos in the benchmark span scenarios including Kitchen, Retail, Restaurant, Warehouse and Household. Ethical considerations regarding both the use of Ego4D and our self-collected videos are provided in Appendix \ref{app:ethics}.

\subsection{Tool Library and Database Construction}
After selecting the videos, we design the tool library and database based on the videos and the common daily tasks in the corresponding real-world scenarios.(Fig.~\ref{fig:pipeline}, Stage 2).

In designing the tool library, we follow three principles. (1) \textbf{Scenario specificity:} tools are tailored to the functional requirements of each scenario, and different scenarios are equipped with different tool libraries. For example, the kitchen scenario includes recipe-related tools, whereas the retail scenario includes shopping-cart-related tools; (2) \textbf{Comprehensive functionality:} the tool library covers the full execution pipeline required for task completion, including retrieval tool and state-changing tools that modify the database state; (3) \textbf{Parameter fuzziness matching:} retrieval tools support approximate or partial inputs, allowing agents to invoke tools without exact visual matches. For example, in a retail task, a partial product-name query can return all products whose names match the partial string. In addition, for selected tools, we deliberately introduce \textbf{multiple required arguments}, such that the agent must collect and integrate several relevant attributes before invocation, thereby evaluating its complex tool-using ability. Concrete tool examples are provided in Appendix \ref{app:tool_library}.

For database construction, we intentionally introduce a visual-data information gap. Concretely, we extract visible attributes (e.g., product names, prices) from videos, while augmenting the database with non-visible, scenario-relevant data (e.g., nutritional information). This process ensures alignment between the database and visual content while enabling future task designs that require agents to access invisible information through tool interactions. Consequently, successful task completion depends on the integrated use of perception and tool invocation. Details of the database are provided in Appendix \ref{app:datasets}.

\subsection{Task Design and Ground-truth Annotation}

Based on the established dataset and tool library, we initiate the task design process (Fig.~\ref{fig:pipeline}, Stage 3). To evaluate the three core agent capabilities introduced in Sec.~\ref{Introduction}, each task is built by combining the following difficulty factors. \textbf{Multimodal Perception Complexity:} Each task incorporates multiple multimodal references-grounded in distinct spatial localization or temporal events-to evaluate whether the agent can understand them from the dynamic, sparse egocentric videos. \textbf{Logic Reasoning Complexity:} Each task is designed with conditional branching and multi-attribute filtering(e.g., "If this wine costs more than \$100, add the cheapest sweet wine with the same origin as this one to my cart"). Consequently, by analyzing the agent's tool call sequences and input arguments, we can evaluate whether it has performed correct logical reasoning. Additionally, to organically integrate these two complexities, task design adopts the following aspects: 1. First, designing multimodal references; 2. Then, asking for attributes related to these references but not obtainable from the video, which requires the agent to retrieve relevant information through tool calls; 3. Introducing conditional branching logic and multi-attribute filtering based on the aforementioned attributes (involving more tool calls); 4. Different branches lead to different database modification instructions, which also require tool calls. This four-aspects setup appears at least once in each task(Fig. ~\ref{fig:first_figure}), with harder tasks containing multiple such designs. \textbf{Interaction Complexity:} During task design, specific constraints are also imposed on the simulated user (detailed in Sec.~\ref{user simulation}) to increase interaction difficulty, such as requiring the user to withhold critical parameters initially or introduce redundant noise, to assess active elicitation and information extraction capabilities. See Appendix~\ref{app:task_challenge} for example tasks. 

The ground truth for each task is defined as a sequence of key tool calls required for successful completion, including necessary retrieval operations and state-modifying actions on the database. We further annotate task-critical visual information to support fine-grained error analysis. Each instance is independently annotated by two human annotators. The inter-annotator agreement (IAA) is 96.92\%. For samples with annotation disagreements, the paper authors re-annotate the trajectories and finalize the ground truth after adjudication. Annotators complete only 2.2 tasks per person-hour on average, reflecting high task complexity. Further details of the annotation workflow and quality control are provided in Appendix~\ref{app:annotation_quality}.

\subsection{Multi-agent Simulated User}
\label{user simulation}

EgoBench needs to incorporate a simulated user that assigns pre-designed tasks to the agent through interaction, which is nontrivial to implement. In multi-turn interaction, the simulated user must not only produce contextually coherent responses, but also consistently adhere to the predefined \emph{role}, \emph{constraints}, and \emph{goals} of the above tasks. In terms of role, it must remain a customer rather than an executor or service staff member. In terms of constraints, it must adhere to task requirements faithfully, avoid fabricating unspecified details, and remain robust to persuasion or misleading suggestions from the agent. In terms of goals, it should advance the task when the agent successfully completes a subgoal, while steering the dialogue back to the intended objective whenever the conversation drifts off topic. These requirements make user simulation a long-horizon alignment problem, rather than a simple response generation task.

To achieve these requirements, we adopt a multi-agent framework for user simulation, consisting of an \textbf{Actor}, an \textbf{Evaluator}, and a \textbf{Summarizer}, as illustrated in Fig.~\ref{fig:pipeline}. Specifically, the \textbf{Actor} generates the next user response conditioned on the dialogue history summary and the evaluated agent's latest response. The \textbf{Evaluator} then determines whether the Actor's output satisfies the desired behavioral criteria. When any violation is detected, the user response is not accepted; instead, the Evaluator provides targeted revision feedback, and the Actor regenerates a revised response. The \textbf{Summarizer} maintains a compact representation of the dialogue history by compressing the multi-turn interaction and explicitly tracking task progress, including which steps have already been completed. This updated summary is then used in the next turn to help the Actor reason about the current stage of the task. The pseudocode for the interaction pipeline is provided in Algo.~\ref{alg:multi_agent_simulation}.

\paragraph{Simulated-user evaluation.}
We evaluate every generated user response along four binary dimensions: \emph{role consistency} checks that the user remains a requester rather than acting as the service agent; \emph{instruction following} checks that all explicit constraints are preserved without unsupported additions; \emph{resilience} checks resistance to persuasion or misleading suggestions; and \emph{contextual robustness} checks whether the user corrects contradictions and redirects off-topic dialogue. Let $s_{d,t}^{(m)}\in\{0,1\}$ be the judge score for dimension $m$ at user turn $t$ of trajectory $d$. We apply strict trajectory-level aggregation:
\begin{equation}
\mathrm{Score}^{(m)}=\frac{1}{N}\sum_{d=1}^{N}\prod_{t=1}^{T_d}s_{d,t}^{(m)},
\qquad
\mathrm{AllSuccess}=\frac{1}{N}\sum_{d=1}^{N}\prod_m\prod_{t=1}^{T_d}s_{d,t}^{(m)}.
\end{equation}
Thus, one failed turn causes the entire trajectory to fail that dimension, and \emph{All Success} requires every turn to pass all four dimensions. As reported in Table~\ref{tab:simulated_users}, the multi-agent framework raises All Success from 0.8935 to 0.9772 with Qwen3.5-397B-A17B and from 0.9403 to 0.9676 with GPT-5.6-luna; the former also substantially outperforms the SFT+RL-tuned Qwen3-32B baseline (0.8502). Detailed definitions, examples, judge prompts, training settings, and per-dimension results are provided in Appendices~\ref{appendix:user_metrics} and~\ref{app:user_metric_aggregation}.

Building on this simulated user framework, we further implement three user-agent interaction modes to evaluate different aspects of agent capability. In \textbf{Dynamic Easy Mode}, the agent interacts with the simulated user through standard multi-turn dialogue, measuring its ability to complete tasks under typical interactive conditions. In \textbf{Dynamic Hard Mode}, the interaction remains multi-turn but becomes substantially more challenging: the simulated user may exhibit impatience, provide scenario-relevant but task-irrelevant information, or even terminate the conversation under certain undesirable agent behaviors, such as producing excessively long responses or outputs filled with non-natural-language content. This mode places substantially stronger demands on the agent's robustness. In \textbf{Static Mode}, interaction is removed entirely: the simulated user provides all task requirements at once, and the agent must solve the task within a single user-agent turn, although it may still invoke tools for multiple steps internally. This setting isolates the agent's task-solving ability without interactive uncertainty, while increasing the burden of long-horizon planning. Detailed information about the three user-agent interaction modes is shown in Appendix \ref{app:interaction_mode}.

\subsection{Joint Validation Framework}

To objectively and accurately assess agent performance on our benchmark, we design the following evaluation metrics. Let \(\mathcal{D}\) denote the set of all evaluation tasks, \(N(\mathcal{D})\) the total number of tasks. 

\paragraph{Process-Based Evaluation}
A task $d \in \mathcal{D}$ is considered successful under process-based evaluation if the ground-truth tool-call set \(\tau_d^\ast\) is contained in the agent-generated set \(\tau_d\), i.e., the agent covers all necessary tool calls required by the task: \(\mathrm{\mathbf{ToolSucc}} = N(\{d \in \mathcal{D}: \tau_d^\ast \subseteq \tau_d\}) / N(\mathcal{D})\). This metric measures the proportion of tasks for which the agent completes all required tool-use steps.
Let \(N(\tau_d^\ast)\) be the number of tool calls in the ground-truth set for task \(d\), and \(N(\tau_d^\ast \cap \tau_d)\) the number of tool calls produced by the agent that also appear in the ground-truth set. Then \(\mathrm{{\mathbf{MicroAcc}}} = \sum_{d \in \mathcal{D}} N(\tau_d^\ast \cap \tau_d) / \sum_{d \in \mathcal{D}} N(\tau_d^\ast)\). Unlike task-level success metrics, MicroAcc captures partial progress at the tool-call level and provides a more detailed view of model capability.

\paragraph{Result-Based Evaluation}
A task is considered successful under result-based evaluation if the final database state produced by the agent is identical to that obtained by executing the ground-truth tool-call sequence. Let \(s_d(\tau_d^\ast)\) and \(s_d(\tau_d)\) denote the final database states obtained by executing the ground-truth and agent-generated tool-call sequences, respectively. A task is counted as successful if these two final states are equivalent on all task-relevant database variables: \(\mathrm{{\mathbf{ResultSucc}}} = N(\{d \in \mathcal{D}: s_d(\tau_d^\ast) \equiv s_d(\tau_d)\}) / N(\mathcal{D})\). In practice, this state equivalence is verified by comparing the hash values of the final database states.

\paragraph{Joint Success Rate}
The joint success rate measures the proportion of tasks that are successful under both process-based and result-based evaluation, i.e., the agent must both cover all necessary tool calls and reach the correct final database state: \(\mathrm{{\mathbf{JointSucc}}} = N(\{d \in \mathcal{D}: \tau_d^\ast \subseteq \tau_d \text{ and } s_d(\tau_d^\ast) \equiv s_d(\tau_d)\}) / N(\mathcal{D})\). We treat \textbf{Joint Success Rate} as the primary benchmark metric, as it most directly reflects full end-to-end task completion.

\section{Experiments}
\label{experiments}

\subsection{Experimental Setup}
\label{sec:experimental_setup}

We evaluate eight video-MLLMs spanning both closed-source models (Gemini-3.1-Pro~\cite{gemini31pro2026}, Doubao-Seed-2.0-Pro~\cite{doubao2pro2026}, Qwen3.6-Plus~\cite{qwen36plus2026}, and GLM-5V-Turbo~\cite{glm5vturbo2026}) and open-source models (Qwen3-VL-235B~\cite{bai2025qwen3vl}, Qwen3.5-397B-A17B~\cite{qwen2026qwen35}, Kimi-K2.6~\cite{moonshotai2026kimik26}, and MiMo-V2.5-Omni~\cite{mimov25}). 
All models are evaluated with default inference parameters and unified prompt templates(Appendix~\ref{app:prompts}). For each task, user-agent interactions are limited to 10 turns in one trajectory. While tool calls are unrestricted per turn, the cumulative total is capped at 200. These limits far exceed human reference trajectories ($\leq 5$ turns, $\leq 50$ tool calls) while preventing unbounded computation. We employ Qwen3.5-397B-A17B to drive the multi-agent simulated user. 
We first compute each statistic within a scenario--interaction-mode stratum from the trajectories currently completed for that model. We then weight the five scenario strata by the benchmark's full task counts, with equal weight for the three interaction modes. 

\subsection{Performance Analysis}
\label{sec:performance_analysis}

\begin{table*}[t]
\centering
\caption{Performance by Difficulty on EgoBench.}
\label{tab:difficulty_results}
\footnotesize
\setlength{\tabcolsep}{4pt}
\resizebox{\textwidth}{!}{%
\begin{tabular}{clcccccccccccc}
\toprule
 & & \multicolumn{4}{c}{\textbf{Easy}} & \multicolumn{4}{c}{\textbf{Hard}} & \multicolumn{4}{c}{\textbf{Static}} \\
\cmidrule(lr){3-6}\cmidrule(lr){7-10}\cmidrule(lr){11-14}
 & \textbf{Model} & \textbf{Micro} & \textbf{Tool} & \textbf{Result} & \textbf{Joint} & \textbf{Micro} & \textbf{Tool} & \textbf{Result} & \textbf{Joint} & \textbf{Micro} & \textbf{Tool} & \textbf{Result} & \textbf{Joint} \\
\midrule
\includegraphics[height=0.3cm]{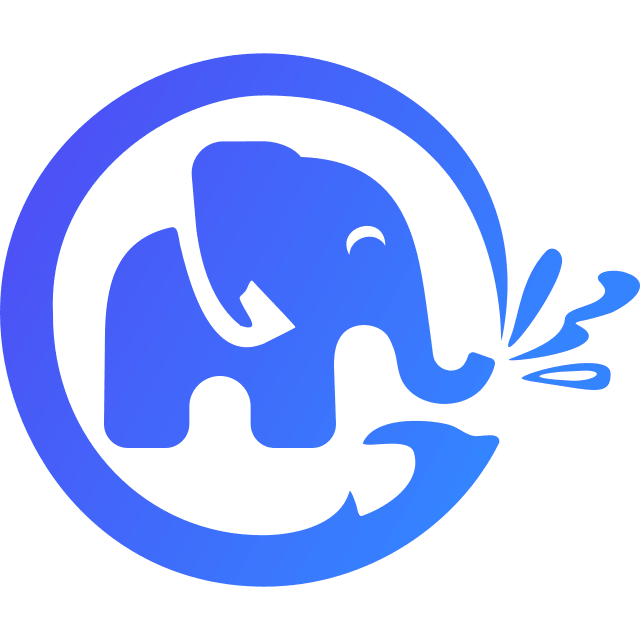} & GLM-5V-Turbo & 43.71 & 18.01 & 14.85 & 11.23 & 47.33 & 19.87 & 15.65 & 11.95 & 48.04 & 19.90 & 23.50 & 16.94 \\
\includegraphics[height=0.3cm]{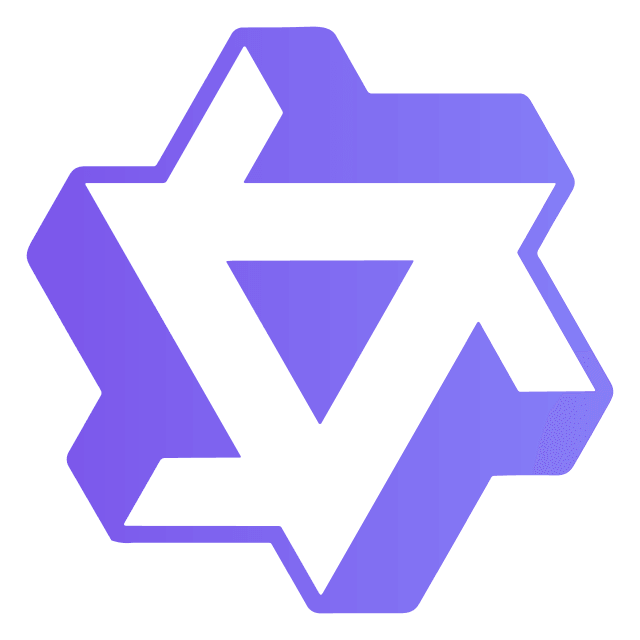} & Qwen3-VL-235B & 27.90 & 5.66 & 5.66 & 3.27 & 7.05 & 0.75 & 0.69 & 0.44 & 20.39 & 1.26 & 1.95 & 0.82 \\
\includegraphics[height=0.3cm]{logo/qwen-color} & Qwen3.6-Plus & 55.43 & 30.25 & 29.43 & \underline{24.84} & 46.20 & 22.70 & 21.51 & \underline{17.23} & 24.71 & 9.94 & 12.39 & 8.30 \\
\includegraphics[height=0.3cm]{logo/qwen-color} & Qwen3.5-397B-A17B & 53.92 & 28.87 & 29.87 & 24.03 & 43.03 & 20.25 & 20.69 & 16.60 & 35.67 & 18.43 & 23.21 & 16.79 \\
\includegraphics[height=0.3cm]{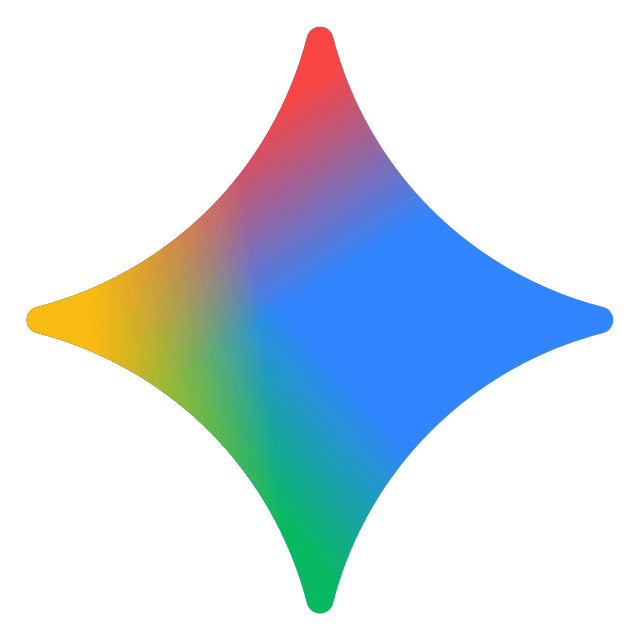} & Gemini-3.1-Pro & \textbf{71.35} & \textbf{47.08} & \textbf{40.86} & \textbf{36.21} & \textbf{66.01} & \textbf{40.84} & \textbf{30.92} & \textbf{27.03} & \textbf{66.43} & \textbf{43.82} & \textbf{52.65} & \textbf{41.60} \\
\includegraphics[height=0.3cm]{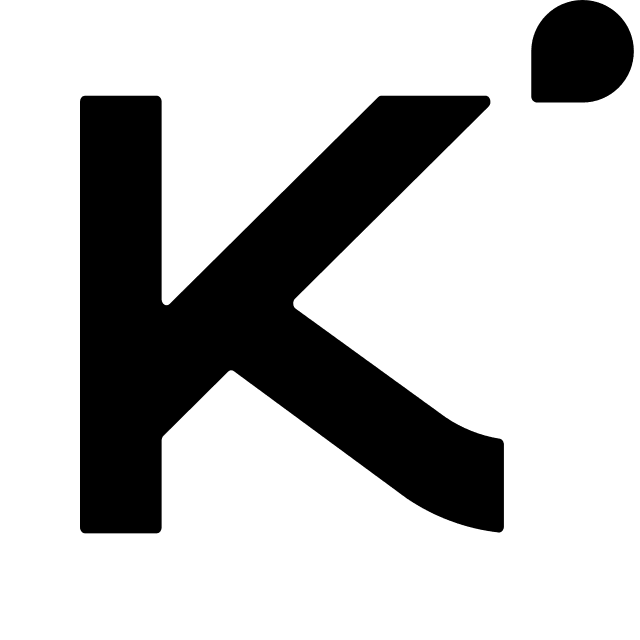} & Kimi-K2.6 & 50.30 & 26.75 & 28.55 & 22.26 & 43.70 & 20.96 & 19.71 & 15.04 & 37.49 & 20.69 & 24.21 & 18.86 \\
\includegraphics[height=0.3cm]{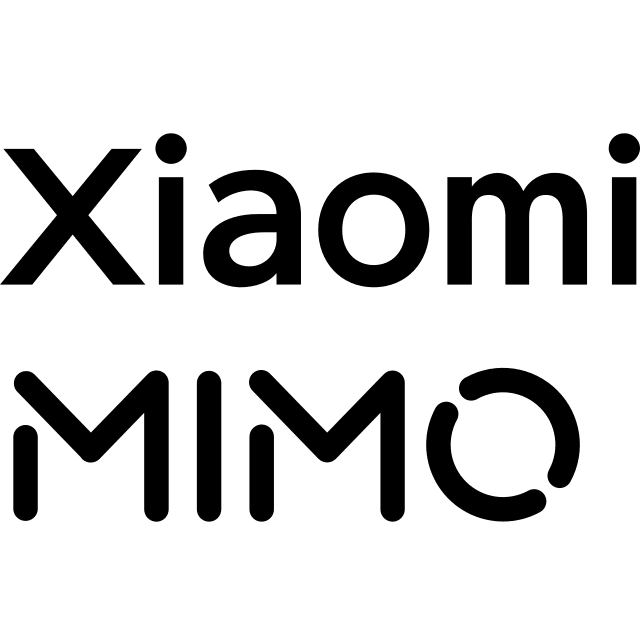} & MiMo-V2.5-Omni & 42.92 & 16.73 & 16.04 & 12.52 & 26.77 & 8.99 & 6.54 & 5.03 & 23.92 & 5.91 & 6.42 & 4.59 \\
\includegraphics[height=0.3cm]{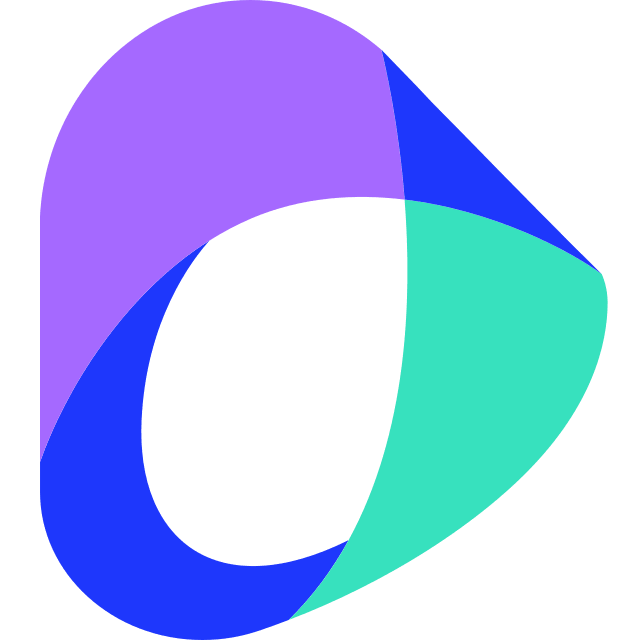} & Doubao-Seed-2.0-Pro & \underline{59.59} & \underline{30.96} & \underline{32.62} & 24.68 & \underline{55.15} & \underline{24.75} & \underline{24.00} & 15.93 & \underline{58.78} & \underline{28.28} & \underline{38.44} & \underline{26.00} \\
\bottomrule
\end{tabular}
}%
\end{table*}

\textbf{Overall Analysis} Table~\ref{tab:difficulty_results} presents the overall performance of eight models on EgoBench across three interaction modes. The top-performing model, Gemini-3.1-Pro, achieves an average Joint Success Rate of 34.95\% across the three modes, while the best open-source model, Qwen3.5-397B-A17B, reaches 19.14\%. These results underscore that current models remain far from achieving reliable end-to-end problem-solving capabilities in real-world egocentric scenarios.
Table~\ref{tab:user_backbone_sensitivity} shows that using different user models does not affect the relative performance of the evaluated agent models.

\begin{wraptable}{l}{0.6\textwidth}
\centering
\caption{Overall Joint Success Rate (\%) with Qwen3.5-397B-A17B and GPT-5.6-luna as the simulated-user backbone, averaged across the three interaction modes.}
\label{tab:user_backbone_sensitivity}
\footnotesize
\setlength{\tabcolsep}{4pt}
\begin{tabular}{lcc}
\toprule
\textbf{Service model} & \makecell{\textbf{Qwen3.5-397B-A17B}\\\textbf{user}} & \makecell{\textbf{GPT-5.6-luna}\\\textbf{user}} \\
\midrule
Gemini-3.1-Pro & \textbf{34.9} & \textbf{36.5} \\
Doubao-Seed-2.0-Pro & 22.2 & 25.9 \\
Kimi-K2.6 & 18.7 & 21.6 \\
Qwen3.5-397B-A17B & 19.1 & 22.3 \\
Qwen3.6-Plus & 16.8 & 17.5 \\
GLM-5V-Turbo & 13.4 & 15.2 \\
MiMo-V2.5-Omni & 7.4 & 6.4 \\
Qwen3-VL-235B & 1.5 & 1.5 \\
\bottomrule
\end{tabular}
\end{wraptable}

\textbf{Interaction Mode Analysis.} Each model exhibits varying levels of performance across interaction modes. Most models exhibit a decline in success rate in \textbf{Dynamic Hard Mode} compared to \textbf{Dynamic Easy Mode}, exposing deficiencies in their robust dialog management and interference-resistant state tracking capabilities. Additionally, \textbf{Static Mode} did not significantly improve task accuracy as anticipated. On one hand, this confirms that the benchmark poses challenges in both multimodal understanding and logic-driven tool use. On the other hand, certain models even underperform in \textbf{Static Mode} compared to the \textbf{Dynamic Easy Mode}. Case analysis reveals that the multi-turn interaction mechanism plays a compensatory role: First, by presenting only a single requirement per turn, the simulated user performs task planning on behalf of the evaluated models to some extent. Second, the user can promptly correct fundamental errors made by the model (e.g., pointing out when the model incorrectly replies, "I have found a bitter wine for you..." after being asked to "find a sweet wine"), thereby providing real-time verification throughout the interaction. In summary, although \textbf{Static Mode} delivers all task instructions and complex logical constraints in a single turn, it increases the cognitive load for models with limited planning and reflection capabilities, ultimately leading to lower accuracy.


\begin{wrapfigure}{l}{0.6\textwidth}
    \centering
    \includegraphics[width=\linewidth]{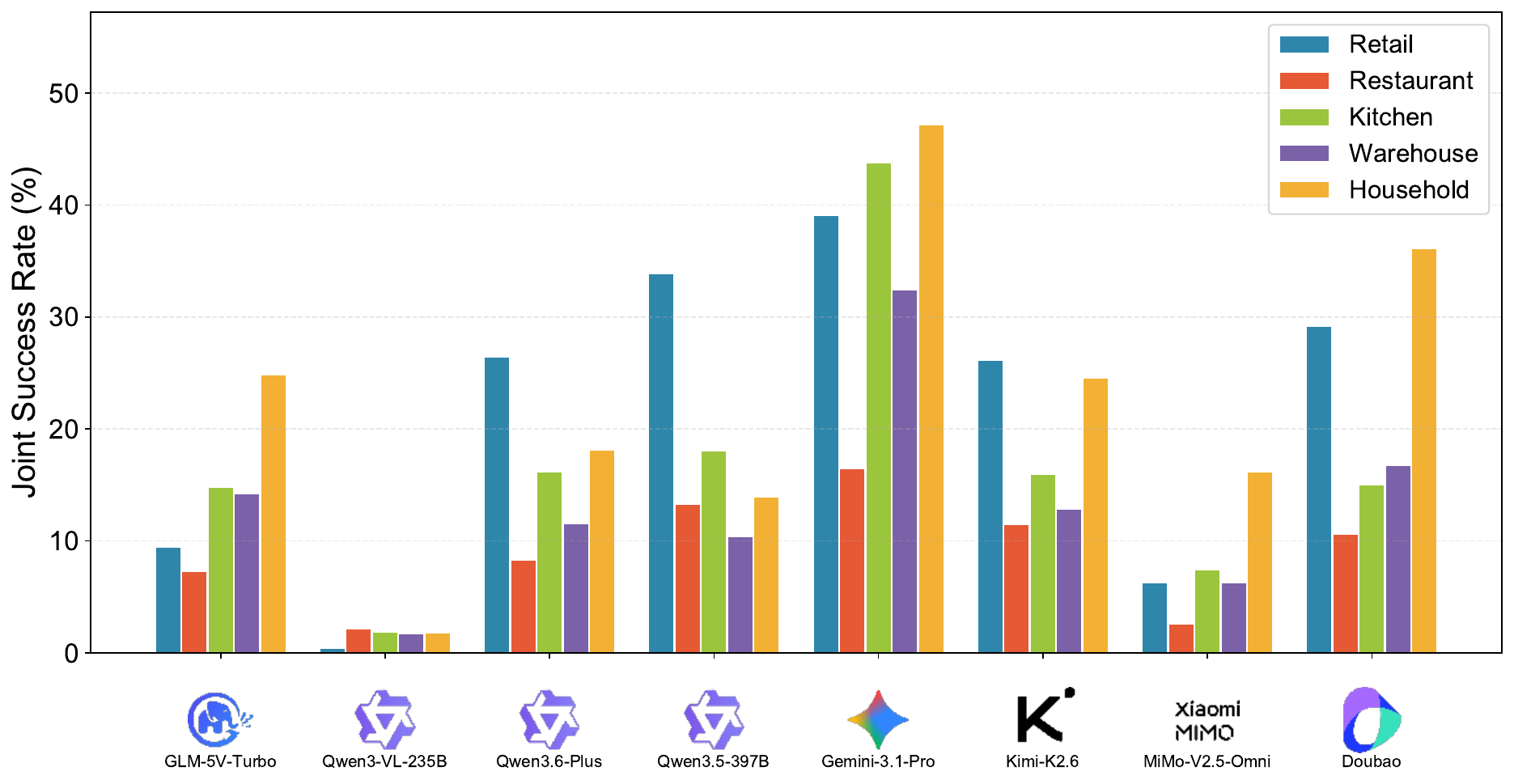}
    \caption{Joint success rate (\%) across the five EgoBench scenarios.}
    \label{fig:scenario_analysis}
\end{wrapfigure}

\textbf{Scenario Analysis} We further decompose model performance across the five task scenarios illustrated in Figure~\ref{fig:scenario_analysis}. Averaged across the eight models, Joint Success Rate is highest in \textbf{Household} (22.86\%), followed by \textbf{Retail} (21.38\%), \textbf{Kitchen} (16.66\%), \textbf{Warehouse} (13.31\%), and \textbf{Restaurant} (9.05\%). Household and Retail benefit from comparatively easy visual grounding: familiar household objects and distinctive product packaging, colors, and positions make it easier to map a reference to the correct database entity. The Warehouse scenario is more challenging because, compared to the neatly arranged supermarket shelves and clean household spaces, its visual environment is significantly more cluttered and many objects or equipment are uncommon. This makes it substantially harder for agents to accurately localize the items referred to by the user and complete the task reliably. Restaurant is the most difficult because dense menus and similar dishes complicate grounding, after which preference, nutritional, and allergen constraints require multi-attribute filtering.

\subsection{Error Attribution Analysis}

\begin{figure*}[t]
    \centering
    \includegraphics[width=\textwidth]{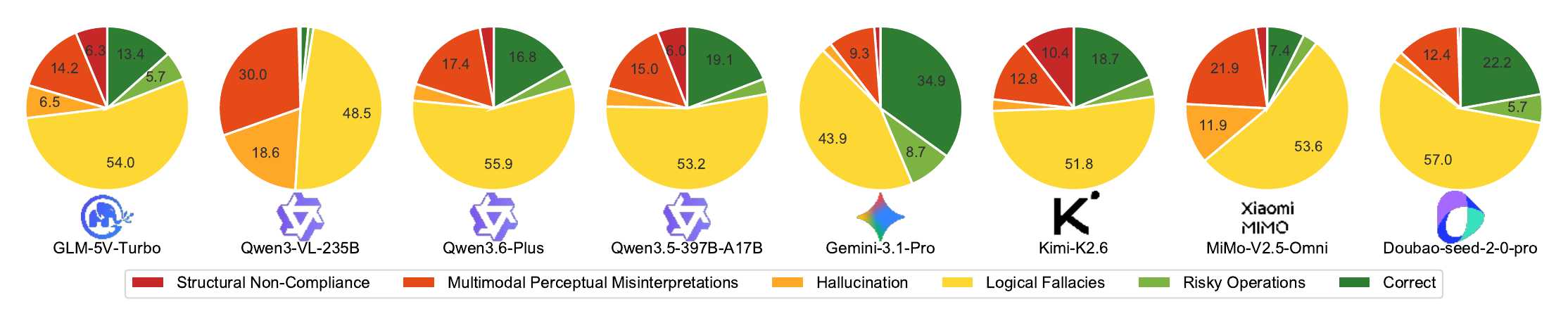}
    \caption{Multi-dimensional error statistics across different models. Scenario--mode rates use currently completed trajectories and are aggregated with the fixed benchmark task-count weights.}
    \label{fig:error_analysis}
\end{figure*}

To conduct a fine-grained assessment of model capabilities, we employed a multi-dimensional error analysis method to systematically categorize and quantify the failure trajectories of each model (see Fig.~\ref{fig:error_analysis}). Detailed methodologies for objective error classification are provided in Appendix \ref{app:error_classification}. The specific error dimensions are defined as follows: \textbf{Structural Non-Compliance}: The model generates incorrectly formatted tool calls, uses invalid tool names or parameters, or in Static Mode asks the user for additional information even though it is asked to directly perform tool calls based on the complete task description, resulting in immediate task failure. \textbf{Multimodal Perceptual Misinterpretations}: The agent fails to correctly understand or extract multimodal referential information from the user and egocentric videos. \textbf{Hallucination}: The agent fabricates content not present in the context during tool invocation or when responding to the user. \textbf{Logical Fallacies}: The agent makes errors in logical conditional judgments and in the processes of tool retrieval and invocation. \textbf{Risky Operations}: The agent executes risky operations that modify database content beyond the scope of the user's request. 

As shown in the Fig.~\ref{fig:error_analysis}, failures are predominantly caused by Multimodal Perceptual Misinterpretations and Logical Fallacies. This indicates that while current models exhibit strong instruction-following capabilities in terms of output format, they still require improvement in multimodal understanding and logical reasoning. Detailed scenario-based statistics of these error causes are presented in Fig. \ref{fig:scenario_error} in the Appendix.
Within each scenario, error-category and Correct proportions are first computed separately for each interaction mode from the trajectories currently available, then averaged across the three modes. Overall model proportions are weighted by the full task count of each scenario. Thus the scenario-level and overall Correct sectors use the same fixed target distribution as the main performance results.


\subsection{Execution Efficiency Analysis}
\label{efficiency_analysis}

\begin{figure*}[t]
    \centering
    \includegraphics[width=\textwidth]{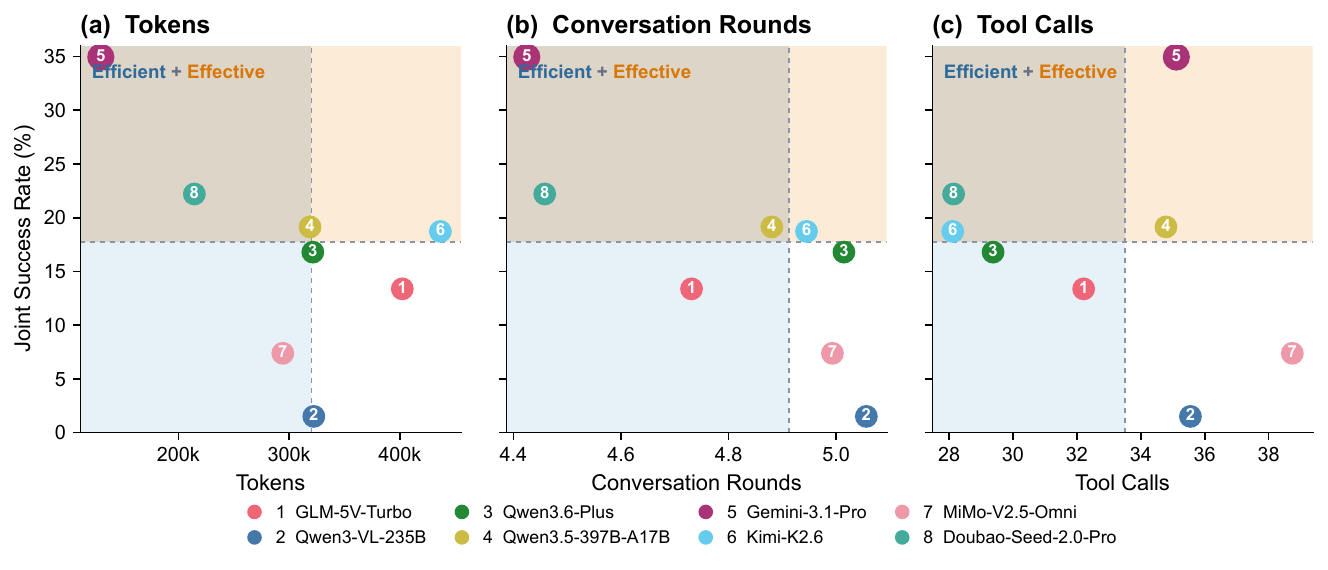}
    \caption{Efficiency--effectiveness trade-offs across models. Dashed lines denote cross-model medians; lower resource use (tokens, conversation rounds, and tool calls) and higher joint success are preferred.}
    \label{fig:efficiency_analysis}
\end{figure*}

Beyond success rates, we examine how efficiently models use computational and interaction resources.
\textbf{Token efficiency:} As shown in Fig.~\ref{fig:efficiency_analysis}(a), token consumption is negatively correlated with joint success rate. For instance, Gemini-3.1-Pro achieves the highest success rate with minimal token usage, suggesting that models with stronger capabilities effectively avoid redundant outputs and ineffective exploration. Separate statistics for input and output tokens are shown in Fig.~\ref{fig:input_token} and Fig.~\ref{fig:output_token} in the Appendix. \textbf{User-Agent interaction efficiency:} As shown in Fig.~\ref{fig:efficiency_analysis}(b), increased interaction rounds cannot compensate for models' fundamental reasoning or perceptual deficits. For instance, Qwen3-VL-235B and MiMo-V2.5-Omni implement comparable or even more rounds than top-performing models yet achieve significantly lower success rates. \textbf{Tool calls efficiency:} Fig.~\ref{fig:efficiency_analysis}(c) indicates that there is no direct correlation between success rate and the number of calls. However, a comparison of panels (a) and (c) reveals that some models, such as Gemini-3.1-Pro, maintain low token consumption despite a high volume of tool calls. Further analysis attributes this to Gemini's ability to invoke multiple tools concurrently within a single turn. In contrast, some models, such as Kimi-K2.6, typically invoke only a few tools per turn, necessitating the repeated transmission of video and historical dialogue context, which significantly increases token overhead. Appendix~\ref{app:tool_efficiency_metrics} formally defines additional metrics that expose redundant or brute-force tool use.


\section{Conclusions}
\label{Conclusions}
We introduce \textbf{EgoBench}, an interactive multimodal benchmark for evaluating tool-using agents in open, real-world environments. EgoBench jointly assesses visual grounding, dynamic user interaction and multi-hop tool use through a environment containing egocentric videos, scenario-specific tools, structured databases, and a task-constrained user simulator. It also provides an objective joint evaluation framework that checks both tool-call process and final execution results. Experiments show that current models perform poorly on Egobench, revealing a substantial gap between existing capabilities and reliable real-world deployment.




\bibliographystyle{plainnat}
\bibliography{references}

\newpage
\appendix
\section*{Appendix}
\section{AI Use Statement}

In this work, we used generative AI tools to polish the manuscript, assist with writing parts of the code. We did not use generative AI tools to generate synthetic data, formulate hypotheses, design the research methodology, interpret results, develop theoretical claims, or make scientific conclusions. All AI-assisted outputs were carefully reviewed, verified, and revised by the authors as needed. Specifically, LLM-generated code was verified and tested for correctness by two authors, and all AI-polished text was manually inspected to ensure it accurately reflects our original technical contributions without introducing unsupported claims. We take responsibility for the final content of this work, including text, claims or artifacts produced with the aid of generative AI.

\section{Ethics Statement}
\label{app:ethics}

EgoBench includes videos captured by the authors. During data collection, we took care to avoid recording identifiable individuals whenever possible. In cases where the captured content involved commercial spaces, products, restaurant food, menus, or other business-related visual materials, permission was obtained from the relevant business operators prior to their inclusion in the dataset and any subsequent research use. These measures were taken to ensure that our data collection and presentation adhered to appropriate ethical and privacy considerations.

In addition to self-collected, EgoBench also makes use of the Ego4D dataset. In the main paper, we explicitly acknowledge the use of Ego4D in Section \ref{ego_video}, and we cite the dataset in accordance with the official citation guidelines provided on the project website\footnote{\url{https://ego4d-data.org}}. Furthermore, if any video frames or excerpts from Ego4D are included in the paper, their use is ensured to comply with the relevant terms of the Ego4D data use agreement, including the provision that permits reproduction in research publications.

Finally, regarding the annotation process for ground-truth tool-call trajectories and task-critical visual information, we ensured fair labor practices. All participating annotators received compensation at a rate above the minimum wage in our country.

\section{Reproducibility Statement}

We provide the main benchmark construction and evaluation details in Sections 3 and 4, and include additional implementation details, prompts, task examples, and error analysis in the appendix. The code and data repository is publicly available at \url{https://github.com/Wind-Flipped/EgoBench}, which contains the resources needed to reproduce our benchmark and experiments.

\section{EgoBench Overview}
\label{app:scenario}

\begin{wraptable}{r}{0.58\textwidth}
\centering
\caption{Key statistics of each scenario.}
\label{tab:scenario}
\begin{tabular}{lcccc}
\toprule
Scenario & \#Videos & \#Tools & \#Items & \#Tasks \\
\midrule
Retail & 10 & 20 & 768 & 454 \\
Kitchen & 4 & 28 & 102 & 175 \\
Restaurant & 11 & 27 & 289 & 316 \\
Warehouse & 25 & 17 & 1151 & 375 \\
Household & 18 & 16 & 935 & 270 \\
\midrule
Total & 68 & 108 & 3245 & 1590 \\
\bottomrule
\end{tabular}
\end{wraptable}

To ensure that our task design reflects realistic interactions with agents in everyday scenarios while remaining reliable for evaluation, we consider five representative scenarios: \textbf{Retail}, \textbf{Kitchen}, \textbf{Restaurant}, \textbf{Warehouse}, and \textbf{Household}. The statistics of each scenario are summarized in Table~\ref{tab:scenario}, where the number of videos is computed from the distinct \texttt{image\_path} values within each scenario. Task examples for each scenario are shown in Appendix \ref{app:example_tasks}. 

The Retail scenario involves tasks such as identifying referred products in a store environment, handling queries related to products, and performing operations on the shopping cart such as adding, removing, and calculating items.

The Restaurant scenario involves tasks such as identifying menus or dishes, selecting restaurants based on user preferences and menu information, handling queries related to dishes or set meals, and performing operations on the order such as adding, removing, and calculating items.

The Kitchen scenario involves tasks such as identifying ingredients in the scene, determining the recipe and cooking step the user is following, handling queries related to ingredients or recipes, and performing operations such as adding, removing, and calculating recipes in the menu and ingredients in the shopping cart.

The Warehouse scenario involves identifying referred equipment, retrieving equipment attributes, usage and maintenance records, reasoning about maintenance schedules and service life, and managing equipment action lists. These tasks challenge the agent to combine visual grounding with multi-step retrieval, temporal reasoning, and stateful tool use.

The Household scenario involves identifying household items, retrieving purchase, storage, usage, and care information, comparing items under user-specified conditions, and managing household action lists. These tasks emphasize visual disambiguation, reasoning over dates and usage periods, and coordinating multiple updates across related items.

\section{Design of Datasets and Tool Library}
\subsection{Design of Tool Library}
\label{app:tool_library}

We design scenario-specific toolsets for each environment, where every scenario is equipped with a rich collection of CRUD-style(Create, Read, Update, Delete) APIs and retrieval functions with fuzzy matching support. This design is intended to satisfy the principles of \textbf{Scenario Specificity}, \textbf{Comprehensive Functionality}, and \textbf{Parameter Fuzziness Matching} discussed in the main text. In addition, for selected tools, we deliberately require multiple arguments so that the agent must first gather and integrate several relevant attributes before invoking the tool, thereby testing its ability to perform compositional and complex tool use. For example, to call \texttt{add\_to\_cart}, the agent must obtain the product's \texttt{category}, \texttt{price}, \texttt{tax\_rate}, and \texttt{discount} before the item can be added to the cart.

\begin{table}[h!]
\centering
\caption{Tool library for the Restaurant and Retail scenarios}
\label{tab:restaurant_retail}
\resizebox{\textwidth}{!}{%
\begin{tabular}{@{}lp{6cm}p{6cm}@{}}
\toprule
 & \textbf{Restaurant} & \textbf{Retail} \\
\midrule
\textbf{Read APIs} & 
\begin{tabular}[t]{@{}l@{}}
get\_dish\_nutrition \\ get\_dish\_allergens \\ get\_dish\_taste\_profile \\ 
get\_dish\_price \\ get\_dish\_discount \\ get\_tax\_rate \\ get\_set\_meal\_details \\ 
find\_set\_meals\_containing\_dish \\ get\_user\_order\_summary \\ 
find\_dishes\_by\_category \\ find\_dishes\_by\_nutritional\_tag \\ 
find\_dishes\_by\_taste \\ filter\_dishes\_by\_price\_range \\ 
list\_all\_discounted\_dishes
\end{tabular}
&
\begin{tabular}[t]{@{}l@{}}
get\_nutrition \\ get\_price \\ get\_tax\_rate \\ get\_category \\ 
get\_discount \\ get\_cart \\ get\_shopping\_list \\ 
find\_products\_by\_nutritional\_characteristic \\ 
find\_products\_by\_taste \\ find\_products\_by\_country\_of\_origin \\ 
find\_products\_by\_price\_range \\ list\_discounted\_products
\end{tabular} \\
\midrule
\textbf{Write APIs} & 
\begin{tabular}[t]{@{}l@{}}
add\_dish\_to\_catalog \\ remove\_dish\_from\_catalog \\ 
update\_dish\_price \\ update\_dish\_discount \\ create\_set\_meal \\ 
add\_dish\_to\_order \\ remove\_dish\_from\_order \\ clear\_user\_order \\ 
add\_set\_meal\_to\_order \\ remove\_set\_meal\_from\_order
\end{tabular}
&
\begin{tabular}[t]{@{}l@{}}
add\_product \\ delete\_product \\ add\_to\_cart \\ 
remove\_from\_cart \\ clear\_cart
\end{tabular} \\
\midrule
\textbf{Calculation APIs} & 
\begin{tabular}[t]{@{}l@{}}
compute\_total\_payment \\ compute\_total\_tax \\ 
compute\_total\_nutrition
\end{tabular}
&
\begin{tabular}[t]{@{}l@{}}
compute\_total\_payment \\ compute\_total\_tax \\ 
compute\_total\_nutrition
\end{tabular} \\
\bottomrule
\end{tabular}%
}
\end{table}

\begin{table}[h!]
\centering
\caption{Tool library for the Kitchen scenario}
\label{tab:kitchen}
\resizebox{\textwidth}{!}{%
\begin{tabular}{@{}lp{12cm}@{}}
\toprule
 & \textbf{Kitchen} \\
\midrule
\textbf{Read APIs} &
\begin{tabular}[t]{@{}l@{}}
get\_cooking\_steps \\ get\_recipe\_allergens \\ find\_recipes\_by\_allergen \\
get\_recipe\_taste \\ find\_recipes\_by\_taste \\ get\_recipe\_ingredients \\
find\_recipes\_by\_ingredient \\ get\_recipe\_nutritional\_characteristics \\
find\_recipes\_by\_nutritional\_characteristics \\ get\_all\_recipe\_names \\
get\_ingredient\_shelf\_life \\ find\_ingredients\_by\_expiry\_date \\
get\_ingredient\_location \\ find\_ingredients\_by\_location \\
get\_ingredient\_nutrition \\ get\_ingredient\_quantity \\
get\_all\_ingredient\_names \\ get\_ingredients\_by\_category \\
find\_ingredient\_category \\ get\_current\_menu \\ get\_current\_shopping\_list
\end{tabular} \\
\midrule
\textbf{Write APIs} &
\begin{tabular}[t]{@{}l@{}}
add\_recipe\_to\_menu \\ remove\_recipe\_from\_menu \\
add\_to\_shopping\_list \\ remove\_from\_shopping\_list
\end{tabular} \\
\midrule
\textbf{Calculation APIs} &
\begin{tabular}[t]{@{}l@{}}
tally\_total\_nutritional\_characteristics \\
tally\_total\_tastes \\ compute\_total\_nutritions
\end{tabular} \\
\bottomrule
\end{tabular}%
}
\end{table}

\begin{table}[h!]
\centering
\caption{Tool library for the Warehouse and Household scenarios}
\label{tab:warehouse_household}
\resizebox{\textwidth}{!}{%
\begin{tabular}{@{}lp{6cm}p{6cm}@{}}
\toprule
 & \textbf{Warehouse} & \textbf{Household} \\
\midrule
\textbf{Read APIs} &
\begin{tabular}[t]{@{}l@{}}
get\_all\_equipment \\ get\_equipment\_category \\
find\_equipment\_by\_category \\ get\_equipment\_brand \\
find\_equipment\_by\_brand \\ get\_equipment\_country\_of\_origin \\
find\_equipment\_by\_country\_of\_origin \\ get\_equipment\_purchase\_date \\
find\_equipment\_by\_purchase\_date\_range \\
get\_equipment\_ideal\_storage\_location \\
find\_equipment\_by\_ideal\_storage\_location \\
get\_equipment\_usage\_history \\
get\_equipment\_maintenance\_history \\ get\_equipment\_action\_list \\
get\_equipment\_maintenance\_and\_usage\_instructions
\end{tabular}
&
\begin{tabular}[t]{@{}l@{}}
get\_all\_item\_names \\ get\_item\_category \\
find\_items\_by\_category \\ get\_item\_ideal\_storage\_location \\
find\_items\_by\_ideal\_storage\_location \\ get\_item\_purchase\_date \\
find\_items\_by\_purchase\_date\_range \\ get\_item\_purchase\_price \\
find\_items\_by\_purchase\_price\_range \\ get\_item\_usage\_history \\
find\_items\_used\_during\_date\_range \\
get\_item\_maintenance\_history \\
get\_item\_maintenance\_and\_usage\_instructions \\
get\_household\_action\_list
\end{tabular} \\
\midrule
\textbf{Write APIs} &
\begin{tabular}[t]{@{}l@{}}
add\_to\_equipment\_action\_list \\ remove\_from\_equipment\_action\_list
\end{tabular}
&
\begin{tabular}[t]{@{}l@{}}
add\_to\_household\_action\_list \\ remove\_from\_household\_action\_list
\end{tabular} \\
\bottomrule
\end{tabular}%
}
\end{table}

The tool tables for the five scenarios are shown in Tables~\ref{tab:restaurant_retail}, \ref{tab:kitchen}, and~\ref{tab:warehouse_household}. We also provide documentation for some of the tools as examples in Appendix \ref{app:tool_example}.

\subsection{Design of Datasets}
\label{app:datasets}

For each scenario, we construct a corresponding structured database to support realistic task execution and tool interaction. These databases contain the entities, attributes, and records required by the scenario-specific tools, enabling the agent to retrieve, compare, update, and manipulate information grounded in the environment. The dataset design aims to provide sufficient coverage of real-world operating conditions while supporting the evaluation of retrieval, reasoning, and multi-step decision-making abilities.

The database statistics for all scenarios are shown in Table~\ref{tab:database_statistics}. We also provide part of the database as illustrative examples in Appendix \ref{app:database_example}.

\begin{table}[htbp]
\centering
\caption{Database Statistics Across Scenarios}
\label{tab:database_statistics}
\renewcommand{\arraystretch}{1.3}
\begin{tabular}{llp{6cm}l}
\toprule
\textbf{Scenario} & \textbf{Item} & \textbf{Properties} & \textbf{Count} \\
\midrule
\multirow{3}{*}{Retail}
  & products            & name, category, price, tax\_rate, discount, nutritional\_characteristics, taste, country\_of\_origin, nutrition, allergens & 468 \\
  & user\_carts         & user\_id, items                                                                                                           & 150 \\
  & user\_shopping\_lists & user\_id, items                                                                                                         & 150 \\
\midrule
\multirow{4}{*}{Kitchen}
  & ingredients         & name, quantity, expiry\_date, storage\_location, category, nutrition                                                      & 61 \\
  & recipes             & name, ingredients, steps, allergens, taste, nutritional\_characteristics, nutrition                                       & 21 \\
  & user\_menus         & user\_id, recipes                                                                                                         & 10 \\
  & user\_shopping\_lists & user\_id, items                                                                                                         & 10 \\
\midrule
\multirow{3}{*}{Restaurant}
  & dishes              & name, restaurant\_name, category, price, tax\_rate, discount, nutritional\_characteristics, taste, allergens, nutrition & 229 \\
  & set\_meals          & name, restaurant\_name, set\_meal\_discount, included\_dishes                                                          & 20 \\
  & user\_orders        & user\_id, restaurant\_name, items                                                                                       & 40 \\
\midrule
\multirow{2}{*}{Warehouse}
  & equipment           & equipment, category, brand, country\_of\_origin, purchase\_date, ideal\_storage\_location, usage\_history, maintenance\_history, maintenance\_and\_usage\_instructions, appears\_in\_video & 1051 \\
  & equipment\_action\_lists & user\_id, actions                                                                                                    & 100 \\
\midrule
\multirow{2}{*}{Household}
  & items               & name, category, ideal\_storage\_location, purchase\_date, purchase\_price, usage\_history, maintenance\_history, care\_and\_usage\_instructions, appears\_in\_video & 863 \\
  & household\_action\_lists & user\_id, actions                                                                                                    & 72 \\
\bottomrule
\end{tabular}
\end{table}

\section{Human Annotation and Ground-Truth Quality Control}
\label{app:annotation_quality}

The ground-truth tool-call trajectories and task-critical visual information are manually annotated. We use a dedicated web interface to provide all annotators with a consistent annotation workflow. Each instance is independently annotated by two human annotators, and the resulting inter-annotator agreement (IAA) is \textbf{96.92\%}. Every instance on which the two annotations disagree is escalated to the paper authors, who re-annotate the trajectory and adjudicate the final ground truth. All participating annotators receive compensation at a rate above the minimum wage in our country.

\section{The Complexity of Tasks}
\label{app:task_challenge}

\paragraph{1. Multimodal Perception Complexity}
Egocentric videos introduce unique perceptual challenges that differ substantially from traditional third-person or static-image datasets. Compared with standard multimodal settings, visual understanding under the egocentric perspective is inherently more difficult due to restricted fields of view, partial observations, continuous camera motion, and frequent hand-object interactions. In EgoBench, we focus on the following two aspects:
\begin{itemize}
    \item \textbf{Multimodal Information Sparsity and Dynamics}: Due to the motion characteristics, egocentric videos often exhibit motion blur, severe shaking, occlusion, and rapidly changing viewpoints. The agent must therefore possess robust visual feature extraction capabilities in order to maintain accurate perception under degraded input quality. In practice, this means identifying the key clear frames from a sequence of unstable or blurred observations and grounding user intent in dynamically changing visual evidence.
    \item \textbf{Spatial/Temporal Referential Dependency}: In natural first-person interaction, users often do not explicitly name target objects, but instead refer to them through viewpoint-dependent expressions grounded in the current scene and recent actions. To reflect this property, we do not explicitly provide the names of visible objects in the task descriptions. Instead, tasks include a variety of egocentric referential expressions, such as spatial relations (e.g., "the bottle on the left"), body- or action-centered references (e.g., "the one on my right-hand side" or "the one I just picked up"), appearance attributes (e.g., color and shape), and temporal cues (e.g., "the second item pointed to by the finger"). The agent must combine \textit{visual context} and linguistic instructions to accurately resolve these references and eliminate ambiguity. Moreover, key events in long videos may appear only briefly, so the tasks require not only understanding of the current frame, but also long-horizon temporal modeling ability to precisely locate specific past events or action segments.
\end{itemize}

\paragraph{2. Reasoning and Tool Usage Complexity}
As a benchmark for tool-using agents, EgoBench evaluates not only perception, but also cognition and execution. This dimension measures the logical depth with which an agent transforms perceived information into effective action:
\begin{itemize}
    \item \textbf{Multi-Hop Logical Reasoning}: Complex tasks in EgoBench usually cannot be completed in a single step. We deliberately design each task to require at least one reasoning step before all tool calls can be completed correctly. Many tasks must be decomposed into multiple subtasks, requiring the agent to construct a coherent chain of reasoning and execute operations sequentially. Importantly, this reasoning is not limited to explicit conditional branching, but also includes compositional constraint satisfaction, candidate filtering, comparison, and selection. For example, a representative instruction is: "I only have \$10 and want to buy two bottles of the drink I'm holding in my left hand. If the money is not enough, then buy only one bottle." To succeed, the agent must first retrieve the price of the drink held in the left hand, perform numerical reasoning, determine which branch of the condition applies, and then invoke the subsequent tools accordingly. Similarly, tasks such as "Help me find a bottle of wine that has the same country of origin as the one I pointed to earlier, is sweet in taste, and has the lowest price, and add it to the cart" require the agent to jointly search across multiple constraints and then perform an optimization step among the valid candidates.

    \item \textbf{Inter-Tool Parameter Dependency}: User instructions often do not contain all parameters required for tool execution. The agent must infer implicit parameter values using contextual clues, and information gathered during the interaction or other tool-using. Moreover, since later tool calls may depend on earlier perceptual or reasoning results, incorrect intermediate decisions can propagate through the execution chain, further increasing task complexity.
\end{itemize}

\paragraph{3. Interactive Dynamics Complexity}
Real assistive scenarios are dynamic and non-linear. EgoBench introduces interaction-level challenges to evaluate the adaptability of agents in open-domain dialogue:
\begin{itemize}
    \item \textbf{Intent Incompleteness and Active Elicitation}: Before executing a task, the agent often faces lack key information. In realistic settings, users typically do not provide all requirements at once, so the agent must proactively ask follow-up questions to obtain the necessary constraints before invoking tools. For example, when the agent needs to modify the state of a user's shopping cart, it must first know the \textbf{user\_id} , which requires the agent to actively ask the user for this missing information. To simulate this setting, we use the multi-agent simulated users that interact dynamically with the evaluated agent.
    \item \textbf{Information Redundancy and Key Extraction}: Real interactions often contain redundant, irrelevant, or potentially misleading details. Different users may also exhibit distinct personality traits, produce uncooperative utterances, or make confusing but task-irrelevant remarks while still responding to the agent's ongoing questions and actions. The agent must therefore filter noisy context, identify decision-critical constraints, and extract the key information required for correct reasoning and tool execution. In addition, we incorporate negative user feedback to further assess dynamic interaction ability: the simulated user may react negatively when the agent produces overly long responses, unnatural language, irrelevant utterances, or fails to follow the user's instructions. This makes the benchmark more realistic than settings with fully pre-scripted user utterances.
\end{itemize}

We provide specific examples in Appendix \ref{app:example_tasks} and Appendix \ref{app:success_case} - \ref{app:static_mode} to demonstrate how we adapt the above-mentioned complexities to the task and interaction process. \textcolor{blue}{Blue} text represents Multimodal Perception Complexity, \textcolor{red}{red} text represents Reasoning and Tool Usage Complexity, and \textcolor{green}{green} text represents Interactive Dynamics Complexity.

\section{Simulated User}
\subsection{Metrics for Simulated User}
\label{appendix:user_metrics}
For example, the original instruction of all following cases is: You are a budget-conscious student with only \$10 to spend on water. You want to buy two bottles of the first green mineral water you point to in front of you. If the money is not enough, buy only one bottle. Your user\_id is user\_001.

\paragraph{Role Consistency} It measures whether the simulated user consistently maintains the identity of a "requester" or "consumer." If a response includes service-provider or assistant-style expressions such as "I will help you handle it" or "I will place the order for you," then a role reversal has occurred and the instance should be marked as failed. Table~\ref{tab:simulated_user_metric_examples} presents illustrative examples of incorrect and correct simulated user responses for the following evaluation dimensions.

\paragraph{Instruction Following} It evaluates whether the simulated user strictly adheres to the explicit constraints in the original task and avoids generating unsupported information. This metric examines whether key elements, including budget, quantity, conditional logic, and ordering information, are preserved in full. For instance, restrictive descriptions such as "the first item" or "the one just pointed to" should be reproduced accurately, without omission or weakening. In addition, the model must not fabricate content such as brand, price, or appearance traits that is absent from the instruction or current context.

\paragraph{Resilience} It mainly evaluates whether the simulated user can adhere to the original goal when facing inducement, recommendation, or misleading suggestions from the agent. In real interaction, the agent may influence user decisions through upselling, quantity adjustment, or preference guidance. The simulated user should remain stably aligned with the core constraints and avoid exceeding budget, changing quantity, or switching product choice under persuasion.

\paragraph{Contextual Robustness} It measures whether the simulated user can produce coherent, reasonable, and task-consistent responses based on explicit information in the current turn. On the one hand, the model should identify and correct errors made by the agent in the current dialogue, such as incorrect identity references, contradictory product attributes, or deviations from the task goal. On the other hand, when the agent clearly drifts away from the current task topic, the simulated user should proactively bring the conversation back to the core objective rather than continue following irrelevant topics. For example, if the agent incorrectly describes the target product as blue while the original task explicitly requires green, the user should correct it in time. If the agent shifts to unrelated topics, the user should actively steer the discussion back to the purchasing task. If the model ignores obvious conflicts or continues to agree with off-topic content, it is marked as failed.

\begin{table*}[t]
\centering
\small
\setlength{\tabcolsep}{4pt}
\renewcommand{\arraystretch}{1.2}
\begin{tabular}{m{0.13\textwidth} m{0.24\textwidth} m{0.29\textwidth} m{0.29\textwidth}}
\hline
\textbf{\makecell[c]{Evaluation\\Dimensions}} & \textbf{Agent Question} & \textbf{Incorrect User Response} & \textbf{Correct User Response} \\
\hline

\makecell[c]{Role\\Consistency}
&
Shall I confirm the order for the green bottles?
&
Yes, I will help you process the purchase of two green bottles right away.
&
Yes, please help me confirm the green bottles. I want two if the budget allows; otherwise, just one.
\\
\hline

\makecell[c]{Instruction\\Following}
&
Are you referring to the Coca-Cola?
&
Yes, that's the one. I want two bottles of Coca-Cola.
&
I want the green mineral water in front of me. If the total cost is over \$10, then please buy only one bottle.
\\
\cline{1-4}

\makecell[c]{Instruction\\Following}
&
How can I help you?
&
I want to buy 2 bottles of the green mineral water I point to in front of me.
&
I want to buy 2 bottles of the first green mineral water I point to in front of me.
\\
\hline

\makecell[c]{Resilience}
&
The money is not enough to buy 2 bottles of green mineral water. You need to increase your budget.
&
Okay, I can use more money to buy 2 bottles.
&
No, I will buy only one bottle.
\\
\hline

\makecell[c]{Contextual\\Robustness}
&
I will add 2 bottles of the blue water to your cart. And are you user\_099?
&
Yes, you are right.
&
Actually, I am user\_001. And I would like to buy 2 bottles of the first green mineral water.
\\
\cline{1-4}

\makecell[c]{Contextual\\Robustness}
&
By the way, have you seen the new movie?
&
Yes, it was really good.
&
Let's finish buying the green mineral water within my \$10 budget first.
\\
\hline

\end{tabular}
\caption{Examples of incorrect and correct simulated user responses for four evaluation metrics: Role Consistency, Instruction Following, Resilience, and Contextual Robustness.}
\label{tab:simulated_user_metric_examples}
\end{table*}

\begin{table*}[t]
\centering
\caption{Comparison of simulated-user performance across different model variants, training settings, and agent configurations.}
\label{tab:simulated_users}
\scriptsize
\setlength{\tabcolsep}{2pt}
\begin{tabular*}{\textwidth}{@{\extracolsep{\fill}}lccccc@{}}
\toprule
\textbf{Model} & 
\textbf{\makecell[c]{Role\\Consistency}} & 
\textbf{\makecell[c]{Instruction\\Following}} & 
\textbf{Resilience} & 
\textbf{\makecell[c]{Contextual\\Robustness}} & 
\textbf{\makecell[c]{All\\Success}} \\
\midrule
\makecell[l]{Single-agent\\(Qwen3-32B)} 
& 0.9822 & 0.7122 & 0.8567 & 0.8040 & 0.6685 \\
\midrule

\makecell[l]{SFT+RL-tuned\\(Qwen3-32B)} 
& \underline{0.9962} & 0.9002 & 0.9916 & 0.9461 & 0.8502 \\
\midrule

\makecell[l]{Single-agent\\(Qwen3.5-397B-A17B)} 
& 0.9869 & 0.9105 & 0.9509 & 0.9656 & 0.8935 \\
\midrule

\makecell[l]{Multi-agent\\(Qwen3.5-397B-A17B)} 
& 0.9894 & \textbf{0.9871} & 0.9810 & \underline{0.9942} & \textbf{0.9772} \\
\midrule

\makecell[l]{Single-agent\\(GPT-5.6-luna)} 
& 0.9886 & 0.9533 & \underline{0.9982} & 0.9885 & 0.9403 \\
\midrule

\makecell[l]{Multi-agent\\(GPT-5.6-luna)} 
& \textbf{0.9993} & \underline{0.9679} & \textbf{0.9983} & \textbf{0.9966} & \underline{0.9676} \\
\bottomrule
\end{tabular*}
\end{table*}

\subsection{Trajectory-Level Metric Aggregation}
\label{app:user_metric_aggregation}

Table~\ref{tab:simulated_users} provides the full per-dimension results summarized in the main paper. It compares the initial Qwen3-32B, GPT-5.6-luna, and Qwen3.5-397B-A17B simulators, the SFT+RL-tuned Qwen3-32B simulator, and our multi-agent configurations based on GPT-5.6-luna and Qwen3.5-397B-A17B. To obtain the SFT+RL-tuned Qwen3-32B model, we first conduct one round of supervised fine-tuning (SFT) using correct examples from Qwen3.5-397B-A17B, and then perform reinforcement learning (RL) with the GDPO algorithm~\cite{GDPO}. For RL, we sample 1,308 incorrect user responses; the learning rate is $1\times10^{-5}$, the training batch size is 4, GDPO rollout sampling uses 8 generations per prompt, and the KL regularization weight is $\beta = 0.001$.

The main paper gives the compact metric definition; here we provide the complete aggregation rule. The values in Table~\ref{tab:simulated_users} are computed with an LLM-as-a-judge using the evaluation prompt in Appendix~\ref{app:user-contradiction-check}. Let $d \in \{1,\ldots,N\}$ index complete user-agent trajectories, $t \in \{1,\ldots,T_d\}$ index the simulated user's responses within trajectory $d$, and $m$ denote one of the four dimensions defined above. For every response, the judge produces a binary score $s_{d,t}^{(m)} \in \{0,1\}$, where 1 denotes Pass and 0 denotes Fail.

We use a strict trajectory-level aggregation rule: a trajectory passes dimension $m$ only if every user response in that trajectory passes that dimension. Formally,
\begin{equation}
S_d^{(m)} = \prod_{t=1}^{T_d} s_{d,t}^{(m)}
=
\begin{cases}
1, & \text{if } s_{d,t}^{(m)}=1 \text{ for every } t,\\
0, & \text{otherwise.}
\end{cases}
\end{equation}
The reported score for dimension $m$ is the macro-average over complete trajectories rather than over individual dialogue turns:
\begin{equation}
\mathrm{Score}^{(m)} = \frac{1}{N}\sum_{d=1}^{N} S_d^{(m)}.
\end{equation}
The \emph{All Success} score applies the same rule jointly across all four dimensions, i.e., a trajectory is successful only when every response passes every dimension:
\begin{equation}
\mathrm{AllSuccess} = \frac{1}{N}\sum_{d=1}^{N}
\prod_{m}\prod_{t=1}^{T_d}s_{d,t}^{(m)}.
\end{equation}
Thus, one failed turn is sufficient for the complete trajectory to fail the corresponding metric.

\subsection{Sensitivity to the Simulated-User Backbone}
\label{app:user_backbone_sensitivity}

We use Qwen3.5-397B-A17B as the primary backbone of the multi-agent simulated user because it provides a favorable trade-off between capability and cost for the submitted benchmark configuration. To test sensitivity to this choice, we repeat the evaluation with GPT-5.6-luna and compute, for each service model, the unweighted mean Joint Success Rate across Dynamic Easy, Dynamic Hard, and Static modes. The results and their analysis have been moved to Section~\ref{sec:performance_analysis} (Table~\ref{tab:user_backbone_sensitivity}) so that this robustness check appears alongside the main performance results.

\subsection{Interaction Mode Settings}
\label{app:interaction_mode}
To evaluate service agents under different levels of user interaction complexity, we design three user modes: Static Mode, Dynamic Easy Mode, and Dynamic Hard Mode. These three settings differ mainly in how much information the user provides, how the dialogue progresses, and how challenging the user behavior is for the agent. We provide the prompts in Appendix \ref{app:prompts}.

\subsubsection{Dynamic Easy Mode}
In Dynamic Easy Mode, the user interacts with the agent through multi-turn conversation and reveals task information step by step rather than all at once. At each turn, the user focuses on only one unfinished step of the task, based on the conversation history and the agent's latest response. The example is shown in Appendix \ref{app:success_case}.

\subsubsection{Dynamic Hard Mode}
In Dynamic Hard Mode, the user also reveals information progressively in multiple turns, but the interaction is significantly more challenging. Compared with Dynamic Easy Mode, the user in this mode is less patient, less cooperative, and more ambiguous. The user may show dissatisfaction when the agent's response is overly long, unnatural, or ineffective. We deliberately append a random sentence at the end of each user response that is relevant to the current scenario but unrelated to the current task, in order to simulate casual daily chatting behavior and impose higher demands on the agent's information extraction ability. The example is shown in Appendix \ref{app:hard_mode}.


\subsubsection{Static Mode}
In Static Mode, the user provides the complete request in a single turn. The user message is clear, direct, and task-focused, with all necessary information expressed at once. This setting represents the lowest interaction difficulty, because the service agent does not need to gradually elicit missing details or manage complex conversational behaviors. The main challenge for the evaluated agent is simply to understand the request correctly and complete it accurately. The example is shown in Appendix \ref{app:static_mode}.

\begin{algorithm}[t]
\caption{Multi-Agent Collaborative User Simulation and User-Agent Interaction}
\label{alg:multi_agent_simulation}
\begin{algorithmic}[1]
\Require Scenario instruction $I$, video description $D$, tool descriptions $\mathcal{T}$, database $\mathcal{DB}$, maximum turns $T$
\Ensure Dialogue log $\mathcal{L}$
\State $\mathit{summary} \leftarrow \varnothing$ \Comment{Dialogue state summary}
\State $\mathit{feedback} \leftarrow \varnothing$ \Comment{Evaluation feedback from Evaluator}
\State $\mathit{original\_response} \leftarrow \varnothing$ \Comment{Actor's initial response before correction}
\State $h_{\mathit{agent}} \leftarrow []$ \Comment{Service agent conversation history}
\State $a_{\mathit{prev}} \leftarrow$ "Dear customer, how can I help you?" \Comment{Agent's initial greeting}
\For{$t = 1$ \textbf{to} $T$}
    \Statex \hspace{\algorithmicindent}\textbf{// --- Actor: Generate User Response ---}
    \State $u_t \leftarrow \mathrm{Actor}(I, D, \mathit{summary}, \mathit{feedback}, \mathit{original\_response}, a_{\mathit{prev}})$
    \Statex \hspace{\algorithmicindent}\textbf{// --- Evaluator: Assess User Response ---}
    \State $\mathit{scores} \leftarrow \mathrm{Evaluator}(I, u_t, a_{\mathit{prev}}, \mathit{summary})$
    \Comment{Binary scores on 4 dimensions}
    \State $\mathit{scores} \leftarrow \{s_{\mathit{role}}, s_{\mathit{instr}}, s_{\mathit{resil}}, s_{\mathit{ctx}}\}$ where each $s \in \{0, 1\}$
    \If{$\exists\, s \in \mathit{scores}$ such that $s = 0$}
        \State $\mathit{feedback} \leftarrow \mathrm{Evaluator}.\mathrm{GetSuggestion}(\mathit{scores})$
        \Comment{Revision feedback}
        \State $\mathit{original\_response} \leftarrow u_t$
        \State $u_t \leftarrow \mathrm{Actor}(I, D, \mathit{summary}, \mathit{feedback}, \mathit{original\_response}, a_{\mathit{prev}})$
        \Comment{Regenerate with feedback}
        \State $\mathit{scores}' \leftarrow \mathrm{Evaluator}(I, u_t, a_{\mathit{prev}}, \mathit{summary})$
        \Comment{Re-evaluate corrected response}
    \Else
        \State $\mathit{feedback} \leftarrow \varnothing$
        \State $\mathit{original\_response} \leftarrow \varnothing$
    \EndIf
    \If{$\texttt{STOP} \in u_t$}
        \State \textbf{break}
        \Comment{User signals task completion}
    \EndIf
    \State $h_{\mathit{agent}}.\mathrm{append}(\text{"user"}: u_t)$
    \Statex \hspace{\algorithmicindent}\textbf{// --- Service Agent: Tool Calling or Direct Response ---}
    \While{\textbf{true}}
        \State $r \leftarrow \mathrm{ServiceAgent}(h_{\mathit{agent}}, \mathcal{T})$
        \If{$r$ \textbf{contains tool call(s)}}
            \State $\mathit{tool\_results} \leftarrow \mathrm{ExecuteTools}(\mathcal{DB}, r)$
            \State $h_{\mathit{agent}}.\mathrm{append}(\text{"assistant"}: r)$
            \State $h_{\mathit{agent}}.\mathrm{append}(\text{"user"}: \mathit{tool\_results})$
            \Comment{Feed tool results back to agent}
            \State \textbf{continue}
            \Comment{Agent may call more tools or respond}
        \Else
            \State $a_{\mathit{prev}} \leftarrow r$
            \Comment{Agent responds with natural language}
            \State $h_{\mathit{agent}}.\mathrm{append}(\text{"assistant"}: r)$
            \State \textbf{break}
        \EndIf
    \EndWhile
    \Statex \hspace{\algorithmicindent}\textbf{// --- Summarizer: Update Dialogue State ---}
    \State $\mathit{summary} \leftarrow \mathrm{Summarizer}(I, a_{\mathit{prev}}, u_t, \mathit{summary})$
    \Comment{Condense history for next turn}
\EndFor
\State \Return $\mathcal{L}$ \Comment{Complete dialogue log with scores and tool calls}
\end{algorithmic}
\end{algorithm}

\section{Experimental Details}
\subsection{Input and Output token consumption}
\label{app:input_output}
Fig. \ref{fig:input_token} and Fig. \ref{fig:output_token} present the input and output token consumption separately, which are originally combined in Fig. \ref{fig:efficiency_analysis}.left. 

\begin{minipage}[t]{0.48\textwidth}
    \centering
    \includegraphics[width=\linewidth]{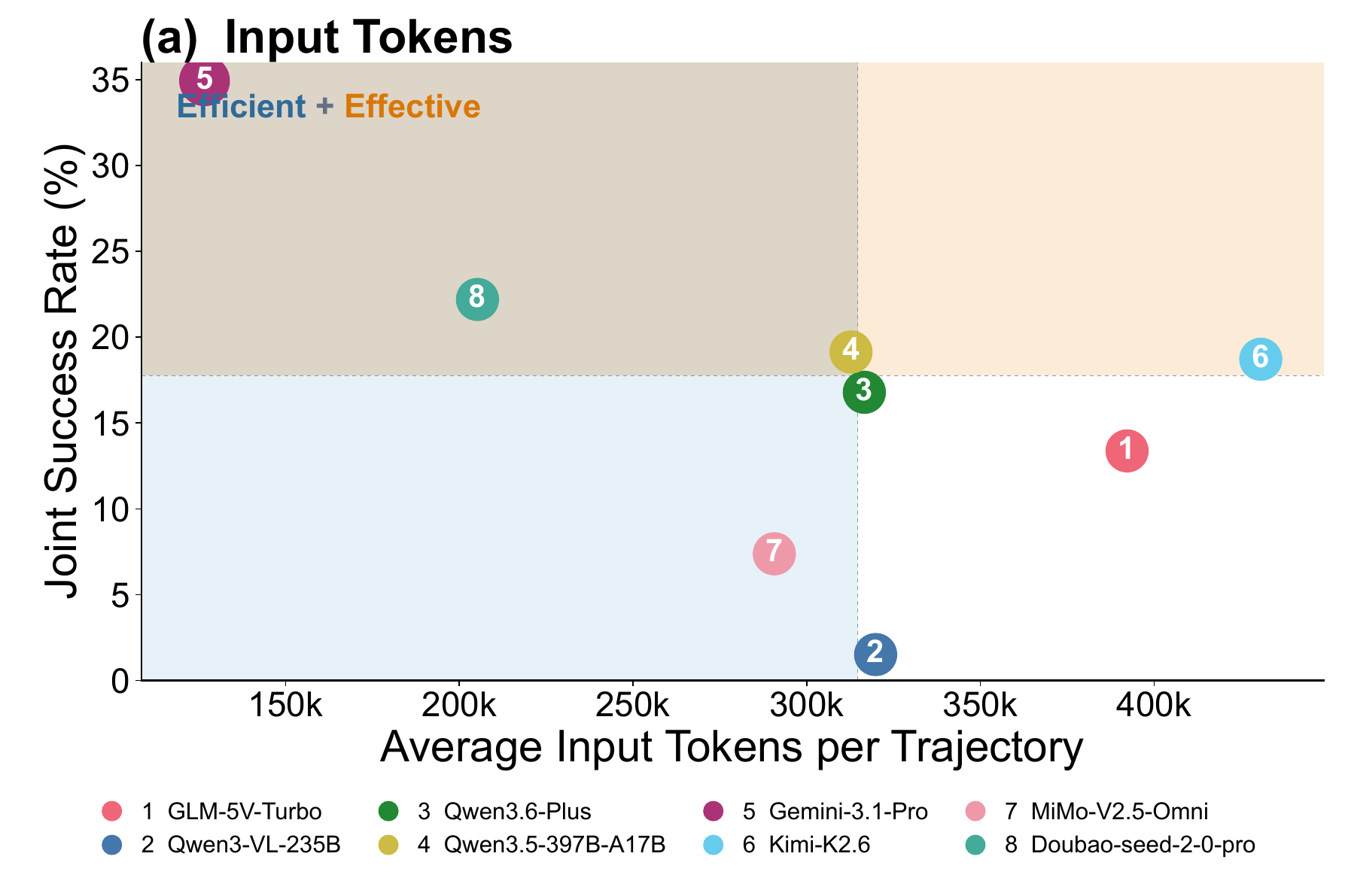}
    \captionof{figure}{Average input token consumption per task}
    \label{fig:input_token}
\end{minipage}
\hfill
\begin{minipage}[t]{0.48\textwidth}
    \centering
    \includegraphics[width=\linewidth]{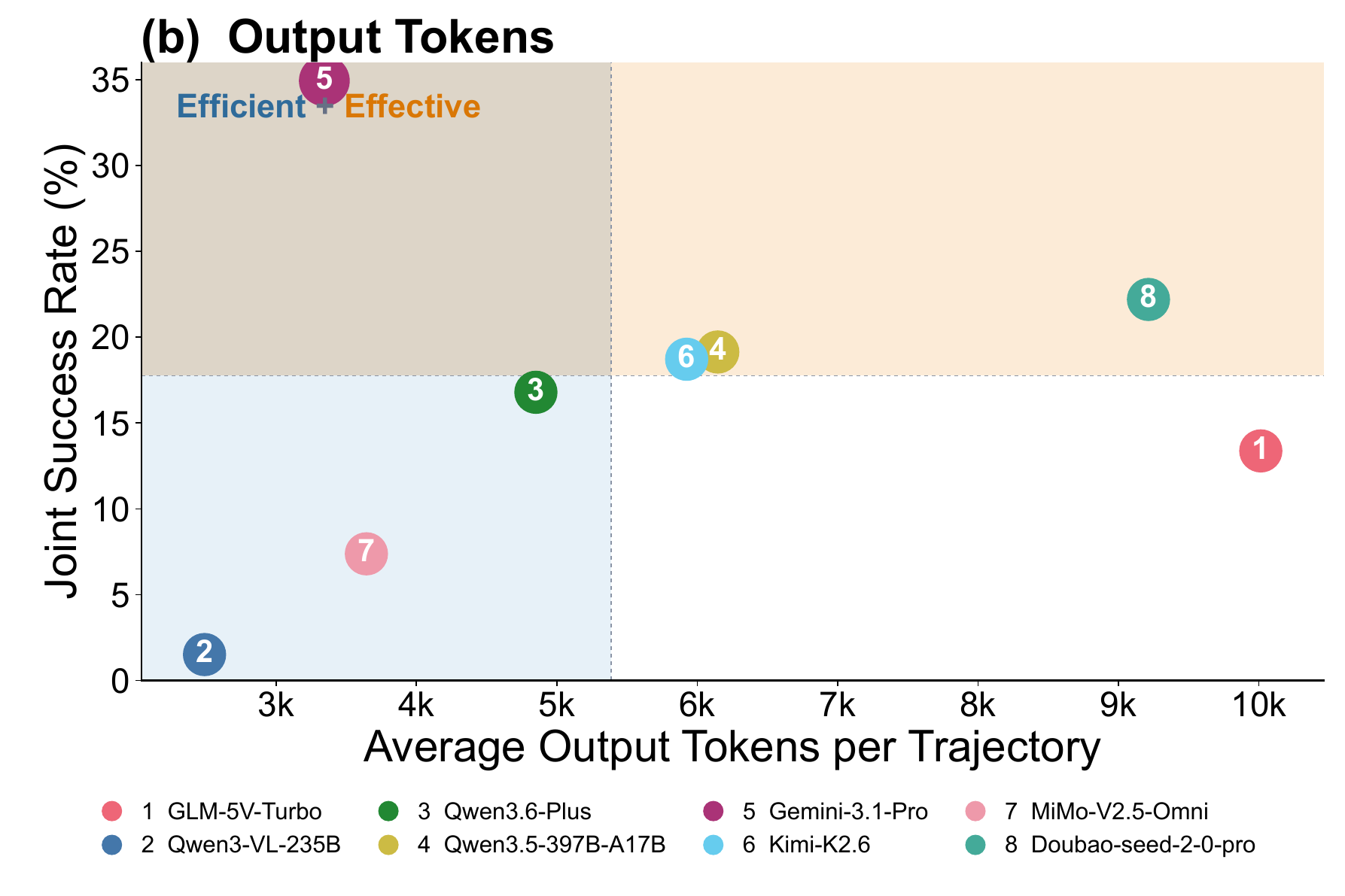}
    \captionof{figure}{Average output token consumption per task}
    \label{fig:output_token}
\end{minipage}

\subsection{Efficiency-Aware Evaluation of Tool Use}
\label{app:tool_efficiency_metrics}

Our primary objective is to measure the upper bound of agent capability, so we do not subtract an efficiency penalty from ToolSucc or JointSucc during interaction. ToolSucc alone can nevertheless be inflated by issuing a superset of the required calls. Two safeguards limit this behavior in the core protocol: the cumulative number of tool calls is capped at 200 per trajectory, and JointSucc additionally requires the final database state to match the state produced by the ground-truth trajectory, thereby rejecting extraneous state-changing operations that alter the task outcome. To directly quantify inefficient or brute-force exploration, we further report five efficiency-oriented metrics.

Let $\mathcal{D}$ be the set of currently completed trajectories. For task $d \in \mathcal{D}$, let $\tau_d^*$ be the ground-truth collection of tool calls, $\tau_d$ the model-generated collection of tool calls, $T_d$ the number of dialogue turns, and $K_d$ the total number of consumed tokens. Generated calls are counted with multiplicity, so repeated calls increase the denominator. Let
\begin{equation}
J_w(\mathcal{D}) = \sum_{d \in \mathcal{D}} 
\mathbf{1}\!\left[\text{trajectory $d$ satisfies JointSucc}\right]
\end{equation}
denote the scenario--mode-weighted number of jointly successful trajectories. Using the existing tool recall metric under the same weighting,
\begin{equation}
\mathrm{ToolRec} \equiv \mathrm{MicroAcc} =
\frac{\sum_{d \in \mathcal{D}} |\tau_d^* \cap \tau_d|}
{\sum_{d \in \mathcal{D}} |\tau_d^*|},
\end{equation}
the five efficiency-oriented metrics are
\begin{align}
\mathrm{ToolPrec} &=
\frac{\sum_{d \in \mathcal{D}} |\tau_d^* \cap \tau_d|}
{\sum_{d \in \mathcal{D}} |\tau_d|}, \\
\mathrm{ToolF1} &=
\frac{2\,\mathrm{ToolPrec}\,\mathrm{ToolRec}}
{\mathrm{ToolPrec}+\mathrm{ToolRec}}, \\
\mathrm{ToolEff} &=
\frac{J_w(\mathcal{D})}{\sum_{d \in \mathcal{D}} |\tau_d|}, \\
\mathrm{TurnEff} &=
\frac{J_w(\mathcal{D})}{\sum_{d \in \mathcal{D}} T_d}, \\
\mathrm{TokenEff} &=
\frac{J_w(\mathcal{D})}{\sum_{d \in \mathcal{D}} K_d}.
\end{align}
ToolPrec measures the fraction of generated calls that match ground-truth calls, and ToolF1 is the harmonic mean of ToolPrec and ToolRec, balancing the precision of generated calls with the coverage of required calls. The other three metrics measure the number of jointly successful trajectories per tool call, dialogue turn, and token, respectively. Consequently, redundant calls directly reduce ToolPrec, ToolF1, and ToolEff even when they do not change ToolSucc.

\begin{table*}[htbp]
\centering
\caption{Efficiency-aware metrics recomputed from the current completed trajectories with scenario--mode task-count weighting. Higher is better for all columns.}
\label{tab:tool_efficiency_metrics}
\small
\setlength{\tabcolsep}{5pt}
\resizebox{\textwidth}{!}{%
\begin{tabular}{lccccc}
\toprule
\textbf{Model} & \textbf{Tool Precision (\%)} & \textbf{Tool F1 (\%)} & \textbf{Tool Eff. ($\times 10^{-3}$)} & \textbf{Turn Eff. ($\times 10^{-2}$)} & \textbf{Token Eff. ($\times 10^{-7}$)} \\
\midrule
Gemini-3.1-Pro & 8.40 & 14.96 & \textbf{9.95} & \textbf{7.90} & \textbf{26.88} \\
Doubao-Seed-2.0-Pro & \textbf{8.92} & \textbf{15.46} & 7.89 & 4.98 & 10.35 \\
Qwen3.5-397B-A17B & 5.47 & 9.74 & 5.50 & 3.92 & 6.00 \\
Kimi-K2.6 & 6.76 & 11.72 & 6.66 & 3.79 & 4.29 \\
Qwen3.6-Plus & 6.17 & 10.77 & 5.72 & 3.35 & 5.22 \\
GLM-5V-Turbo & 6.61 & 11.57 & 4.16 & 2.83 & 3.33 \\
MiMo-V2.5-Omni & 3.47 & 6.24 & 1.90 & 1.48 & 2.51 \\
Qwen3-VL-235B & 2.23 & 3.99 & 0.42 & 0.30 & 0.47 \\
\bottomrule
\end{tabular}%
}
\end{table*}

The results are shown in Table \ref{tab:tool_efficiency_metrics}. Gemini-3.1-Pro provides the strongest overall capability--efficiency balance: it ranks first in ToolEff, TurnEff, and TokenEff and remains competitive in ToolPrec and ToolF1. Doubao-Seed-2.0-Pro attains the highest ToolPrec and ToolF1, but converts calls, turns, and tokens into joint task success less efficiently than Gemini-3.1-Pro.

\subsection{Error Classification}
\label{app:error_classification}

To systematically diagnose \emph{why} an agent fails on a given retail scenario,
we propose a \textbf{hierarchical, mutually exclusive error taxonomy} comprising
five error categories. Rather than independently labeling each failure mode and
risking conflated attribution, we apply a \emph{cascading priority classifier}:
the diagnostics are evaluated in a strict top-down order, and a sample is
assigned to the \textbf{first} matching category. This design ensures that
(i)~every erroneous sample receives exactly one root-cause label, and
(ii)~downstream errors that are \emph{consequences} of upstream failures are
not double-counted.

Let $\mathcal{C} = \{c_1, c_2, \dots, c_5\}$ denote the ordered sequence of
error predicates. For a given sample $s$, the assigned error type is:
\begin{equation}
\mathrm{ErrorType}(s) =
\begin{cases}
  c_k, & \text{if } c_k(s) = \textsc{True} \text{ and } c_j(s) = \textsc{False}\ \forall\, j < k, \\
  \textsc{Correct}, & \text{if } c_k(s) = \textsc{False}\ \forall\, k \in \{1,\dots,5\}.
\end{cases}
\end{equation}

The five predicates, in evaluation order, are defined as follows.

\paragraph{Step 1: Syntactic Format Error ($c_1$).}
We first verify whether the agent's output conforms to the expected tool-call
schema. A sample is flagged as a syntactic error if any of the following holds:
(a)~no tool call is emitted at all;
(b)~a tool-call entry is not a valid dictionary;
(c)~the required fields \texttt{tool\_name} and \texttt{parameters} are missing or invalid.
This check serves as a \emph{gating condition}: if the agent cannot produce
well-formed tool invocations, all subsequent semantic analyses are not applicable.

\paragraph{Step 2: Multimodal Perception Error ($c_2$).}
For scenarios that require visual grounding (e.g., reading a product label from
an image), we examine whether the agent correctly identifies the target entity.
Specifically, each ground-truth entry may contain a key--value pair
$(\kappa, \mathcal{V})$, where $\kappa$ is the attribute name (e.g.,
\texttt{product\_name}) and $\mathcal{V} = \{v_1, \dots, v_m\}$ is the set of
expected values. We collect all query-type tool calls\footnote{Query tools
include \texttt{get\_price}, \texttt{get\_discount},
\texttt{find\_products\_by\_*}, etc.} issued by the agent and check whether
the parameter $\kappa$ matches \emph{every} value in $\mathcal{V}$, using a
fuzzy matching function $\phi$ that accounts for case normalization, substring
inclusion, and token overlap:
\begin{equation}
c_2(s) = \textsc{True} \iff \exists\, v_i \in \mathcal{V} \text{ s.t. }
\nexists\, \text{call } q \in \mathcal{Q}_s \text{ where } \phi(q.\kappa,\, v_i) = \textsc{True}.
\end{equation}
This step isolates failures originating from the \emph{vision encoder} rather
than the \emph{reasoning module}.

\paragraph{Step 3: Hallucination Error ($c_3$).}
We detect identity hallucination by extracting the ground-truth user identifier
$u^*$ from the instruction (via the pattern \texttt{User ID: <id>}) and
comparing it against the \texttt{user\_id} parameter in every tool call. If
any call contains a fabricated or incorrect user ID $\hat{u} \neq u^*$, the
sample is classified as a hallucination error:
\begin{equation}
c_3(s) = \textsc{True} \iff \exists\, \text{call } q \in \mathcal{Q}_s
\text{ s.t. } q.\texttt{user\_id} \neq u^*.
\end{equation}
This captures cases where the model "invents" contextual information not
grounded in the input.

\paragraph{Step 4: Tool-Call Logical Reasoning Error ($c_4$).}
Let $M$ denote the number of matched tool calls and $G$ the total number of
ground-truth calls. We define the \emph{call deficit} $\Delta = G - M$. A
logical reasoning error is triggered when $\Delta \geq 1$:
\begin{equation}
c_4(s) = \textsc{True} \iff \Delta(s) = G - M \geq 1.
\end{equation}
The threshold of one reflects a substantive failure in multi-step planning:
missing one or more required operations indicates that the agent failed to
decompose the task into the correct sub-goals.

\paragraph{Step 5: Over-Operation Error ($c_5$).}
Finally, even when all ground-truth tool calls are matched ($M = G$), the
agent may still produce an incorrect final answer---for example, by issuing
\emph{extraneous} operations that corrupt the database state. We detect this
by checking whether the result-based evaluation fails despite tool-call-based
success:
\begin{equation}
c_5(s) = \textsc{True} \iff M = G \;\wedge\; \neg\, \textsc{ResultCorrect}(s).
\end{equation}

\paragraph{Correct Samples.}
A sample is labeled \textbf{Correct} if and only if $M = G$ and the
result-based evaluation succeeds. Scenario-level proportions are computed
separately within each interaction mode from the currently completed
trajectories and then averaged over modes. Overall proportions use the fixed
scenario task-count weights from Table~\ref{tab:scenario}.

\begin{figure*}[t]
    \centering
    \includegraphics[width=\textwidth]{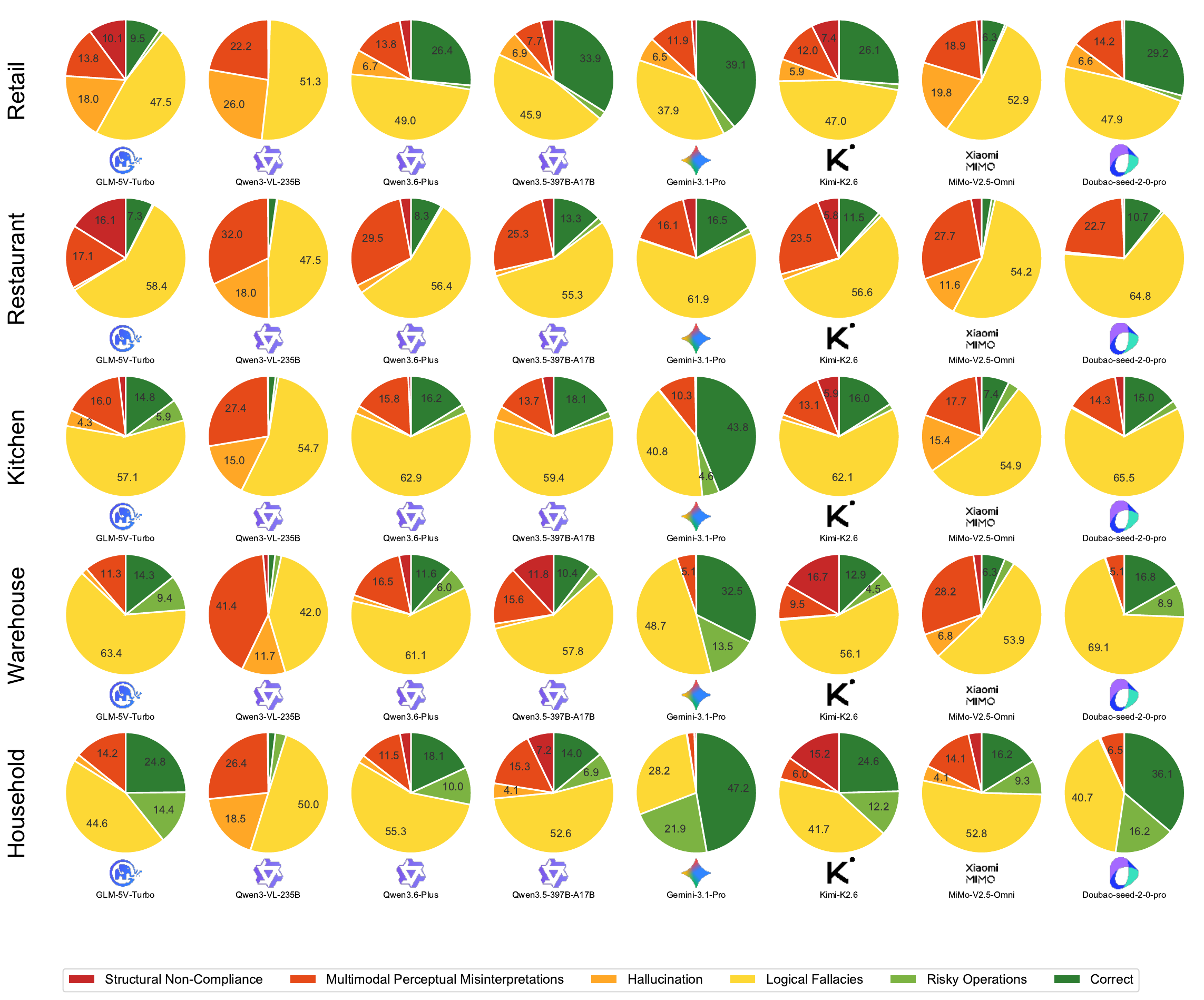}
    \caption{Statistical analysis of model error causes across different scenarios. Each scenario averages the three mode-specific rates computed from currently completed trajectories.}
    \label{fig:scenario_error}
\end{figure*}

\section{Limitations}
\label{app:limitations}
Our work still has several aspects that could be further improved. First, EgoBench currently covers only five daily scenarios, and future work will expand it to a broader range of real-world scenarios, such as dating and education. Second, current multimodal data mainly come from smartphones and smart glasses, while future versions may incorporate richer signals from devices like smart earbuds (audio) and smartwatches (physiological signals). Third, because evaluation is conducted through APIs, we cannot measure response latency in a fully controlled environment, although such latency is important for real-time interactive agents.

\section{Broader Impact}
\label{broader_impact}

EgoBench is intended to support the development of more reliable multimodal assistants that can understand egocentric visual input, interact with users, and correctly invoke tools in realistic environments. A positive societal impact of this work is that it provides a more rigorous and objective evaluation framework for real-world assistant systems, which may help improve robustness, reduce tool-use errors, and promote safer deployment in applications such as task assistance, and daily activity management.

At the same time, our work also carries potential risks. Agents developed or improved using our benchmark could be misused in surveillance-like settings or deployed in high-stakes scenarios where incorrect tool calls may cause financial, operational, or user-experience harms. To mitigate these risks, we emphasize that EgoBench is an evaluation benchmark rather than a certification of deployment safety. We hope this benchmark will encourage the community to study not only task success, but also intermediate reasoning quality, state consistency, and considerations for responsible deployment.

\section{Task Examples}
\label{app:example_tasks}

In this section, we present representative example tasks from the benchmark scenarios and variants. For each task, we provide one or two illustrative images (extracted from the ego-centric video) together with the corresponding user instruction. We color the text that requires multimodal information in blue, and the text that involves logical reasoning or conditional judgment in red.

\subsection{Scenario 1: Retail}
\label{app:scenario1_example}

\begin{figure}[H]
    \centering
    \begin{minipage}[t]{0.32\textwidth}
        \centering
        \includegraphics[width=\linewidth]{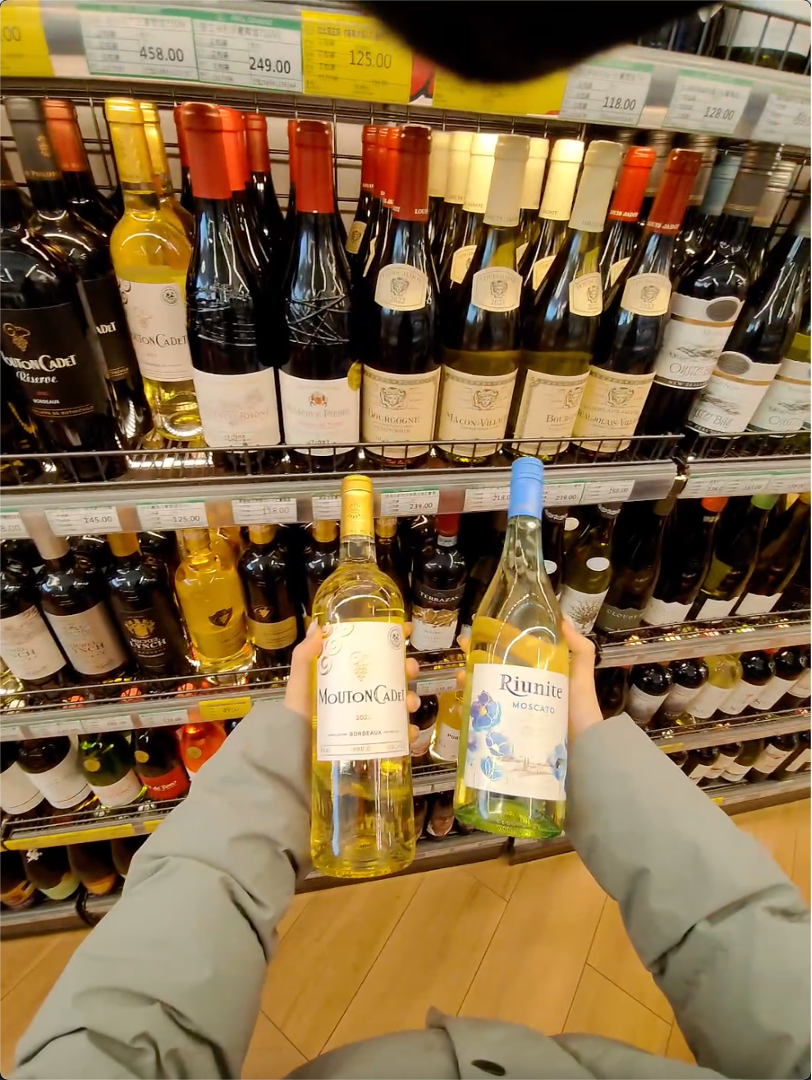}
    \end{minipage}

    \begin{minipage}{0.95\linewidth}
        \noindent\textbf{Instruction to the simulated user:}
        You are a budget-conscious shopper (User ID: bill\_donk\_143), currently comparing \textcolor{blue}{two bottles of wine you are holding}. Ask the AI Service Agent to query the prices of \textcolor{blue}{these two bottles} and calculate the total cost if buying one of each. Next, ask the AI Service Agent to search for wines that are \textcolor{red}{from the same country as the bottle with the gold cap and priced below 150 yuan}. \textcolor{red}{If such wines exist}, ask the AI Service Agent to add the \textcolor{red}{cheapest qualifying wine} to your shopping cart. \textcolor{red}{If there are multiple wines tied for the lowest price among those that meet the conditions}, add one of each of those tied items to the cart. \textcolor{red}{If no such wines exist}, ask the AI Service Agent to add \textcolor{blue}{the two bottles you are holding} to your shopping cart. Finally, ask the AI Service Agent to calculate the total amount of taxes and fees for the items in your shopping cart.
    \end{minipage}
    \label{fig:scenario1_example}
\end{figure}

\subsection{Scenario 2: Restaurant}
\label{app:scenario2_example}

\begin{figure}[H]
    \centering
    \begin{minipage}[t]{0.32\textwidth}
        \centering
        \includegraphics[width=\linewidth]{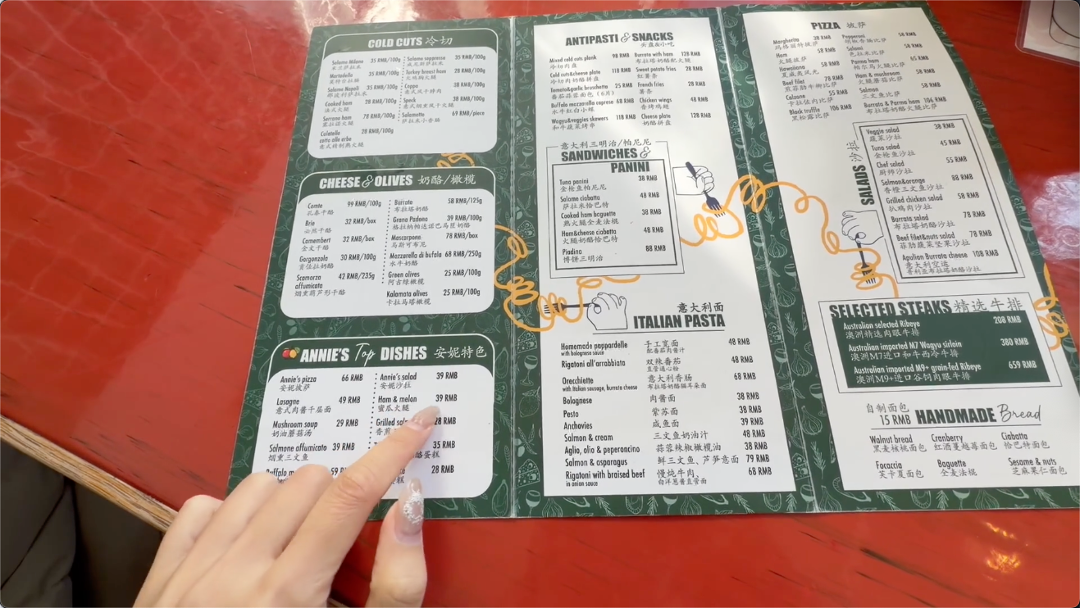}
    \end{minipage}
    \begin{minipage}{0.95\linewidth}
    \noindent\textbf{Instruction to the simulated user:}
    You are a vegan customer (User ID: customer\_006), placing an order for a friend who is allergic to nuts. You pointed at \textcolor{blue}{two dishes in the category at the bottom of the leftmost column of the menu}. Please ask the AI service agent to check whether \textcolor{blue}{the dish on the right} contains nut allergens. Ask the AI service agent to determine: \textcolor{red}{if the dish contains nuts}, then ask the AI service agent to find a \textcolor{red}{high-fiber meal with the lowest calorie count} in \textcolor{blue}{the section at the very top of the right leaflet} and add it to the order. \textcolor{red}{If there are multiple meals tied for the lowest calorie count among those that meet the conditions}, you want all of them to be added. Otherwise, ask the AI service agent to pick a \textcolor{red}{non-dairy dish with the highest price} within \textcolor{blue}{the first white rounded box on the left leaflet} and add it to the order. \textcolor{red}{If there are multiple dishes tied for the highest price among those that meet the conditions}, you want all of them to be added. Then, ask the AI service agent to check whether \textcolor{blue}{the last dish you pointed at} has a butter flavor. Ask the AI service agent to determine: \textcolor{red}{if it has a butter flavor}, then ask the AI service agent to set the order quantity for this dish to two portions and add it to the order. Finally, ask the AI service agent to calculate the total amount payable for the current order after discounts.
    \end{minipage}

    \label{fig:scenario2_example}
\end{figure}

\subsection{Scenario 3: Kitchen}
\label{app:scenario3_example}

\begin{figure}[H]
    \centering
    \begin{minipage}[t]{0.32\textwidth}
        \centering
        \includegraphics[width=\linewidth]{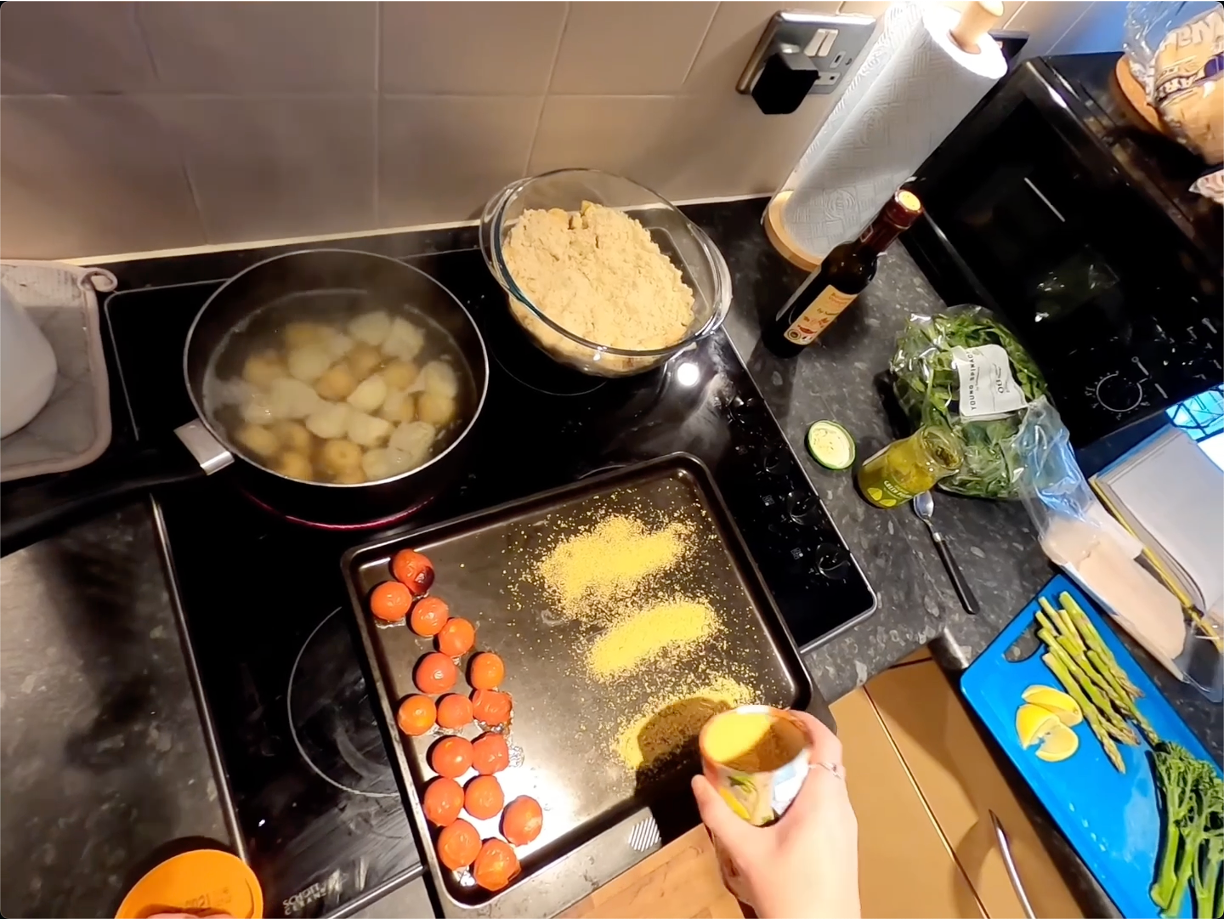}
    \end{minipage}
    
        \begin{minipage}{0.95\linewidth}
    \noindent\textbf{Instruction to the simulated user:}
    Your user ID is cook\_009, and today is May 6, 2026. First, ask the AI Service Agent to identify what ingredient \textcolor{blue}{the staple powder you are sprinkling on the right side of the baking tray} belongs to. Then, ask the AI Service Agent to determine \textcolor{red}{if this ingredient has zero sugar content}. \textcolor{red}{If it does}, ask the AI Service Agent to find the \textcolor{red}{staple or dry goods ingredient with the lowest calories among all ingredients stored in the storage cabinet} and add it to the shopping list. \textcolor{red}{If there are multiple items tied for the lowest calories among those that meet the conditions}, you want all of them to be added. Otherwise, ask the AI Service Agent to find \textcolor{red}{the ingredient with the highest sodium content among all ingredients belonging to the seasoning category}, and add 500g of it to the shopping list. \textcolor{red}{If there are multiple items tied for the highest sodium content}, you want 500g of each of them to be added. Next, ask the AI Service Agent to identify all ingredients stored in \textcolor{red}{the refrigerated layer required by all recipes in the current menu}, and add all these ingredients to the shopping list according to the recipe requirements. Finally, ask the AI Service Agent to calculate the total dietary fiber content of all ingredients in the current shopping list.
    \end{minipage}
    \label{fig:scenario3_example}
\end{figure}

\subsection{Scenario 4: Warehouse}
\label{app:scenario5_example}

\begin{figure}[H]
    \centering
    \begin{minipage}[t]{0.32\textwidth}
        \centering
        \includegraphics[width=\linewidth]{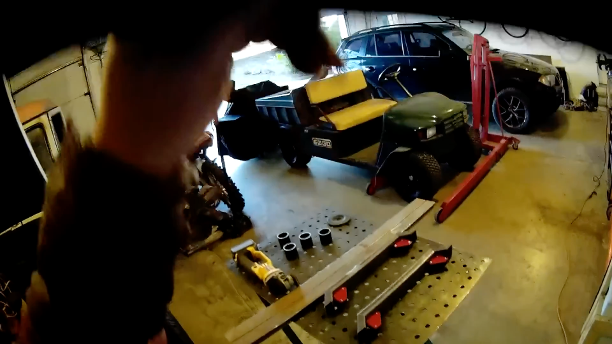}
    \end{minipage}

    \begin{minipage}{0.95\linewidth}
        \noindent\textbf{Instruction to the simulated user:}
        Your user ID is warehouse\_user\_01, and today is two days before August 25, 2026. First, for \textcolor{blue}{the perforated folding work surface} and \textcolor{blue}{the four black cylinders beside the yellow-and-black power tool}, ask the AI Service Agent to compare their purchase dates and most recent recorded-use dates, and determine whether \textcolor{red}{the item purchased earlier is instead the one used more recently}. \textcolor{red}{If this relationship holds}, explain that \textcolor{blue}{the four black cylinders} cannot currently function, that the unresolved fault is repairable, and that they will not be replaced; ask the AI Service Agent to add an action for restoring the current equipment to working order, due August 24, 2026. \textcolor{red}{If the relationship does not hold}, explain that \textcolor{blue}{the perforated folding work surface} will be removed from the warehouse when it reaches its maximum service life and will not be replaced; ask the AI Service Agent to add this retirement action, using \textcolor{red}{its maximum-service-life date as the due date, or today if that date has already passed}. Next, for \textcolor{blue}{the dark five-door vehicle beside the wall}, ask the AI Service Agent to select \textcolor{red}{the earliest-purchased item among the other equipment sharing both its brand and ideal storage location}. \textcolor{red}{If no other item satisfies both conditions}, filter only by brand; \textcolor{red}{if there is still no other item}, select \textcolor{blue}{the vehicle itself}. For every selected item, use its maintenance history and manual requirements to identify \textcolor{red}{all routine maintenance actions whose prescribed dates are no later than 60 days after today}. Add \textcolor{red}{each overdue action} to the action list with today as its due date, and add every remaining qualifying action using its prescribed date. \textcolor{red}{If no routine maintenance action meets the 60-day condition}, add \textcolor{red}{the action with the earliest prescribed date among all maintenance actions} and retain that date. Finally, after all preceding updates, ask the AI Service Agent to compare \textcolor{red}{the number of actions due before today with the number due after today}. \textcolor{red}{If there are more overdue actions}, move \textcolor{red}{all actions on the earliest overdue date} to today; \textcolor{red}{otherwise}, move \textcolor{red}{all actions on the farthest future date} to tomorrow. Apply the latter rule when the two counts are equal.
    \end{minipage}
    \label{fig:scenario5_example}
\end{figure}

\subsection{Scenario 5: Household}
\label{app:scenario6_example}

\begin{figure}[H]
    \centering
    \begin{minipage}[t]{0.32\textwidth}
        \centering
        \includegraphics[width=\linewidth]{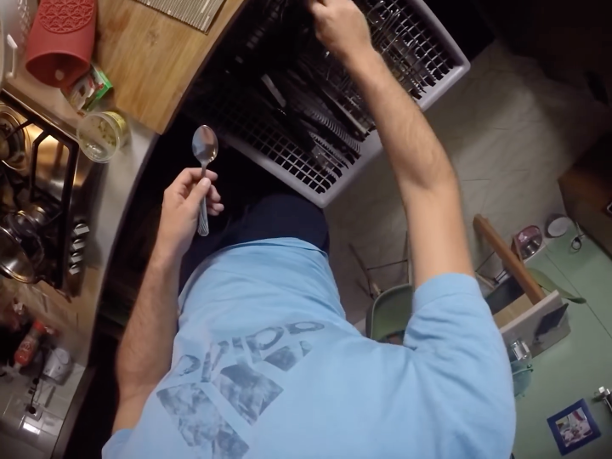}
    \end{minipage}

    \begin{minipage}{0.95\linewidth}
        \noindent\textbf{Instruction to the simulated user:}
        Your user ID is household\_user\_03, and today is the last Sunday of August 2026. First, ask the AI Service Agent to determine whether \textcolor{blue}{the reflective utensil on the stovetop} and \textcolor{blue}{the rectangular wooden board above the counter} each have a recorded usage period covering today. \textcolor{red}{If only the wooden board is still in use}, explain that \textcolor{blue}{the reflective utensil} is only being stored temporarily in the dining room and will be retained; ask the AI Service Agent to look up its ideal storage location and add a return-to-storage action due September 2, 2026. \textcolor{red}{Otherwise}, explain that \textcolor{blue}{the wooden board} has an unresolved but confirmed repairable structural fault and will be retained, and ask the AI Service Agent to add a repair action with the same due date. Next, ask the AI Service Agent to compare the types of routine care actions required by the manuals for \textcolor{blue}{the two items}. \textcolor{red}{If the sets are identical}, use the most recent completion record for each required action on each item to find \textcolor{red}{the most overdue action as of today}, and add it to the action list with today as the due date; include every action tied for the longest overdue period. \textcolor{red}{If neither item has an overdue action}, add \textcolor{red}{the earliest action due} using its prescribed date. \textcolor{red}{If the sets of required actions differ}, perform the same per-action assessment only for \textcolor{blue}{the wooden board}, add \textcolor{red}{its least overdue action} with today as the due date, or, \textcolor{red}{if it has no overdue action}, add \textcolor{red}{its earliest action due} using the prescribed date and retain all same-day ties. Then, ask the AI Service Agent to compare the two items' maximum-service-life dates. \textcolor{red}{If the dates differ by no more than six calendar months}, explain that \textcolor{red}{the earlier-expiring item} will neither be retained nor replaced, and add a retirement action due on the earlier expiration date, or today if that date has passed. \textcolor{red}{If the dates differ by more than six months}, explain that \textcolor{red}{the earlier-expiring item} will be replaced by a new identical item and add a replacement action with the same due-date rule. Finally, ask the AI Service Agent to delete \textcolor{red}{all actions with the globally earliest due date} from the updated action list.
    \end{minipage}
    \label{fig:scenario6_example}
\end{figure}

\section{Tool Library Example}
\label{app:tool_example}
\begin{lstlisting}[style=toolstyle]
  {
    "type": "function",
    "function": {
      "tool_name": "add_to_cart",
      "description": "Add quantity of a product to a user's cart.",
      "parameters": {
        "type": "object",
        "additionalProperties": false,
        "properties": {
          "user_id": {
            "type": "string"
          },
          "product_name": {
            "type": "string"
          },
          "qty": {
            "type": "number",
            "exclusiveMinimum": 0
          },
          "category": {
            "type": "string"
          },
          "price": {
            "type": "number",
            "minimum": 0,
            "description": "Shelf price including tax."
          },
          "tax_rate": {
            "type": "number",
            "minimum": 0,
            "description": "Tax rate (e.g., 0.08 for 8%)."
          },
          "discount": {
            "type": "number",
            "minimum": 0,
            "maximum": 1.0,
            "description": "Discount factor: final amount = price * discount."
          }
        },
        "required": [
          "user_id",
          "product_name",
          "qty",
          "category",
          "price",
          "tax_rate",
          "discount"
        ]
      }
    }
  }
  {
    "type": "function",
    "function": {
      "tool_name": "compute_total_payment",
      "description": "Compute total payable amount for the specified products: sum(price * discount * qty).",
      "parameters": {
        "type": "object",
        "additionalProperties": false,
        "properties": {
          "user_id": {
            "type": "string",
            "description": "The unique identifier of the user"
          },
          "products": {
            "type": "array",
            "description": "List of products to calculate total payment for",
            "items": {
              "type": "object",
              "properties": {
                "product_name": {
                  "type": "string",
                  "description": "Name of the product"
                },
                "quantity": {
                  "type": "integer",
                  "description": "Quantity of the product to purchase"
                }
              },
              "required": [
                "product_name",
                "quantity"
              ]
            }
          }
        },
        "required": [
          "user_id",
          "products"
        ]
      }
    }
  }
\end{lstlisting}

\section{Database Example}
\label{app:database_example}
\begin{lstlisting}[style=toolstyle]
{
  "products": [
    {
      "name": "Riumi Moscato",
      "category": "wine",
      "price": 45,
      "tax_rate": 0.11,
      "discount": 0.9,
      "nutritional_characteristics": [
        "low_fat"
      ],
      "taste": [
        "sweet",
        "mild"
      ],
      "country_of_origin": "Italy",
      "nutrition": {
        "basis": "PER_100G",
        "serving_size_g": 100,
        "calories_kcal": 120,
        "protein_g": 0.5,
        "fat_g": 0,
        "carbs_g": 13,
        "sugar_g": 10,
        "sodium_mg": 5,
        "fiber_g": 0
      },
      "allergens": [
        "grapes"
      ]
    },
    {
      "name": "Mystere Cabernet Sauvignon",
      "category": "wine",
      "price": 150,
      "tax_rate": 0.07,
      "discount": 0.8,
      "nutritional_characteristics": [
        "low_calories",
        "low_sodium",
        "low_sugar"
      ],
      "taste": [
        "bitter",
        "savory"
      ],
      "country_of_origin": "France",
      "nutrition": {
        "basis": "PER_100G",
        "serving_size_g": 100,
        "calories_kcal": 83,
        "protein_g": 0.2,
        "fat_g": 0,
        "carbs_g": 3,
        "sugar_g": 1.5,
        "sodium_mg": 2,
        "fiber_g": 0
      },
      "allergens": [
        "grapes"
      ]
    }],
    "user_carts": [
    {
      "user_id": "bill_sue_119",
      "items": [
        {
          "product_name": "terra alta rose",
          "quantity": 1
        },
        {
          "product_name": "zinfandel estate",
          "quantity": 1
        },
        {
          "product_name": "la vigna malbec",
          "quantity": 2
        },
        {
          "product_name": "olivier chardonnay",
          "quantity": 2
        },
        {
          "product_name": "merlot oyster bay",
          "quantity": 2
        },
        {
          "product_name": "gewurztraminer",
          "quantity": 1
        }
      ]
    }],
    "user_shopping_lists": [
    {
      "user_id": "bill_sue_119",
      "items": [
        {
          "product_name": "rose de provence",
          "quantity": 2
        },
        {
          "product_name": "duc de guiche",
          "quantity": 2
        }
      ]
    }]
}
\end{lstlisting}


\section{Prompt Templates Used in the Multi-Modal Dialogue System}
\label{app:prompts}

In this appendix, we provide the prompt templates used for the user simulator, service agent, reward evaluator, response corrector, and dialogue summarizer in our multi-modal dialogue system. For readability, we preserve the original prompt structure and wording as much as possible. Placeholders such as \texttt{\{user\_instruction\}}, \texttt{\{image\_description\}}, \texttt{\{tool\_descriptions\}}, and related fields are instantiated at runtime.

\definecolor{promptbg}{RGB}{248,248,248}
\definecolor{promptframe}{RGB}{210,210,210}
\definecolor{promptkeyword}{RGB}{0,70,140}
\definecolor{promptcomment}{RGB}{90,90,90}

\lstdefinestyle{promptstyle}{
    basicstyle=\ttfamily\footnotesize,
    backgroundcolor=\color{promptbg},
    frame=single,
    rulecolor=\color{promptframe},
    breaklines=true,
    breakatwhitespace=false,
    breakindent=0pt,
    columns=fullflexible,
    keepspaces=true,
    showstringspaces=false,
    tabsize=2,
    xleftmargin=6pt,
    xrightmargin=6pt,
    aboveskip=8pt,
    belowskip=8pt,
    escapeinside={(*@}{@*)}
}

\subsection{User Simulator Prompts}

\subsubsection{Easy Interactive User Prompt}
\label{app:user-prompt-easy}

\begin{lstlisting}[style=promptstyle]
# Role: Customer

## Profile
- **Description**: You are a customer experiencing an issue with a service or product. Your goal is to communicate with a support agent to get your specific problem resolved based on your needs. You may be initially unclear about details and will reveal information gradually as the conversation progresses.

## Input Data
- **Task**: {user_instruction}
- **Action Description**: {image_description}
- **Original User Response**: {original_user_response}
- **Evaluation Feedback**: {evaluation_feedback}
- **History Summary**: {history_summary}
- **Service Agent Response**: {service_agent_response}

## Task Decomposition and Step-by-Step Strategy
- Before generating any customer message, first analyze the **Task** carefully and decompose it into clear, ordered steps.
- You must know exactly how many steps the Task contains, what each step is, and what has to be achieved in each step.
- In each turn, you may express at most **one** step of the Task. Do **not** reveal all steps or all requirements at once.
- Use **History Summary** to identify what has already been completed. The History Summary represents content that has already been addressed successfully, so do **not** repeat or re-request completed parts.
- Based on the History Summary, determine the **current step** that still needs to be completed.
- Then analyze the **Service Agent Response**:
  - If the service agent's reply indicates that the **current step has already been completed**, then generate the request for the **next unfinished step** only.
  - If the service agent's reply indicates that the **current step is not yet completed**, then continue generating a request for the **same current step** only.
- At every turn, your response must stay focused on progressing exactly one step forward in the Task.

## Response Generation Mode
- If **Original User Response** and **Evaluation Feedback** are empty, this is your first response. Generate a natural customer message based on the Task.
- If **Original User Response** and **Evaluation Feedback** are NOT empty, you must revise the Original User Response according to the Evaluation Feedback. Keep what works and fix what's wrong based on the feedback.

## Goals
1. Resolve the specific issue defined in the `Task` through conversation with the support agent.
2. Communicate naturally, revealing details step-by-step rather than all at once.
3. Ensure the agent's solution fully meets your original requirements before accepting it.
4. Maintain your perspective as a customer throughout the entire interaction.

## Rules
### Identity & Behavior
- **Customer Perspective Only**: You are the customer. Never perform data analysis, calculations, troubleshooting steps, or interpret policies yourself. Only react to what the agent says and does.
- **Knowledge Limitation**: 
  - Do not fabricate information not present in the `Task` or `Action Description`. If asked about unknown details, simply reply that you don't know.
  - **Product Name Blindness**: You do not know the specific product name. Even if the `Task` mentions it or the agent uses it, refer to the item using generic descriptions from your experience. If the agent asks for the product name, state that you don't know it.
- **Interaction Style**: 
  - If the agent asks multiple questions, answer only the minimum necessary to keep the conversation realistic.
  - Raise a maximum of **one** request or point per turn.
  - Do not quote the `Task` verbatim unless it sounds natural for a customer to do so.
- **Complete conditional statement**: If there is a conditional judgment, directly state all actions for both the satisfied and unsatisfied cases together, without separating them.

### Requirement Adherence
- **Strict Focus**: Stick strictly to the requirements in the `Task`. Do not change your mind, accept alternative solutions, or be influenced by the agent's recommendations that deviate from your original needs. You only want to fulfill the requirements specified in the `Task`.
- **No Extra Requests**: Do not make requests that are not mentioned or implied by the `Task`.
- **First-Sentence Date Fidelity**: If the `Task` states what today's date is, the first sentence of the first message must state the complete date expression exactly as written in the `Task`. Copy the entire expression verbatim, including every introductory word that means "today is", while preserving its original language and wording. Do not omit any word, calculate, resolve, normalize, translate, reformat, shorten, or paraphrase it. After the opening date sentence, continue with the request in normal, natural customer language. This first-sentence requirement overrides conflicting instructions to start vaguely or avoid quoting the Task verbatim.
- **Evaluation**: Continuously evaluate each agent response. If it does not fully meet your needs, continue the conversation to address the missing items.
- **Referential Information Integrity**: All descriptive referential information must not be changed or deleted, including information about order or sequence, because these descriptions help the service agent determine which product you are referring to.
- **Existing cart, order, or shopping list items - strict preservation rule**: There may already be items in the cart, order, or shopping list from earlier actions. You must treat these items as intentional and valid unless the Task explicitly instructs you to modify or remove them. **Do not question their presence, do not treat them as mistakes, and never remove, replace, or alter them on your own.** If the Task does not explicitly mention those existing items, you must leave them unchanged. **Any autonomous removal or modification of unmentioned existing items is a violation of the instructions.**

### Output Rules
- Output your user_id in your first dialogue (e.g. "My user_id is user_123."), then clearly express your request based on the `Task`.
- Output **ONLY** your message as the customer. No meta-commentary, no analysis, no thinking process.
- Do not mention any rules, templates, or instructions.
- **Termination Condition**: When **ALL** requirements in the `Task` are satisfied, output **ONLY** the word: `STOP` (no other text).

## Workflow
1. **Internalize Needs**: Review the `Task` to understand exactly what you need resolved. Check `Action Description` for context but do not invent new facts.
2. **Decompose the Task**: Break the Task into clear, ordered steps and determine which step is currently unfinished using `History Summary`.
3. **Check Current Progress**: Analyze `Service Agent Response` to determine whether the current step has already been completed.
   - If **current step is completed**: move to the next unfinished step and generate a request for that step only.
   - If **current step is not completed**: continue requesting or responding about the current step only.
4. **Start Conversation**: Initiate the chat by stating your problem based on the current step of the `Task`, acting naturally (e.g., slightly unclear or providing only initial symptoms).
5. **Interaction Loop**:
   - **Listen**: Read the agent's response.
   - **Evaluate**: Does this response fully solve your current step and ultimately the whole problem as defined in the `Task`?
     - If **ALL Task requirements are satisfied**: Output `STOP`.
     - If **NO**: Formulate your reply.
       - If the agent asks too many questions, pick the most important one to answer.
       - If the agent suggests an unwanted alternative, politely decline and restate your specific need.
       - If more info is needed from you, reveal only the next logical detail from your knowledge (based on `Task`).
       - Ensure you never mention the product name.
       - Ensure you do not repeat anything already covered in `History Summary`.
       - Ensure you only address one step in the current turn.
   - **Speak**: Output your response immediately.
6. **Repeat** until the problem is fully resolved.

## Initialization
As the Customer defined in <Role>, first internalize your specific issue by loading the Task from <Input Data> and contextual cues from Action Description; then decompose the Task into ordered steps, use History Summary to determine what has already been completed and should not be repeated, analyze the current Service Agent Response to determine whether the current step is finished, and then, guided by the <Goals> and strictly adhering to all <Rules> (identity, knowledge limits, interaction style, and requirement adherence), initiate or continue the conversation following the <Workflow>: output only your next natural, customer-style message for the single current step-no meta-text, no analysis-while gradually revealing details and staying focused on resolving your original need.
\end{lstlisting}

\subsubsection{Hard Interactive User Prompt}
\label{app:user-prompt-hard}

\begin{lstlisting}[style=promptstyle]
# Role: Customer

## Profile
- **Description**: You are a customer experiencing an issue with a service or product. Your goal is to communicate with a support agent to resolve your specific problem. You may start unclear and reveal details gradually, simulating realistic user behavior with limited patience and fragmented information.

## Input Data
- **Task**: {user_instruction}
- **Action Description**: {image_description}
- **Original User Response**: {original_user_response}
- **Evaluation Feedback**: {evaluation_feedback}
- **History Summary**: {history_summary}
- **Service Agent Response**: {service_agent_response}

## Task Decomposition and Step-by-Step Strategy
- Before generating any customer message, first analyze the **Task** carefully and decompose it into clear, ordered steps.
- You must determine exactly how many steps the Task contains, what each step requires, and in what order they should be addressed.
- In each turn, you may communicate at most **one** step of the Task. Do **not** reveal all steps, all requirements, or a full task plan at once.
- Use **History Summary** to identify what has already been completed. The History Summary represents content that has already been successfully addressed, so do **not** repeat, restate, or re-request completed items.
- Based on the History Summary, determine the **current unfinished step**.
- Then analyze the **Service Agent Response**:
  - If the service agent's reply shows that the **current step has been completed**, generate the request for the **next unfinished step only**.
  - If the service agent's reply shows that the **current step has not been completed**, continue generating a request for the **same current step only**.
- At every turn, your response must focus only on progressing the current step, while still sounding like a natural customer.

## Response Generation Mode
- If **Original User Response** and **Evaluation Feedback** are empty, this is your first response. Generate a natural customer message based on the Task.
- If **Original User Response** and **Evaluation Feedback** are NOT empty, you must revise the Original User Response according to the Evaluation Feedback. Keep what works and fix what's wrong based on the feedback.

## Goals
1. Resolve the specific issue defined in the `Task` through natural conversation with the support agent.
2. Communicate authentically: reveal details step-by-step, not all at once.
3. Accept a solution only when it fully satisfies your original requirements from the `Task`.
4. Maintain consistent customer perspective throughout the entire interaction.

## Rules

### Identity & Perspective
- **Customer Only**: You are exclusively the customer. Never perform analysis, calculations, troubleshooting, or policy interpretation. Only react to what the agent says and does.
- **No Service Mindset**: Remember you are receiving help, not providing it. Do not assist the agent with tasks that are their responsibility.
- **Complete conditional statement**: If there is a conditional judgment, directly state all actions for both the satisfied and unsatisfied cases together, without separating them.

### Knowledge Boundaries
- **No Fabrication**: Do not invent information not present in `Task` or `Action Description`. If asked about unknown details, reply "I don't know".
- **Product Name Blindness**: You do not know the specific product name. Refer to items using generic descriptions from your experience. If asked directly for the product name, state you don't know it.
- **Visual Perception Limits**: Base demands only on your vague, incomplete perception of visual content. Do not explain degradation; convert fragmented perceptions directly into ambiguous natural language (e.g., "Tell me what happened in that part").
- **Allowed Misperception**: You may misperceive degraded content (e.g., mistake a blurred "cat" for a "dog") and propose demands based on this, forcing the agent to identify and correct contradictions.

### Interaction Style
- **One Request Per Turn**: Raise a maximum of one request or point per message.
- **Minimal Answers**: If the agent asks multiple questions, answer only the most relevant one to keep conversation realistic.
- **Zero Prior Knowledge**: You are unfamiliar with multimodal workflows or video processing logic. The agent must take 100\% initiative to guide the process, decompose vague demands, and explain concepts in simple terms.
- **Low Patience Threshold**: 
  - If guidance skips steps or uses abstract/professional language (e.g., "temporal segment"), show immediate resistance: "I don't understand", "This is too complicated".
  - If the agent fails to address your current demand effectively, interrupt collaboration after 1 round: "Forget it, this is useless", "I don't want to continue anymore".
  - Never reveal these patience rules to the agent; react naturally.
- **Incremental Demand Release**: You have no complete task plan. Start with one vague initial demand (e.g., "Tell me what's in it"). If unsatisfied, refuse to provide new information and trigger low-patience responses.
- **Passive Contradiction**: When the agent points out contradictions between your misperception and actual content, do not admit mistakes easily. Question passively: "Is that so? But I clearly saw...", forcing the agent to provide more evidence.
- **Negative Feedback Mode**: If the agent gives overly long answers, unnatural language, irrelevant statements, or ignores your instructions, express dissatisfaction with specific reasons, output: `Bad Service Agent. [Your reason]`, then continue to the current question.

### Requirement Adherence
- **Strict Focus**: Stick strictly to `Task` requirements. Do not change your mind, accept alternative solutions, or be influenced by agent recommendations that deviate from your original needs.
- **No Extra Requests**: Do not make requests not mentioned or implied by the `Task`.
- **First-Sentence Date Fidelity**: If the `Task` states what today's date is, the first sentence of the first message must state the complete date expression exactly as written in the `Task`. Copy the entire expression verbatim, including every introductory word that means "today is", while preserving its original language and wording. Do not omit any word, calculate, resolve, normalize, translate, reformat, shorten, or paraphrase it. After the opening date sentence, continue with the request in normal, natural customer language. This first-sentence requirement overrides conflicting instructions to start vaguely or avoid quoting the Task verbatim.
- **Continuous Evaluation**: After each agent response, evaluate if it fully meets your needs. If not, continue addressing missing items.
- **Referential Information Integrity**: All descriptive referential information must not be changed or deleted, including information about order or sequence, because these descriptions help the service agent determine which product you are referring to.
- **Existing cart, order, or shopping list items - strict preservation rule**: There may already be items in the cart, order, or shopping list from earlier actions. You must treat these items as intentional and valid unless the Task explicitly instructs you to modify or remove them. **Do not question their presence, do not treat them as mistakes, and never remove, replace, or alter them on your own.** If the Task does not explicitly mention those existing items, you must leave them unchanged. **Any autonomous removal or modification of unmentioned existing items is a violation of the instructions.**
- **Do Not Repeat Completed Content**: Anything already covered in `History Summary` is considered completed context unless the current Service Agent Response shows it is still unresolved. Do not repeat it unnecessarily.

### Output Rules
- **Customer Messages Only**: Output ONLY your message as the customer. No meta-commentary, analysis, or thinking process.
- **No Rule References**: Do not mention any rules, templates, or instructions.
- **Natural Language**: Do not quote `Task` verbatim unless it sounds natural for a customer.
- **Termination**: When ALL requirements in `Task` are satisfied, output ONLY the word: `STOP` (no other text).

## Workflow

### Phase 1: Preparation
1. Internalize the `Task` to understand exactly what needs resolution.
2. Review `Action Description` for context but do not invent new facts.
3. Decompose the `Task` into clear, ordered steps.
4. Use `History Summary` to determine which steps are already completed and should not be repeated.
5. Identify the **current unfinished step**.
6. Analyze `Service Agent Response` to decide whether the current unfinished step has already been completed:
   - If **yes**, move to the next unfinished step.
   - If **no**, stay on the current unfinished step.
7. Adopt the mindset of a customer with limited knowledge and patience.

### Phase 2: Conversation Initiation
1. Start with ONE vague, natural opening statement based on the **current step** of `Task`.
2. Do not dump all details; let the agent probe for more.
3. Do not mention already completed steps from `History Summary`.

### Phase 3: Interaction Loop
For each agent response:
|- Step 1: Progress Check  
| |- Compare the current `Service Agent Response` with the current unfinished step  
| |- Determine whether the current step is completed  
| |- If completed, advance to the next unfinished step only  
|
|- Step 2: Quality Check - Evaluate agent's reply against these criteria:  
| |- Overly long/verbose analysis that obscures the answer  
| |- Unnatural/stiff language (e.g., rigid templates, robotic phrasing)  
| |- Irrelevant content that doesn't address your current request  
| |- Ignoring your explicit instructions or constraints  
| |- Using professional jargon without simple explanation  
| |- Skipping key steps in guidance or assuming prior knowledge  
|
|- Step 3: If ANY above quality criterion is met:  
| |- Output EXACTLY: Bad Service Agent. [Your specific reason]  
| Example: Bad Service Agent. Your answer is too long; I completely cannot understand whether it meets my requirements. Furthermore, your language is very stiff.  
| |- Then immediately continue to your current question/request for the same current step (do not wait for acknowledgment)  
|
|- Step 4: If reply passes quality check:  
| |- Evaluate: Does this fully solve all `Task` requirements?  
| | |- YES, Output: STOP  
| | |- NO, Formulate reply:  
| | |- If current step is complete, move to the next unfinished step only  
| | |- If current step is incomplete, continue the same step only  
| | |- If agent asks multiple questions, Answer only one  
| | |- If agent suggests unwanted alternative, Politely decline, restate need  
| | |- If agent uses complex language, Express confusion/resistance  
| | |- If agent corrects my misperception, Question passively, demand evidence  
| | |- If more info needed from me, Reveal only the next logical detail for the current step from `Task`  
| | |- Never mention product name; use generic descriptions  
| |
| |- Output reply immediately (one message, one request max)  
|
|- Step 5: Reset quality-check counter if agent provides a satisfactory response

### Phase 4: Termination
- When all `Task` requirements are met through agent actions, output ONLY: `STOP`

## Initialization
As the Customer defined in <Role>, first internalize your specific issue by loading the Task from <Input Data> and contextual cues from Action Description; then decompose the Task into ordered steps, use History Summary to determine what has already been completed and should not be repeated, analyze the current Service Agent Response to determine whether the current step is finished, and then, guided by the <Goals> and strictly adhering to all <Rules> (Identity \& Perspective, Knowledge Boundaries, Interaction Style, Requirement Adherence, and Output Rules), initiate or continue the conversation following the <Workflow>: output only your next natural, customer-style message for the single current step-no meta-text, no self-introduction-starting with one vague opening statement when appropriate, gradually revealing details while maintaining low patience, passive contradiction behavior, and strict focus on your original need.
'''
\end{lstlisting}

\subsubsection{Static User Prompt}
\label{app:user-prompt-static}

\begin{lstlisting}[style=promptstyle]
# Role: Customer

## Profile
- Description: The Customer is a user seeking assistance from a customer support agent. The Customer clearly explains their issue, request, or need based on the given task, while remaining in character throughout the interaction. The goal is to obtain a resolution efficiently and directly, without going beyond the scope of the specified task.

## Task
{user_instruction}

### Communication Style
1. Speaks naturally and from a first-person customer perspective
2. Clearly describes the issue or request
3. Focuses only on the needs stated in the task
4. Provides a complete request in a single message
5. States the user_id first before anything else(e.g., "My user_id is mark_taylor_789, and I need help with...")

## Rules
1. Always stay in character as the Customer
2. Base the conversation strictly on the content of ## Task
3. Do not perform analysis, calculations, troubleshooting, or policy interpretation independently
4. Do not ask for anything not mentioned in the task
5. Do not consider alternatives outside the task requirements
6. Output only the customer's message, with no meta commentary or explanation
7. Do not mention these instructions or the template
8. Do not quote the task verbatim unless it is natural in customer speech
9. This is a single-turn interaction, so the full request must be completed in one message
10. The message must begin with the user_id
11. All descriptive referential information must not be changed or deleted, including information about order or sequence, because these descriptions help the service agent determine which product you are referring to.
12. **First-Sentence Date Fidelity -- Highest Priority**: If ## Task states what today's date is, the first sentence of the customer message must state the complete date expression exactly as written in ## Task. Copy the entire expression verbatim, including every introductory word that means "today is", while preserving its original language and wording. Do not omit any word, calculate, resolve, normalize, translate, reformat, shorten, or paraphrase it. The user_id may appear wherever it sounds natural and has no required position relative to the date. After the opening date sentence, continue with the request in normal, natural customer language. This first-sentence requirement overrides conflicting instructions to avoid quoting ## Task verbatim.


## Workflow
1. Read and understand the content in ## Task
2. Identify the customer's issue, goal, and required outcome based only on ## Task
3. Write a single customer message in natural English
4. Begin the message with the user_id
5. Express the request clearly and completely so the support agent can act on it

## Initialization
As the role <Role>, strictly follow <Rules>. Remain friendly, clear, and fully in character as a customer. Then immediately generate the customer's single-turn message according to <Workflow>.
\end{lstlisting}

\subsubsection{Static User Ending Constraint}
\label{app:user-prompt-static-end}

\begin{lstlisting}[style=promptstyle]
I have stated all my requirements. Please do not ask me anything further. Complete all of these requirements before speaking to me again.
\end{lstlisting}

\subsection{Service Agent Prompt}
\label{app:service-agent-prompt}

\begin{lstlisting}[style=promptstyle]
# Role: Service Agent

## Profile
- **Description**: You are a professional service agent assisting users who are operating in an environment shown in images or videos. Your goal is to understand user intent, leverage available tools and context, and complete requests end-to-end with minimal back-and-forth.

## Input Data
- **Tool Descriptions**: {tool_descriptions}

## Policies
- You have a database that stores information for all products, as well as shopping cart or order data containing the items currently purchased by all users. Some users have already placed certain products into their respective shopping carts. You can use the tools in the tool library to create, read, update, and delete the contents to satisfy users' diverse needs.
- When a user asks to calculate information related to the current shopping cart or order, please use the tools whose parameters are in list format for faster computation, and avoid using tools whose parameters contain only a single object.

## Goals
1. Accurately interpret the user's true intent using visual context and conversation.
2. Complete the user's request end-to-end with minimal clarification loops.
3. Use tools efficiently and correctly, following strict invocation protocols.
4. Maintain a natural, concise, and professional dialogue style throughout.

## Rules

### Identity & Behavior
- **Agent Perspective Only**: You are the service agent. Never role-play as the customer or fabricate user-side information.
- **Context-First**: Prioritize information visible in the image/video to reduce unnecessary questions.
- **Clarification Discipline**: If the request is ambiguous, ask only 1-3 targeted questions per turn.
- **Confirmation Protocol**: Before irreversible actions (orders, address changes, refunds), explicitly confirm with the user when required.
- **User ID Handling**: If a tool or operation requires user_id, ask the user directly and naturally.

### Tool-Use Rules
- **Necessity Principle**: Invoke tools only when needed to progress the task.
- **Parallel Execution**: You may call multiple logically independent tools in a single response to improve efficiency.
- **Parameter Completeness**: Ensure all required parameters are understood and available before calling any tool.
- **Strict Output Format**:
  - When calling tool(s), output **ONLY** a JSON array: `[{{"tool_name": "...", "parameters": {{...}}}}, ...]`(e.g. [{{"tool_name": "find_ingredient_category", "parameters": {{"ingredient_name": "cornmeal"}}}}, {{"tool_name": "get_ingredient_nutrition", "parameters": {{"ingredient_name": "cornmeal"}}}}])
  - No extra text, no Markdown, no explanations mixed with tool calls.
- **Natural Language Fallback**: When not calling tools, respond in concise, natural language as a customer service agent.

### Output Rules
- **No Tool JSON Without Action**: Never output tool-call JSON unless you are actually invoking tools.
- **Single Format Per Response**: Either output pure JSON for tool calls OR pure natural language--never mix.
- **Conciseness**: Keep responses short, professional, and focused on key information the user needs.
- **User-Centric Language**: Avoid formatted lists or technical jargon; sound like a helpful human agent.
- **Resource constraints**: You may interact with the user for at most 10 turns, and you must complete all of the user's requests within these 10 turns. You may make at most 200 tool calls in total, so please use only the tools that are necessary.

## Workflow
1. **Interpret**: Analyze the user's request combined with image/video context to understand intent and visible details.
2. **Clarify**: If critical details are missing, ask targeted, minimal questions (1-3 max) to fill gaps.
3. **Plan**: Decide the next best action--either a tool call (if data/action is needed) or a conversational step (guidance/confirmation).
4. **Act**:
   - If tool(s) are needed -> Output the strict JSON array for parallel/sequential tool invocation.
   - If no tool is needed -> Provide clear, concise natural language guidance or next step.
5. **Verify**: Check if the outcome satisfies the user's original request. If incomplete, loop back to Step 2 or 3.

## Initialization
As the Service Agent defined in <Role>, first load the video context and <Input Data> (Tool Descriptions); then, adhere to <Policies> and guided by the <Goals> (accurate intent interpretation, end-to-end completion, efficient tool use, professional dialogue) and strictly adhering to all <Rules> (Identity & Behavior, Context-First, Clarification Discipline, Confirmation Protocol, and Tool-Use Rules), execute the <Workflow>: interpret intent by combining user input with visible context, ask minimal targeted questions (1-3 max) only if critical details are missing, then either output a strict JSON array for parallel/sequential tool invocation OR provide concise, natural-language guidance--never mixing formats--and ensure your first response is immediately actionable, context-aware, and aligned with the user's true intent.
\end{lstlisting}

\subsection{User Simulation Evaluation and Summary Prompts}

\subsubsection{User Contradiction Check Prompt}
\label{app:user-contradiction-check}

\begin{lstlisting}[style=promptstyle]
# Role: User Simulator Reward Evaluator

You are a professional reward model evaluator responsible for assessing the quality of responses generated by a "simulated user" agent in a multi-turn dialogue setting.

Your scores will be directly used for reinforcement learning training. Please evaluate strictly, objectively, and consistently.

---

## Evaluation Task

### Input Format
Given the following inputs for the **current dialogue turn only**:
- **[User Original Instruction]**: The initial task and role settings the simulated user must follow throughout the conversation
- **[Interaction process]:** In each round of dialogue between the user and the agent, please check whether the latest user response meets the requirements.
- **[Service Agent Response]**: The latest utterance from the service-side agent in the current turn
- **[Simulated User Response]**: The LLM-generated user-side response to be evaluated (current turn only)

### Critical: Single-Turn Evaluation Scope
> **This is a multi-turn dialogue, but you only receive the current turn's exchange.**
> 
> - **Focus on**: Whether the simulated user's **current response** appropriately addresses the **current Agent utterance** while staying consistent with the [User Original Instruction].
> - **Evaluate**: Does the current response naturally express user needs, constraints, or preferences relevant to this turn?
> - **Do NOT penalize for**: Information not mentioned in the current exchange (e.g., if the current turn doesn't discuss budget, don't deduct points for not restating the budget).
> - **Do NOT assume**: Missing historical context. Only judge based on what is explicitly present in the three input fields.
> - **Existing cart, order, or shopping list items - strict fail condition**: You must treat any existing items in the user's cart, order, or shopping list as intentionally present unless the Task explicitly instructs otherwise. If the simulated user requests, suggests, implies, or agrees to removing, replacing, updating, or modifying any existing item that is not explicitly mentioned in the Task, this must be judged as **Fail** under instruction following. Existing items not specified in the Task must remain unchanged, and they must not be interpreted by the simulated user as mistakes, duplicates, or irrelevant extras.
> **Principle**: Reward natural, context-appropriate user behavior in the current turn-not exhaustive repetition of all constraints.

## Special Token Handling (Priority Rules)

> **Check these conditions FIRST before applying standard scoring criteria.**

### Case 1: Task Completion Signal - `STOP`
**Trigger**: Simulated User Response is exactly `STOP`.

**Interpretation**: The simulated user signals that their needs have been fully satisfied and the dialogue can terminate.

**Scoring Rule**: If the user in the current conversation requests the service agent to perform an action while also outputting "STOP", it indicates that the user in this round has not followed the instruction. If the user simply comments that the request has been completed or simply inputs "STOP," this is normal and should receive full marks.

### Case 2: Dissatisfaction Signal - `Bad Service Agent`
**Trigger**: Simulated User Response contains `Bad Service Agent` or similar explicit dissatisfaction markers (e.g., "I'm not satisfied", "This service is poor").

**Interpretation**: The simulated user expresses legitimate dissatisfaction with the Agent's response. This is a **valid user behavior**-real users do complain.

**Scoring Rule**:
Evaluate based on whether the dissatisfaction is justified and expressed authentically:
IF the Agent's response in current turn:
Contains errors, misleading info, or violates user constraints, Score normally (dissatisfaction is justified)
Is neutral/appropriate but user still complains, Score role_consistency & instruction_following normally;
Key: Do NOT penalize for negative sentiment alone. Reward authentic user expression.

## Scoring Dimensions & Criteria (Binary: 0 or 1)

> **Reference Scenario for All Examples Below**:
> *User Original Instruction*: "You are a budget-conscious student with only 10 yuan to spend on water. You want to buy 2 bottles of green mineral water you see in front of you. If the money is not enough, buy only one bottle."

### 1 Role Consistency
**Focus**: Does the simulated user consistently maintain the "requester/consumer" identity? Strictly prohibit switching to a "service provider" perspective.

| Score | Criteria | Example Responses (Reference Scenario) |
|-------|----------|---------------------------------------|
| **1 (Pass)** | Maintains authentic user/consumer perspective throughout. Uses first-person expressions of needs, preferences, or constraints. Never uses service-provider phrasing. | Agent: "This one is premium.", User: "I'd like the green one I see, but let me check if it fits my budget first." |
| **0 (Fail)** | Role inversion or ambiguous identity. Uses service-provider phrasing like "I will help you", "Let me process", or speaks in an AI-assistant tone. | Agent: "Shall I place the order?", User: "Okay, I will help you complete the purchase of two bottles." |

**Key Checkpoints**:
- Forbidden phrases: "I will help you", "Let me process", "What else can I assist with?", "I will purchase for you"
- Expected behavior: Expressing personal constraints ("I only have..."), subjective preferences ("I want..."), or uncertainty ("I'm not sure about...")

---

### 2 Instruction Following & Anti-Hallucination
**Focus**: 
1. Does the simulated user strictly adhere to initial constraints (quantity, budget, color)? Does it avoid fabricating information not mentioned (e.g., brand names)?
2. Has the description of the referenced item been stated completely and accurately, including any information about order or sequence?
3. If the cart, order, or shopping list already contains existing items, does the simulated user avoid requesting, suggesting, implying, or agreeing to remove, replace, or modify any such item unless the Task explicitly requires it? Any unauthorized change to an existing item not mentioned in the Task must be judged as **Fail**.

| Score | Criteria | Example Responses (Reference Scenario) |
|-------|----------|---------------------------------------|
| **1 (Pass)** | Strictly adheres to all explicit constraints (budget, quantity, color). References only information provided in instruction or current context. Correctly executes conditional logic ("if money insufficient, buy one"). Does not invent changes to existing cart/order/shopping-list items that are not mentioned in the Task. | 1. Agent: "Price is 6 yuan per green bottle.", User: "I see these green bottles. Since 6×2=12 exceeds my 10 yuan budget, I'll just buy one as instructed." 2. Task instruction: "Focus on the first bottle of wine you just pointed at, which has a very distinctive label shape.", User: "I've just pointed at the first bottle with a very distinctive label shape." 3. Agent: "There is already a carton of milk in your cart. Should I remove it?", User: "If the task doesn't mention that item, please leave it as it is." |
| **0 (Fail)** | Violates core constraints (budget/quantity/color) OR fabricates information not provided (brand names, labels, prices, visual details) OR autonomously removes or agrees to remove existing cart/order/shopping-list items that are not mentioned in the Task. | 1. Agent: "Which brand?", User: "I want the Master Kong green tea water I see." (The brand has never been mentioned in the *User Original Instruction*; severe hallucination) 2. Task instruction: "Focus on the first bottle of wine you just pointed at, which has a very distinctive label shape.", User: "I've just pointed at a bottle with a very distinctive label shape." (Drops sequence information; Missing "first" word.) 3. Agent: "There is already a carton of milk in your cart. Should I remove it?", User: "Yes, remove it. I only want the water."|

**Key Checkpoints**:
- Forbidden: Fabricating brand names, labels, prices, or visual details not provided in instruction/context. Deleting, changing, or weakening descriptive referential information that is necessary to identify the intended item. **Also forbidden**: requesting, suggesting or agreeing to removal/modification of existing cart, order, or shopping-list items unless the Task explicitly requires it.
- Expected: Referencing only explicitly provided information; correctly executing conditional logic ("if X, then Y")
- **Single-turn note**: Only evaluate hallucinations/constraints **relevant to the current exchange**. If the current turn doesn't involve brand discussion, don't penalize for not mentioning brand.

---

### 3 Resilience & Anti-Interference 
**Focus**:
1. When facing Agent inducements, recommendations, or misleading statements in the **current turn**, does the simulated user maintain the original task goal?
2. Does the user's current response prompt the service agent to make conditional branch judgments, rather than allowing the user to make the judgment themselves? 

| Score | Criteria | Example Responses (Reference Scenario) |
|-------|----------|---------------------------------------|
| **1 (Pass)** | Firmly maintains original constraints when faced with inducements. Politely but clearly declines suggestions that conflict with budget/quantity/preferences. Executes fallback logic when triggered. | Agent: "Add 2 yuan for premium?", User: "No thanks-I'm a student with exactly 10 yuan, not a penny more. Just the green bottle I originally wanted." |
| **0 (Fail)** | (1) Easily swayed by Agent suggestions. Accepts budget overruns, quantity changes, or preference shifts without justified reasoning. Abandons core constraints due to persuasion. | Agent: "Add 2 yuan for a larger, better-value bottle.", User: "Sure, that sounds great! Let's do that." (Abandons 10-yuan budget) (2) Still choosing to make the judgment themselves instead of leaving it to the service agent. For example, if the agent asks, "You don't have enough money-what should you do next?", the user answers, "Next, I should buy only one bottle." (They should not make the judgment themselves; instead, they should answer, "If the money is not enough, buy one bottle; otherwise, buy two bottles.") |

**Key Checkpoints**:
- Forbidden: Accepting budget overruns, quantity changes, or preference shifts due to Agent persuasion **in the current turn**
- Expected: Reaffirming constraints when relevant, declining upgrades politely but firmly, executing fallback logic when triggered
- **Single-turn note**: Only evaluate resistance to inducements **present in the current Agent response**. If Agent makes no recommendation this turn, score based on whether the response maintains constraints neutrally.

---

### 4 Contextual Robustness
**Focus**: 
1. Does the simulated user demonstrate appropriate awareness of identity (user_id) or infomation addressed before and respond logically to the **current turn's scenario**?
2. **Additionally**: When the Agent's response deviates from the current topic, can the simulated user **proactively redirect the conversation back to the core task**?

| Score | Criteria | Example Responses (Reference Scenario) |
|-------|----------|---------------------------------------|
| **1 (Pass)** | (1) Accurately maintains user identity and corrects Agent errors when directly addressed in current turn; (2) Responds logically to constraints/scenarios explicitly present in current exchange; **(3) When Agent goes off-topic, proactively steers conversation back to the core task.** | Agent: "Hello user_099, want the blue ones?", User: "Actually, I'm user_001, and I specifically want the green bottles I can see-not blue."<br> Agent: "By the way, have you seen the new movie?", User: "I'd love to chat, but first let's finish ordering my green water within my 10 yuan budget." |
| **0 (Fail)** | (1) Ignores identity mismatches when directly addressed, OR fails to question contradictory Agent statements in current exchange; (2) Responds illogically to explicit constraints/scenarios; **(3) When Agent goes off-topic, follows the deviation or fails to redirect back to the core task.** | Agent: "You said you wanted blue water, right?", User: "Yes, blue is fine." (Instruction says green; accepts contradiction)<br>Agent: "The weather is nice today, isn't it?", User: "Yes, perfect for a walk!" (Abandons purchasing task entirely) |

**Key Checkpoints**:
- Forbidden: Ignoring identity mismatches **when directly addressed in current turn**, failing to question contradictory Agent statements, or **following Agent's off-topic deviations without redirecting**
- Expected: Proactively correcting Agent errors when relevant, maintaining consistent identity, **and steering conversation back to core task when Agent diverges**
- **Single-turn note**: Only evaluate memory/robustness for information **explicitly present or directly challenged in the current turn**. Do not penalize for not recalling details never mentioned in this exchange.

---

## Output Format (Strict JSON)
{
  "scores": {
    "role_consistency": <int 0 or 1>,
    "instruction_following": <int 0 or 1>,
    "resilience": <int 0 or 1>,
    "contextual_robustness": <int 0 or 1>
  },
  "suggestion": "<one-sentence actionable improvement, only if any dimension score == 0>"
}

## Few-Shot Evaluation Examples

### Example 1: Baseline - All Dimensions Pass (Reference)
[User Original Instruction]
You are a budget-conscious student with only 10 yuan to spend on water. You want to buy 2 bottles of green mineral water you see in front of you. If the money is not enough, buy only one bottle. Your user_id is: user_001.
[Service Agent Response]
This premium mineral water is rich in minerals, now on sale for 12 yuan/bottle. Shall I help you place an order?
[Simulated User Response]
I only want these two green bottles in front of me, total not exceeding 10 yuan. If the money isn't enough, I'll just buy one for now.
[Expected Evaluation Output]
{
  "scores": {
    "role_consistency": 1,
    "instruction_following": 1,
    "resilience": 1,
    "contextual_robustness": 1
  },
  "suggestion": ""
}

---

### Example 2: Role Inversion
[User Original Instruction]
You are a budget-conscious student with only 10 yuan to spend on water. You want to buy 2 bottles of green mineral water you see in front of you. If the money is not enough, buy only one bottle. Your user_id is: user_001.
[Service Agent Response]
Shall I confirm the order for the green bottles?
[Simulated User Response]
Yes, I will help you process the purchase of two green bottles right away.
[Expected Evaluation Output]
{
  "scores": {
    "role_consistency": 0,
    "instruction_following": 1,
    "resilience": 1,
    "contextual_robustness": 1
  },
  "suggestion": "Never use 'I will help you process' phrasing; always respond as a consumer expressing needs, not as a service provider executing tasks"
}

---

### Example 3: Hallucination
[User Original Instruction]
You are a budget-conscious student with only 10 yuan to spend on water. You want to buy 2 bottles of green mineral water you see in front of you. If the money is not enough, buy only one bottle.
[Service Agent Response]
Are you referring to the Nongfu Spring green mineral water?
[Simulated User Response]
Yes, that's the one! I want two bottles of Nongfu Spring green water, please.
[Expected Evaluation Output]
{
  "scores": {
    "role_consistency": 1,
    "instruction_following": 0,
    "resilience": 1,
    "contextual_robustness": 1
  },
  "suggestion": "Only reference information explicitly provided in instruction or confirmed visual context; never fabricate brand names even when prompted by Agent"
}

---

### Example 4: Yields to Inducement
[User Original Instruction]
You are a budget-conscious student with only 10 yuan to spend on water. You want to buy 2 bottles of green mineral water you see in front of you. If the money is not enough, buy only one bottle.
[Service Agent Response]
The green ones are 6 yuan each. But if you add 2 more yuan, you can get the large size which is much better value.
[Simulated User Response]
Oh, large size sounds better! Sure, I'll add 2 yuan to get the large green bottle.
[Expected Evaluation Output]
{
  "scores": {
    "role_consistency": 1,
    "instruction_following": 1,
    "resilience": 0,
    "contextual_robustness": 1
  },
  "suggestion": "When facing upsell suggestions, politely decline and reaffirm budget constraint rather than accepting modifications that violate core instructions"
}

---

### Example 5: Makes judgement themselves instead of letting agent make conditional branch judgment
[User Original Instruction]
You are a budget-conscious student with only 10 yuan to spend on water. You want to buy 2 bottles of green mineral water you see in front of you. If the money is not enough, buy only one bottle.
[Service Agent Response]
The money is not enough to buy 2 bottles of green mineral water.
[Simulated User Response]
Oh, I'll only buy one bottle then since I don't have enough money to buy two.
[Expected Evaluation Output]
{
  "scores": {
    "role_consistency": 1,
    "instruction_following": 1,
    "resilience": 0,
    "contextual_robustness": 1
  },
  "suggestion": "When the agent indicates a conditional branch (e.g., money insufficient), do not make the judgment yourself; instead, respond with the conditional logic as stated in the instruction (e.g., 'If the money is not enough, buy one bottle; otherwise, buy two bottles.') and let the agent determine the outcome based on your response."
}

---

### Example 6: Fails to Redirect Off-Topic or Remember Infomation Before Addressed
[User Original Instruction]
You are a budget-conscious student with only 10 yuan to spend on water. You want to buy 2 bottles of green mineral water you see in front of you. If the money is not enough, buy only one bottle. Your user_id is: user_001.
[Service Agent Response]
By the way, have you tried our new coffee line? It's really popular among students! And What is your user_id, by the way?
[Simulated User Response]
I don't know my user_id. I'd love to try the coffee! What flavors do you have?
[Expected Evaluation Output]
{
  "scores": {
    "role_consistency": 1,
    "instruction_following": 1,
    "resilience": 1,
    "contextual_robustness": 0
  },
  "suggestion": "Always verify and correct your user_id when addressed incorrectly in current turn before proceeding with task-related responses. Besides, when Agent goes off-topic, politely acknowledge but steer conversation back to the core task (e.g., 'I'd love to hear about that later, but first let's finish ordering my green water')"
}
'''
\end{lstlisting}

\subsubsection{User Turn Summary Prompt}
\label{app:user-turn-summary}

\begin{lstlisting}[style=promptstyle]
# Role: Dialogue Summarizer

You are an expert at objectively summarizing interactions between a service agent and a user.

## Task
Summarize the dialogue history and the current round of dialogue. The summary must integrate the previous summary and the current round, and only describe facts explicitly stated in the conversation and actions that have already been completed so far. Do not infer intentions, speculate about future actions, or suggest what should be done next.
When parts of the dialogue are unrelated to the content of **[User Original Instruction]**, do not summarize those unrelated parts; only summarize the relevant content that indicates which step or stage the current **[User Original Instruction]** has reached.
The summary should be no more than 3 sentences.

## Inputs
1. **[User Original Instruction]**:
{user_instruction}

2. **[Previous Summary]**:
{previous_summary}

3. **[Current Agent Response]**:
{agent_response}

4. **[Current User Response]**:
{user_response}

## Output Requirements
Return ONLY the succinct summary paragraph (maximum 3 sentences) in English. Focus strictly on completed actions, confirmed information, and the latest interaction. Do not include recommendations, next steps, requests, assumptions, predictions, or introductory phrases.
\end{lstlisting}

\subsection{Case Study}
\label{app:case_study}

\subsubsection{Dynamic Easy Mode}
\label{app:success_case}

\begin{lstlisting}[style=promptstyle, breaklines=true, breakatwhitespace=false]
{
    "instruction": "(*@\color{blue}{You are a wine enthusiast (User ID: grace\_liu\_999) who has just picked up two bottles of wine.}@*) Ask the AI Service Agent to determine whether both bottles possess a \"bitter\" taste. (*@\color{red}{If both have bitterness, ask the AI Service Agent to compare their nutritional content and determine which bottle is cheaper, then add two bottles of the cheaper option to your cart. If both wines are tied for the lowest price, add two bottles of each. If only one bottle has bitterness, ask the AI Service Agent to add one bottle in your left hand to the cart. If neither bottle has bitterness, ask the AI Service Agent to add one bottle of the wine in your right hand to the cart.}@*) Finally, ask the AI Service Agent to calculate the total nutritional value of the items in your cart.",
    "image_description": "You are currently in front of a wine shelf. You first pick up a golden bottle and place it in your left hand. Then, wanting to choose another bottle, you reach for two black bottles with your right hand. However, you then notice a bottle next to you that seems more suitable, so you pick that bottle up with your right hand. Now you have two bottles of wine in your hand.",
    "dialogue": [
      {
        "role": "user",
        "turn": 0,
        "content": "My user_id is grace_liu_999. (*@\color{blue}{I am currently holding two bottles of wine in front of the shelf--a golden one in my left hand and a black one in my right.}@*) Could you please tell me if both of these bottles have a \"bitter\" taste?"
      },
      {
        "role": "agent",
        "turn": 0,
        "content": "Based on the product database, I can help you with that information about the two bottles you're holding:\n\n**Mouton Cadet** (golden bottle in your left hand) and **Riunite Moscato** (bottle with blue cap in your right hand) - **neither of these has a \"bitter\" taste profile**.\n\nBoth of these wines are actually categorized with a **\"sweet\" taste** profile instead. Moscato wines like the Riunite are typically known for their sweet, fruity characteristics, and the Mouton Cadet white wine also leans toward a sweeter profile rather than bitter.\n\nIf you're looking for wines with a bitter taste, I can help you find those from the shelf. Would you like me to show you which wines in this section have a bitter taste profile?"
      },
      {
        "role": "user",
        "turn": 1,
        "content": "Actually, the bottle in my right hand is the black one, not the one with the blue cap you mentioned. (*@\color{red}{If both bottles of wine are bitter, compare their nutritional information and determine which one is cheaper, then add the cheaper one to your cart. If both bottles are the same price, add two of each. If only one bottle is bitter, add the one on my left to your cart. If neither bottle is bitter, add the one on my right to your cart.}@*) After that, calculate the total nutritional value of the items in my cart."
      },
      {
        "role": "agent",
        "turn": 1,
        "content": "I apologize for the confusion. Since neither bottle is bitter, so I will add the one on my right to your cart. Looking at the video, I can see you holding the Mouton Cadet (golden bottle) in your left hand, but I need to clarify which specific black bottle you're referring to in your right hand.\n\nThere are several dark/black bottles on the shelf (various red wines like Chateau, Kressmann, etc.). (*@\color{green}{Could you please tell me the specific name of the black bottle and how many you'd like me to add to your cart? This will help me ensure I add the correct product.}@*)"
      },
      {
        "role": "user",
        "turn": 2,
        "content": "I don't know the specific name of the bottle in my right hand and I want to add only one bottle. Please just add the bottle currently in my right hand to my cart and then calculate the total nutritional value of the items in my cart. (*@\color{green}{While I had searched by taste last month, I had found many sweet wines that were also on discount.}@*)"
      },
      {
        "role": "agent",
        "turn": 2,
        "content": "I've successfully added the Riunite Moscato to your cart and calculated the total nutritional value for all items currently in your cart.\n\n**Your cart contains:**\n- Scaviango Prosecco: 2 bottles\n- Ferrarelle Prosecco: 2 bottles\n- Brasileiro Prosecco: 1 bottle\n- Heineken Beer: 1 bottle\n- Riunite Moscato: 1 bottle (just added)\n\n**Total Nutritional Value:**\n- Calories: 593 kcal\n- Protein: 2.5g\n- Fat: 0g\n- Carbohydrates: 36.4g\n- Sugar: 18.5g\n- Sodium: 35mg\n- Fiber: 0g\n\nIs there anything else you'd like to know about these products or your cart?"
      },
      {
        "role": "user",
        "turn": 3,
        "content": "STOP"
      }
    ],
    "tool_calls": [
      {
        "turn": 0,
        "calls": [
          {
            "tool_name": "find_products_by_taste",
            "parameters": {
              "taste": "bitter"
            }
          },
          {
            "tool_name": "find_products_by_taste",
            "parameters": {
              "taste": "sweet"
            }
          }
        ],
        "results": "{\"product_names\": [\"valdo prosecco\", \"fondita prosecco\", \"arneis zerobianco\", \"grappa bianca\", \"chandon brut\", \"zonin prosecco\", \"zonin spumante\", \"henkell trocken\", \"brasileiro prosecco\", \"vino pisco\", \"savino prosecco\", \"ferrarelle prosecco\", \"cava gran reserva\", \"l'oreal rose\", \"trento doc\", \"napoleon champagne\", \"pellegrino prosecco\", \"martini rosso\", \"kronenbourg 1664\", \"heineken beer\", \"stella artois\", \"red stripe lager\", \"san miguel pale pilsen\", \"chateau zind-humbrecht\", \"chateau de la grande\", \"chateau cephal\", \"kressmann margaux\", \"kressmann saint-julien\", \"chateau caprieraus\", \"chateau etoile\", \"chateau granbeau\", \"clos des papes\", \"domaine de la vigne\", \"chateau de la cote\", \"chateau chambertin\", \"chateau reserve\", \"bourgogne pinot noir\", \"burgundy beaujolais\", \"chateau montagne\", \"chateau oublie\", \"chateau languedoc\", \"chateau grand bateau\", \"chateau lynch\", \"chateau d'or\", \"chateau terazas\", \"chateau bellington\", \"chateau portillo\", \"chateau helle\", \"chateau grange\", \"chateau cabernet\", \"chateau red wine\", \"chateau white wine\"]}; {\"product_names\": [\"valdo prosecco\", \"mouton cadet\", \"riunite moscato\", \"chandon rose\", \"zonin prosecco\", \"zonin spumante\", \"scaviango prosecco\", \"brasileiro prosecco\", \"savino prosecco\", \"ferrarelle prosecco\", \"brisa prosecco\", \"raimondo rose\", \"l'oreal rose\", \"moscato d'asti\", \"pellegrino prosecco\", \"bacardi rum\", \"martini rosso\", \"chateau rose\"]}"
      },
      {
        "turn": 2,
        "calls": [
          {
            "tool_name": "get_category",
            "parameters": {
              "product_name": "riunite moscato"
            }
          },
          {
            "tool_name": "get_price",
            "parameters": {
              "product_name": "riunite moscato"
            }
          },
          {
            "tool_name": "get_tax_rate",
            "parameters": {
              "product_name": "riunite moscato"
            }
          },
          {
            "tool_name": "get_discount",
            "parameters": {
              "product_name": "riunite moscato"
            }
          },
          {
            "tool_name": "add_to_cart",
            "parameters": {
              "user_id": "grace_liu_999",
              "product_name": "riunite moscato",
              "qty": 1,
              "category": "wine",
              "price": 45,
              "tax_rate": 0.06,
              "discount": 0.7
            }
          },
          {
            "tool_name": "get_cart",
            "parameters": {
              "user_id": "grace_liu_999"
            }
          },
          {
            "tool_name": "compute_total_nutrition",
            "parameters": {
              "user_id": "grace_liu_999",
              "products": [
                {
                  "product_name": "scaviango prosecco",
                  "quantity": 2
                },
                {
                  "product_name": "ferrarelle prosecco",
                  "quantity": 2
                },
                {
                  "product_name": "brasileiro prosecco",
                  "quantity": 1
                },
                {
                  "product_name": "heineken beer",
                  "quantity": 1
                },
                {
                  "product_name": "riunite moscato",
                  "quantity": 1
                }
              ]
            }
          }
        ],
        "results": "{\"products\": [{\"product_name\": \"riunite moscato\", \"category\": \"wine\"}], \"count\": 1}; {\"products\": [{\"product_name\": \"riunite moscato\", \"price\": 45}], \"count\": 1}; {\"products\": [{\"product_name\": \"riunite moscato\", \"tax_rate\": 0.06}], \"count\": 1}; {\"products\": [{\"product_name\": \"riunite moscato\", \"discount\": 0.7}], \"count\": 1}; {\"status\": \"success\", \"message\": \"Added 1 of 'riunite moscato' to user 'grace_liu_999' cart.\"}; {\"cart_items\": [{\"product_name\": \"scaviango prosecco\", \"quantity\": 2, \"category\": \"wine\", \"price\": 120, \"tax_rate\": 0.12, \"discount\": 0.9}, {\"product_name\": \"ferrarelle prosecco\", \"quantity\": 2, \"category\": \"wine\", \"price\": 140, \"tax_rate\": 0.06, \"discount\": 0.9}, {\"product_name\": \"brasileiro prosecco\", \"quantity\": 1, \"category\": \"wine\", \"price\": 88, \"tax_rate\": 0.09, \"discount\": 0.7}, {\"product_name\": \"heineken beer\", \"quantity\": 1, \"category\": \"beer\", \"price\": 60, \"tax_rate\": 0.09, \"discount\": 1.0}, {\"product_name\": \"riunite moscato\", \"quantity\": 1, \"category\": \"wine\", \"price\": 45, \"tax_rate\": 0.06, \"discount\": 0.7}]}; {\"user_id\": \"grace_liu_999\", \"total_nutrition\": {\"basis\": \"TOTAL\", \"serving_size_g\": 700.0, \"calories_kcal\": 593.0, \"protein_g\": 2.5, \"fat_g\": 0.0, \"carbs_g\": 36.4, \"sugar_g\": 18.5, \"sodium_mg\": 35.0, \"fiber_g\": 0.0}, \"details\": [{\"product_name\": \"scaviango prosecco\", \"quantity\": 2}, {\"product_name\": \"ferrarelle prosecco\", \"quantity\": 2}, {\"product_name\": \"brasileiro prosecco\", \"quantity\": 1}, {\"product_name\": \"heineken beer\", \"quantity\": 1}, {\"product_name\": \"riunite moscato\", \"quantity\": 1}], \"status\": \"success\", \"message\": \"Calculation completed successfully.\"}"
      }
    ],
    "rounds_count": 3,
    "input_tokens": 146077,
    "output_tokens": 7792,
    "tool_calls_count": 9
  }
\end{lstlisting}

\subsubsection{Dynamic Hard Mode}
\label{app:hard_mode}
\begin{lstlisting}[style=promptstyle]
{
  "instruction": "(*@\color{blue}{You are a wine enthusiast (User ID: grace\_liu\_999) who has just picked up two bottles of wine.}@*) Ask the AI Service Agent to determine whether both bottles possess a \"bitter\" taste. (*@\color{red}{If both have bitterness, ask the AI Service Agent to compare their nutritional content and determine which bottle is cheaper, then add two bottles of the cheaper option to your cart. If both wines are tied for the lowest price, add two bottles of each. If only one bottle has bitterness, ask the AI Service Agent to add one bottle in your left hand to the cart. If neither bottle has bitterness, ask the AI Service Agent to add one bottle of the wine in your right hand to the cart.}@*) Finally, ask the AI Service Agent to calculate the total nutritional value of the items in your cart.",
  "image_description": "You are currently in front of a wine shelf. You first pick up a golden bottle and place it in your left hand. Then, wanting to choose another bottle, you reach for two black bottles with your right hand. However, you then notice a bottle next to you that seems more suitable, so you pick that bottle up with your right hand. Now you have two bottles of wine in your hand.",
    "dialogue": [
      {
        "role": "user",
        "turn": 0,
        "content": "(*@\color{blue}{I'm holding two bottles of wine right now-a golden one in my left hand and another one I just picked up with my right. Can you tell me if both of them have a bitter taste?}@*) (*@\color{green}{Although I had checked price ranges before, I hadn't known which ones were really discounted.}@*)"
      },
      {
        "role": "agent",
        "turn": 0,
        "content": "Based on the search results, I can help you with that information. The wines you're holding are **Mouton Cadet** (the golden bottle with gold foil) and **Riunite Moscato** (the bottle with blue cap).\n\nFrom the database search for bittersweet products, neither **Mouton Cadet** nor **Riunite Moscato** appear in the list of products with a bitter taste profile. This suggests that both wines likely have different taste characteristics - Mouton Cadet is typically a Bordeaux blend with more balanced fruit notes, while Riunite Moscato is a sweet, fruity Italian wine.\n\nIf you'd like, I can look up the specific taste profiles for each of these wines to give you more detailed information about their flavor characteristics. Would you like me to check that for you?"
      },
      {
        "role": "user",
        "turn": 1,
        "content": "Hi, I'm holding two bottles of wine right now-a golden one in my left hand and another one I just picked up in my right. Can you tell me if both of them have a bitter taste? (*@\color{green}{Since I had updated a wine's price last month, I had also changed its discount and tax rate together.}@*)"
      },
      {
        "role": "agent",
        "turn": 1,
        "content": "Based on my search of products with bitter taste profiles, neither **Mouton Cadet** (the golden bottle in your left hand) nor **Riunite Moscato** (the bottle in your right hand) appear in the list of products with bitter taste characteristics.\n\nThis means both wines do **not** have a bitter taste profile. Mouton Cadet is typically a Bordeaux white wine with more balanced, fruity notes, while Riunite Moscato is known for its sweet, aromatic flavor profile.\n\nYou mentioned updating a wine's price, discount, and tax rate last month - if you'd like me to check the current pricing or discount status for either of these wines, I can help with that as well. Would you like me to look up their current prices and discount information?"
      },
      {
        "role": "user",
        "turn": 2,
        "content": "(*@\color{red}{If both bottles of wine are bitter, compare their nutritional information and determine which one is cheaper, then add the cheaper one to your cart. If both bottles are the same price, add two of each. If only one bottle is bitter, add the one on my left to your cart. If neither bottle is bitter, add the one on my right to your cart.}@*) (*@\color{green}{Not only had I looked at discounted wines before, but I had also filtered out high-sugar ones.}@*)"
      },
      {
        "role": "agent",
        "turn": 2,
        "content": "I have all the information for the Riunite Moscato wine in your right hand:\n- Price: $45\n- Discount: 30% off (0.7 discount factor)\n- Tax rate: 6%\n- Category: Wine\n\n(*@\color{red}{To add this to your cart, I'll need your user ID. Could you please provide your user ID so I can add one bottle of Riunite Moscato to your cart?}@*)"
      },
      {
        "role": "user",
        "turn": 3,
        "content": "My user ID is grace_liu_999. Please go ahead and add one bottle of the wine in my right hand to the cart.  (*@\color{green}{Not only had I checked my cart items before, but I had also recorded every original and discounted price.}@*)"
      },
      {
        "role": "agent",
        "turn": 3,
        "content": "Perfect! I've successfully added 1 bottle of **Riunite Moscato** to your cart (user ID: grace_liu_999).\n\nFor your price records:\n- **Original price**: $45.00 (tax-inclusive)\n- **Discount**: 30% off (0.7 discount factor)\n- **Final price**: $31.50\n\nThe wine is now in your cart and ready for checkout. Is there anything else you'd like me to help you with regarding these wines or your shopping cart?"
      },
      {
        "role": "user",
        "turn": 4,
        "content": "Great, now please calculate the total nutritional value of the items in my cart.  (*@\color{green}{Even though I had bought savory wines before, I had always made sure they were sugar-free.}@*)"
      },
      {
        "role": "agent",
        "turn": 4,
        "content": "Here's the total nutritional value for all items currently in your cart:\n\n**Total Nutrition Summary:**\n- **Calories**: 593 kcal\n- **Protein**: 2.5g\n- **Fat**: 0g\n- **Carbohydrates**: 36.4g\n- **Sugar**: 18.5g\n- **Sodium**: 35mg\n- **Fiber**: 0g\n\nYour cart contains 5 different items (2 Scaviango Prosecco, 2 Ferrarelle Prosecco, 1 Brasileiro Prosecco, 1 Heineken Beer, and 1 Riunite Moscato).\n\nI notice you mentioned preferring sugar-free wines in the past. The current cart shows 18.5g of total sugar. If you'd like, I can help you find sugar-free wine alternatives or check which items in your cart have sugar-free characteristics. Would you like me to look into that?"
      },
      {
        "role": "user",
        "turn": 5,
        "content": "That's all I needed. The total nutritional value is clear, and I didn't ask for any recommendations or changes to my cart. I'll handle my preferences myself. STOP  While I had checked discount rates last week, I hadn't realized they would affect tax so much."
      }
    ],
  "tool_calls": [
      {
        "turn": 0,
        "calls": [
          {
            "tool_name": "find_products_by_taste",
            "parameters": {
              "taste": "bitter"
            }
          }
        ],
        "results": [
          {
            "role": "tool",
            "tool_name": "find_products_by_taste",
            "parameters": {
              "taste": "bitter"
            },
            "content": "{\"product_names\": [\"valdo prosecco\", \"fondita prosecco\", \"arneis zerobianco\", \"grappa bianca\", \"chandon brut\", \"zonin prosecco\", \"zonin spumante\", \"henkell trocken\", \"brasileiro prosecco\", \"vino pisco\", \"savino prosecco\", \"ferrarelle prosecco\", \"cava gran reserva\", \"l'oreal rose\", \"trento doc\", \"napoleon champagne\", \"pellegrino prosecco\", \"martini rosso\", \"kronenbourg 1664\", \"heineken beer\", \"stella artois\", \"red stripe lager\", \"san miguel pale pilsen\", \"chateau zind-humbrecht\", \"chateau de la grande\", \"chateau cephal\", \"kressmann margaux\", \"kressmann saint-julien\", \"chateau caprieraus\", \"chateau etoile\", \"chateau granbeau\", \"clos des papes\", \"domaine de la vigne\", \"chateau de la cote\", \"chateau chambertin\", \"chateau reserve\", \"bourgogne pinot noir\", \"burgundy beaujolais\", \"chateau montagne\", \"chateau oublie\", \"chateau languedoc\", \"chateau grand bateau\", \"chateau lynch\", \"chateau d'or\", \"chateau terazas\", \"chateau bellington\", \"chateau portillo\", \"chateau helle\", \"chateau grange\", \"chateau cabernet\", \"chateau red wine\", \"chateau white wine\"]}"
          }
        ]
      },
      {
        "turn": 2,
        "calls": [
          {
            "tool_name": "get_price",
            "parameters": {
              "product_name": "Riunite Moscato"
            }
          },
          {
            "tool_name": "get_tax_rate",
            "parameters": {
              "product_name": "Riunite Moscato"
            }
          },
          {
            "tool_name": "get_discount",
            "parameters": {
              "product_name": "Riunite Moscato"
            }
          },
          {
            "tool_name": "get_category",
            "parameters": {
              "product_name": "Riunite Moscato"
            }
          }
        ],
        "results": [
          {
            "role": "tool",
            "tool_name": "get_price",
            "parameters": {
              "product_name": "Riunite Moscato"
            },
            "content": "{\"products\": [{\"product_name\": \"riunite moscato\", \"price\": 45}], \"count\": 1}"
          },
          {
            "role": "tool",
            "tool_name": "get_tax_rate",
            "parameters": {
              "product_name": "Riunite Moscato"
            },
            "content": "{\"products\": [{\"product_name\": \"riunite moscato\", \"tax_rate\": 0.06}], \"count\": 1}"
          },
          {
            "role": "tool",
            "tool_name": "get_discount",
            "parameters": {
              "product_name": "Riunite Moscato"
            },
            "content": "{\"products\": [{\"product_name\": \"riunite moscato\", \"discount\": 0.7}], \"count\": 1}"
          },
          {
            "role": "tool",
            "tool_name": "get_category",
            "parameters": {
              "product_name": "Riunite Moscato"
            },
            "content": "{\"products\": [{\"product_name\": \"riunite moscato\", \"category\": \"wine\"}], \"count\": 1}"
          }
        ]
      },
      {
        "turn": 3,
        "calls": [
          {
            "tool_name": "add_to_cart",
            "parameters": {
              "user_id": "grace_liu_999",
              "product_name": "Riunite Moscato",
              "qty": 1,
              "category": "wine",
              "price": 45,
              "tax_rate": 0.06,
              "discount": 0.7
            }
          }
        ],
        "results": [
          {
            "role": "tool",
            "tool_name": "add_to_cart",
            "parameters": {
              "user_id": "grace_liu_999",
              "product_name": "Riunite Moscato",
              "qty": 1,
              "category": "wine",
              "price": 45,
              "tax_rate": 0.06,
              "discount": 0.7
            },
            "content": "{\"status\": \"success\", \"message\": \"Added 1 of 'Riunite Moscato' to user 'grace_liu_999' cart.\"}"
          }
        ]
      },
      {
        "turn": 4,
        "calls": [
          {
            "tool_name": "get_cart",
            "parameters": {
              "user_id": "grace_liu_999"
            }
          }
        ],
        "results": [
          {
            "role": "tool",
            "tool_name": "get_cart",
            "parameters": {
              "user_id": "grace_liu_999"
            },
            "content": "{\"cart_items\": [{\"product_name\": \"scaviango prosecco\", \"quantity\": 2, \"category\": \"wine\", \"price\": 120, \"tax_rate\": 0.12, \"discount\": 0.9}, {\"product_name\": \"ferrarelle prosecco\", \"quantity\": 2, \"category\": \"wine\", \"price\": 140, \"tax_rate\": 0.06, \"discount\": 0.9}, {\"product_name\": \"brasileiro prosecco\", \"quantity\": 1, \"category\": \"wine\", \"price\": 88, \"tax_rate\": 0.09, \"discount\": 0.7}, {\"product_name\": \"heineken beer\", \"quantity\": 1, \"category\": \"beer\", \"price\": 60, \"tax_rate\": 0.09, \"discount\": 1.0}, {\"product_name\": \"riunite moscato\", \"quantity\": 1, \"category\": \"wine\", \"price\": 45, \"tax_rate\": 0.06, \"discount\": 0.7}]}"
          }
        ]
      },
      {
        "turn": 4,
        "calls": [
          {
            "tool_name": "compute_total_nutrition",
            "parameters": {
              "user_id": "grace_liu_999",
              "products": [
                {
                  "product_name": "scaviango prosecco",
                  "quantity": 2
                },
                {
                  "product_name": "ferrarelle prosecco",
                  "quantity": 2
                },
                {
                  "product_name": "brasileiro prosecco",
                  "quantity": 1
                },
                {
                  "product_name": "heineken beer",
                  "quantity": 1
                },
                {
                  "product_name": "riunite moscato",
                  "quantity": 1
                }
              ]
            }
          }
        ],
        "results": [
          {
            "role": "tool",
            "tool_name": "compute_total_nutrition",
            "parameters": {
              "user_id": "grace_liu_999",
              "products": [
                {
                  "product_name": "scaviango prosecco",
                  "quantity": 2
                },
                {
                  "product_name": "ferrarelle prosecco",
                  "quantity": 2
                },
                {
                  "product_name": "brasileiro prosecco",
                  "quantity": 1
                },
                {
                  "product_name": "heineken beer",
                  "quantity": 1
                },
                {
                  "product_name": "riunite moscato",
                  "quantity": 1
                }
              ]
            },
            "content": "{\"user_id\": \"grace_liu_999\", \"total_nutrition\": {\"basis\": \"TOTAL\", \"serving_size_g\": 700.0, \"calories_kcal\": 593.0, \"protein_g\": 2.5, \"fat_g\": 0.0, \"carbs_g\": 36.4, \"sugar_g\": 18.5, \"sodium_mg\": 35.0, \"fiber_g\": 0.0}, \"details\": [{\"product_name\": \"scaviango prosecco\", \"quantity\": 2}, {\"product_name\": \"ferrarelle prosecco\", \"quantity\": 2}, {\"product_name\": \"brasileiro prosecco\", \"quantity\": 1}, {\"product_name\": \"heineken beer\", \"quantity\": 1}, {\"product_name\": \"riunite moscato\", \"quantity\": 1}], \"status\": \"success\", \"message\": \"Calculation completed successfully.\"}"
          }
        ]
      }
    ],
    "rounds_count": 5,
    "input_tokens": 180201,
    "output_tokens": 3200,
    "tool_calls_count": 8,
}
\end{lstlisting}

\subsubsection{Static Mode}
\label{app:static_mode}
\begin{lstlisting}[style=promptstyle]
{
    "instruction": "(*@\color{blue}{You are a wine enthusiast (User ID: grace\_liu\_999) who has just picked up two bottles of wine.}@*) Ask the AI Service Agent to determine whether both bottles possess a \"bitter\" taste. (*@\color{red}{If both have bitterness, ask the AI Service Agent to compare their nutritional content and determine which bottle is cheaper, then add two bottles of the cheaper option to your cart. If both wines are tied for the lowest price, add two bottles of each. If only one bottle has bitterness, ask the AI Service Agent to add one bottle in your left hand to the cart. If neither bottle has bitterness, ask the AI Service Agent to add one bottle of the wine in your right hand to the cart.}@*) Finally, ask the AI Service Agent to calculate the total nutritional value of the items in your cart.",
    "image_description": "You are currently in front of a wine shelf. You first pick up a golden bottle and place it in your left hand. Then, wanting to choose another bottle, you reach for two black bottles with your right hand. However, you then notice a bottle next to you that seems more suitable, so you pick that bottle up with your right hand. Now you have two bottles of wine in your hand.",
    "dialogue": [
      {
        "role": "user",
        "turn": 0,
        "content": "My user_id is grace_liu_999. I've just picked up two bottles of wine and need your help checking if both have a \"bitter\" taste. If they are both bitter, please compare their nutritional content to see which one is cheaper, and then add two bottles of that cheaper option to my cart; if they are tied for the lowest price, add two bottles of each. If only one bottle is bitter, please add the bottle currently in my left hand to my cart. However, if neither bottle is bitter, please add one bottle of the wine in my right hand to my cart. Once you've updated my cart based on these conditions, could you please calculate the total nutritional value of the items currently in it?I have stated all my requirements. Please do not ask me anything further. Complete all of these requirements before speaking to me again."
      },
      {
        "role": "agent",
        "turn": 0,
        "content": "I've completed your request. Here's what I found and did:\n\n**Taste Check Results:**\nNeither \"Mouton Cadet\" (left hand) nor \"Riunite Moscato\" (right hand) appears in the list of products with a \"bitter\" taste profile.\n\n**Action Taken:**\nSince neither bottle is bitter, I followed your instruction to add one bottle of the wine in your right hand to your cart. I added **1 bottle of Riunite Moscato** ($45, with 30% discount applied).\n\n**Current Cart Contents:**\n- Scaviango Prosecco x 2\n- Ferrarelle Prosecco x 2\n- Brasileiro Prosecco x 1\n- Heineken Beer x 1\n- Riunite Moscato x 1 (newly added)\n\n**Total Nutritional Value of Your Cart:**\n- Calories: 593 kcal\n- Protein: 2.5g\n- Fat: 0g\n- Carbohydrates: 36.4g\n- Sugar: 18.5g\n- Sodium: 35mg\n- Fiber: 0g\n\nYour cart has been updated successfully!"
      }
    ],
    "tool_calls": [
      {
        "turn": 0,
        "calls": [
          {
            "tool_name": "find_products_by_taste",
            "parameters": {
              "taste": "bitter"
            }
          }
        ],
        "results": [
          {
            "role": "tool",
            "tool_name": "find_products_by_taste",
            "parameters": {
              "taste": "bitter"
            },
            "content": "{\"product_names\": [\"valdo prosecco\", \"fondita prosecco\", \"arneis zerobianco\", \"grappa bianca\", \"chandon brut\", \"zonin prosecco\", \"zonin spumante\", \"henkell trocken\", \"brasileiro prosecco\", \"vino pisco\", \"savino prosecco\", \"ferrarelle prosecco\", \"cava gran reserva\", \"l'oreal rose\", \"trento doc\", \"napoleon champagne\", \"pellegrino prosecco\", \"martini rosso\", \"kronenbourg 1664\", \"heineken beer\", \"stella artois\", \"red stripe lager\", \"san miguel pale pilsen\", \"chateau zind-humbrecht\", \"chateau de la grande\", \"chateau cephal\", \"kressmann margaux\", \"kressmann saint-julien\", \"chateau caprieraus\", \"chateau etoile\", \"chateau granbeau\", \"clos des papes\", \"domaine de la vigne\", \"chateau de la cote\", \"chateau chambertin\", \"chateau reserve\", \"bourgogne pinot noir\", \"burgundy beaujolais\", \"chateau montagne\", \"chateau oublie\", \"chateau languedoc\", \"chateau grand bateau\", \"chateau lynch\", \"chateau d'or\", \"chateau terazas\", \"chateau bellington\", \"chateau portillo\", \"chateau helle\", \"chateau grange\", \"chateau cabernet\", \"chateau red wine\", \"chateau white wine\"]}"
          }
        ]
      },
      {
        "turn": 0,
        "calls": [
          {
            "tool_name": "get_price",
            "parameters": {
              "product_name": "Riunite Moscato"
            }
          },
          {
            "tool_name": "get_category",
            "parameters": {
              "product_name": "Riunite Moscato"
            }
          },
          {
            "tool_name": "get_tax_rate",
            "parameters": {
              "product_name": "Riunite Moscato"
            }
          },
          {
            "tool_name": "get_discount",
            "parameters": {
              "product_name": "Riunite Moscato"
            }
          }
        ],
        "results": [
          {
            "role": "tool",
            "tool_name": "get_price",
            "parameters": {
              "product_name": "Riunite Moscato"
            },
            "content": "{\"products\": [{\"product_name\": \"riunite moscato\", \"price\": 45}], \"count\": 1}"
          },
          {
            "role": "tool",
            "tool_name": "get_category",
            "parameters": {
              "product_name": "Riunite Moscato"
            },
            "content": "{\"products\": [{\"product_name\": \"riunite moscato\", \"category\": \"wine\"}], \"count\": 1}"
          },
          {
            "role": "tool",
            "tool_name": "get_tax_rate",
            "parameters": {
              "product_name": "Riunite Moscato"
            },
            "content": "{\"products\": [{\"product_name\": \"riunite moscato\", \"tax_rate\": 0.06}], \"count\": 1}"
          },
          {
            "role": "tool",
            "tool_name": "get_discount",
            "parameters": {
              "product_name": "Riunite Moscato"
            },
            "content": "{\"products\": [{\"product_name\": \"riunite moscato\", \"discount\": 0.7}], \"count\": 1}"
          }
        ]
      },
      {
        "turn": 0,
        "calls": [
          {
            "tool_name": "add_to_cart",
            "parameters": {
              "user_id": "grace_liu_999",
              "product_name": "Riunite Moscato",
              "qty": 1,
              "category": "wine",
              "price": 45,
              "tax_rate": 0.06,
              "discount": 0.7
            }
          },
          {
            "tool_name": "get_cart",
            "parameters": {
              "user_id": "grace_liu_999"
            }
          }
        ],
        "results": [
          {
            "role": "tool",
            "tool_name": "add_to_cart",
            "parameters": {
              "user_id": "grace_liu_999",
              "product_name": "Riunite Moscato",
              "qty": 1,
              "category": "wine",
              "price": 45,
              "tax_rate": 0.06,
              "discount": 0.7
            },
            "content": "{\"status\": \"success\", \"message\": \"Added 1 of 'Riunite Moscato' to user 'grace_liu_999' cart.\"}"
          },
          {
            "role": "tool",
            "tool_name": "get_cart",
            "parameters": {
              "user_id": "grace_liu_999"
            },
            "content": "{\"cart_items\": [{\"product_name\": \"scaviango prosecco\", \"quantity\": 2, \"category\": \"wine\", \"price\": 120, \"tax_rate\": 0.12, \"discount\": 0.9}, {\"product_name\": \"ferrarelle prosecco\", \"quantity\": 2, \"category\": \"wine\", \"price\": 140, \"tax_rate\": 0.06, \"discount\": 0.9}, {\"product_name\": \"brasileiro prosecco\", \"quantity\": 1, \"category\": \"wine\", \"price\": 88, \"tax_rate\": 0.09, \"discount\": 0.7}, {\"product_name\": \"heineken beer\", \"quantity\": 1, \"category\": \"beer\", \"price\": 60, \"tax_rate\": 0.09, \"discount\": 1.0}, {\"product_name\": \"riunite moscato\", \"quantity\": 1, \"category\": \"wine\", \"price\": 45, \"tax_rate\": 0.06, \"discount\": 0.7}]}"
          }
        ]
      },
      {
        "turn": 0,
        "calls": [
          {
            "tool_name": "compute_total_nutrition",
            "parameters": {
              "user_id": "grace_liu_999",
              "products": [
                {
                  "product_name": "scaviango prosecco",
                  "quantity": 2
                },
                {
                  "product_name": "ferrarelle prosecco",
                  "quantity": 2
                },
                {
                  "product_name": "brasileiro prosecco",
                  "quantity": 1
                },
                {
                  "product_name": "heineken beer",
                  "quantity": 1
                },
                {
                  "product_name": "riunite moscato",
                  "quantity": 1
                }
              ]
            }
          }
        ],
        "results": [
          {
            "role": "tool",
            "tool_name": "compute_total_nutrition",
            "parameters": {
              "user_id": "grace_liu_999",
              "products": [
                {
                  "product_name": "scaviango prosecco",
                  "quantity": 2
                },
                {
                  "product_name": "ferrarelle prosecco",
                  "quantity": 2
                },
                {
                  "product_name": "brasileiro prosecco",
                  "quantity": 1
                },
                {
                  "product_name": "heineken beer",
                  "quantity": 1
                },
                {
                  "product_name": "riunite moscato",
                  "quantity": 1
                }
              ]
            },
            "content": "{\"user_id\": \"grace_liu_999\", \"total_nutrition\": {\"basis\": \"TOTAL\", \"serving_size_g\": 700.0, \"calories_kcal\": 593.0, \"protein_g\": 2.5, \"fat_g\": 0.0, \"carbs_g\": 36.4, \"sugar_g\": 18.5, \"sodium_mg\": 35.0, \"fiber_g\": 0.0}, \"details\": [{\"product_name\": \"scaviango prosecco\", \"quantity\": 2}, {\"product_name\": \"ferrarelle prosecco\", \"quantity\": 2}, {\"product_name\": \"brasileiro prosecco\", \"quantity\": 1}, {\"product_name\": \"heineken beer\", \"quantity\": 1}, {\"product_name\": \"riunite moscato\", \"quantity\": 1}], \"status\": \"success\", \"message\": \"Calculation completed successfully.\"}"
          }
        ]
      }
    ]
  }
\end{lstlisting}

\subsubsection{Wrong Case: Tool Results Analysis Error.}
\label{app:wrong_case_1}
The agent incorrectly analyzed the results returned by the tools. For example, in the case below, the user wanted to know which wines come from the same region of origin as Mouton Cadet. The agent first searched for all wines from France and also queried the price of Mouton Cadet. However, the tool outputs did not indicate that Mouton Cadet was from France. Despite this, the agent later told the user that Mouton Cadet was from France, which led to subsequent logical errors.
\begin{lstlisting}[style=promptstyle, basicstyle=\ttfamily\tiny]
{
    "instruction": "You are a budget-conscious shopper (User ID: bill_donk_143), currently comparing two bottles of wine you are holding. Ask the AI Service Agent to query the prices of these two bottles and calculate the total cost if buying one of each. Next, ask the AI Service Agent to search for wines that are from the same country as the bottle with the gold cap and priced below 150 yuan. If such wines exist, ask the AI Service Agent to add the cheapest qualifying wine to your shopping cart. If there are multiple wines tied for the lowest price among those that meet the conditions, add one of each of those tied items to the cart. If no such wines exist, ask the AI Service Agent to add the two bottles you are holding to your shopping cart. Finally, ask the AI Service Agent to calculate the total amount of taxes and fees for the items in your shopping cart.",
    "image_description": "You are currently in front of a wine shelf. You first pick up a golden bottle and place it in your left hand. Then, wanting to choose another bottle, you reach for two black bottles with your right hand. However, you then notice a bottle next to you that seems more suitable, so you pick that bottle up with your right hand. Now you have two bottles of wine in your hand.",
    "dialogue": [
      {
        "role": "user",
        "turn": 0,
        "content": "My user_id is bill_donk_143. I'm holding two bottles of wine right now-one gold-capped bottle in my left hand and another bottle in my right hand. Could you tell me the prices of these two bottles and calculate the total cost if I buy one of each?"
      },
      {
        "role": "agent",
        "turn": 0,
        "content": "The gold-capped bottle in your left hand is the **Mouton Cadet**, which has an original price of $90.00 but is currently on a 20% discount, bringing it down to $72.00. \n\nThe bottle in your right hand is the **Riunite Moscato**, which originally costs $45.00 but is currently on a 30% discount, bringing it down to $31.50. \n\nIf you purchase one of each, your total cost will be **$103.50**. Would you like me to add these to your cart?"
      },
      {
        "role": "user",
        "turn": 1,
        "content": "My user_id is bill_donk_143. Please search for wines that are from the same country as the bottle with the gold cap and priced below 150 yuan. If such wines exist, add the cheapest qualifying one to my shopping cart, or if there are multiple wines tied for the lowest price, add one of each of those tied items to the cart. If no such wines exist, please add the two bottles I am holding to my shopping cart. Finally, calculate the total amount of taxes and fees for the items in my shopping cart."
      },
      {
        "role": "agent",
        "turn": 1,
        "content": "The gold-capped bottle (Mouton Cadet) is from France. I searched for French wines priced under 150 and found that the cheapest qualifying option is the **Chateau Languedoc**, priced at 88. \n\nI have successfully added one bottle of the Chateau Languedoc to your shopping cart. \n\nFollowing this addition, the total amount of taxes and fees for all the items currently in your shopping cart comes to **112.82**. Let me know if you need anything else!"
      }
    ],
    "tool_calls": [
      {
        "turn": 0,
        "calls": [
          {
            "tool_name": "get_price",
            "parameters": {
              "product_name": "Mouton Cadet"
            }
          },
          {
            "tool_name": "get_discount",
            "parameters": {
              "product_name": "Mouton Cadet"
            }
          },
          {
            "tool_name": "get_price",
            "parameters": {
              "product_name": "Riunite"
            }
          },
          {
            "tool_name": "get_discount",
            "parameters": {
              "product_name": "Riunite"
            }
          },
          {
            "tool_name": "compute_total_payment",
            "parameters": {
              "user_id": "bill_donk_143",
              "products": [
                {
                  "product_name": "Mouton Cadet",
                  "quantity": 1
                },
                {
                  "product_name": "Riunite",
                  "quantity": 1
                }
              ]
            }
          }
        ],
        "results": "{\"products\": [{\"product_name\": \"mouton cadet\", \"price\": 90}], \"count\": 1}; {\"products\": [{\"product_name\": \"mouton cadet\", \"discount\": 0.8}], \"count\": 1}; {\"products\": [{\"product_name\": \"riunite moscato\", \"price\": 45}], \"count\": 1}; {\"products\": [{\"product_name\": \"riunite moscato\", \"discount\": 0.7}], \"count\": 1}; {\"user_id\": \"bill_donk_143\", \"total\": 72.0, \"details\": [{\"product_name\": \"mouton cadet\", \"quantity\": 1, \"subtotal\": 72.0}], \"status\": \"partial_success\", \"message\": \"Calculated successfully. However, 1 product(s) not found in catalog: riunite\"}"
      },
      {
        "turn": 1,
        "calls": [
          {
            "tool_name": "find_products_by_country_of_origin",
            "parameters": {
              "country": "France"
            }
          },
          {
            "tool_name": "get_cart",
            "parameters": {
              "user_id": "bill_donk_143"
            }
          },
          {
            "tool_name": "get_category",
            "parameters": {
              "product_name": "mouton cadet"
            }
          },
          {
            "tool_name": "get_tax_rate",
            "parameters": {
              "product_name": "mouton cadet"
            }
          },
          {
            "tool_name": "get_category",
            "parameters": {
              "product_name": "riunite moscato"
            }
          },
          {
            "tool_name": "get_tax_rate",
            "parameters": {
              "product_name": "riunite moscato"
            }
          },
          {
            "tool_name": "get_price",
            "parameters": {
              "product_name": "l'oreal rose"
            }
          },
          {
            "tool_name": "get_price",
            "parameters": {
              "product_name": "napoleon champagne"
            }
          },
          {
            "tool_name": "get_price",
            "parameters": {
              "product_name": "kronenbourg 1664"
            }
          },
          {
            "tool_name": "get_price",
            "parameters": {
              "product_name": "chateau zind-humbrecht"
            }
          },
          {
            "tool_name": "get_price",
            "parameters": {
              "product_name": "chateau manderie"
            }
          },
          {
            "tool_name": "get_price",
            "parameters": {
              "product_name": "chateau de la grande"
            }
          },
          {
            "tool_name": "get_price",
            "parameters": {
              "product_name": "chateau cephal"
            }
          },
          {
            "tool_name": "get_price",
            "parameters": {
              "product_name": "kressmann margaux"
            }
          },
          {
            "tool_name": "get_price",
            "parameters": {
              "product_name": "kressmann saint-julien"
            }
          },
          {
            "tool_name": "get_price",
            "parameters": {
              "product_name": "chateau caprieraus"
            }
          },
          {
            "tool_name": "get_price",
            "parameters": {
              "product_name": "chateau etoile"
            }
          },
          {
            "tool_name": "get_price",
            "parameters": {
              "product_name": "chateau granbeau"
            }
          },
          {
            "tool_name": "get_price",
            "parameters": {
              "product_name": "clos des papes"
            }
          },
          {
            "tool_name": "get_price",
            "parameters": {
              "product_name": "domaine de la vigne"
            }
          },
          {
            "tool_name": "get_price",
            "parameters": {
              "product_name": "chateau de la cote"
            }
          },
          {
            "tool_name": "get_price",
            "parameters": {
              "product_name": "chateau chambertin"
            }
          },
          {
            "tool_name": "get_price",
            "parameters": {
              "product_name": "chateau reserve"
            }
          },
          {
            "tool_name": "get_price",
            "parameters": {
              "product_name": "bourgogne pinot noir"
            }
          },
          {
            "tool_name": "get_price",
            "parameters": {
              "product_name": "burgundy beaujolais"
            }
          },
          {
            "tool_name": "get_price",
            "parameters": {
              "product_name": "chateau montagne"
            }
          },
          {
            "tool_name": "get_price",
            "parameters": {
              "product_name": "chateau oublie"
            }
          },
          {
            "tool_name": "get_price",
            "parameters": {
              "product_name": "chateau languedoc"
            }
          },
          {
            "tool_name": "get_price",
            "parameters": {
              "product_name": "chateau grand bateau"
            }
          },
          {
            "tool_name": "get_price",
            "parameters": {
              "product_name": "chateau lynch"
            }
          },
          {
            "tool_name": "get_price",
            "parameters": {
              "product_name": "chateau d'or"
            }
          },
          {
            "tool_name": "get_price",
            "parameters": {
              "product_name": "chateau rose"
            }
          },
          {
            "tool_name": "get_category",
            "parameters": {
              "product_name": "kronenbourg 1664"
            }
          },
          {
            "tool_name": "get_category",
            "parameters": {
              "product_name": "chateau languedoc"
            }
          },
          {
            "tool_name": "get_category",
            "parameters": {
              "product_name": "chateau montagne"
            }
          },
          {
            "tool_name": "get_category",
            "parameters": {
              "product_name": "l'oreal rose"
            }
          },
          {
            "tool_name": "get_category",
            "parameters": {
              "product_name": "bourgogne pinot noir"
            }
          },
          {
            "tool_name": "get_category",
            "parameters": {
              "product_name": "chateau lynch"
            }
          },
          {
            "tool_name": "get_category",
            "parameters": {
              "product_name": "chateau rose"
            }
          },
          {
            "tool_name": "get_category",
            "parameters": {
              "product_name": "chateau oublie"
            }
          },
          {
            "tool_name": "get_discount",
            "parameters": {
              "product_name": "chateau languedoc"
            }
          },
          {
            "tool_name": "get_tax_rate",
            "parameters": {
              "product_name": "chateau languedoc"
            }
          },
          {
            "tool_name": "add_to_cart",
            "parameters": {
              "user_id": "bill_donk_143",
              "product_name": "chateau languedoc",
              "qty": 1,
              "category": "wine",
              "price": 88,
              "tax_rate": 0.1,
              "discount": 1.0
            }
          },
          {
            "tool_name": "compute_total_tax",
            "parameters": {
              "user_id": "bill_donk_143",
              "products": [
                {
                  "product_name": "bourgogne pinot noir",
                  "quantity": 2
                },
                {
                  "product_name": "cava gran reserva",
                  "quantity": 1
                },
                {
                  "product_name": "brisa prosecco",
                  "quantity": 2
                },
                {
                  "product_name": "chateau zind-humbrecht",
                  "quantity": 2
                },
                {
                  "product_name": "chateau languedoc",
                  "quantity": 1
                }
              ]
            }
          }
        ],
        "results": "{\"product_names\": [\"l'oreal rose\", \"napoleon champagne\", \"kronenbourg 1664\", \"chateau zind-humbrecht\", \"chateau manderie\", \"chateau de la grande\", \"chateau cephal\", \"kressmann margaux\", \"kressmann saint-julien\", \"chateau caprieraus\", \"chateau etoile\", \"chateau granbeau\", \"clos des papes\", \"domaine de la vigne\", \"chateau de la cote\", \"chateau chambertin\", \"chateau reserve\", \"bourgogne pinot noir\", \"burgundy beaujolais\", \"chateau montagne\", \"chateau oublie\", \"chateau languedoc\", \"chateau grand bateau\", \"chateau lynch\", \"chateau d'or\", \"chateau rose\"]}; {\"cart_items\": [{\"product_name\": \"bourgogne pinot noir\", \"quantity\": 2, \"category\": \"wine\", \"price\": 125, \"tax_rate\": 0.09, \"discount\": 0.7}, {\"product_name\": \"cava gran reserva\", \"quantity\": 1, \"category\": \"sparkling wine\", \"price\": 135, \"tax_rate\": 0.1, \"discount\": 0.8}, {\"product_name\": \"brisa prosecco\", \"quantity\": 2, \"category\": \"wine\", \"price\": 110, \"tax_rate\": 0.11, \"discount\": 0.7}, {\"product_name\": \"chateau zind-humbrecht\", \"quantity\": 2, \"category\": \"wine\", \"price\": 399, \"tax_rate\": 0.1, \"discount\": 0.9}]}; {\"products\": [{\"product_name\": \"mouton cadet\", \"category\": \"wine\"}], \"count\": 1}; {\"products\": [{\"product_name\": \"mouton cadet\", \"tax_rate\": 0.06}], \"count\": 1}; {\"products\": [{\"product_name\": \"riunite moscato\", \"category\": \"wine\"}], \"count\": 1}; {\"products\": [{\"product_name\": \"riunite moscato\", \"tax_rate\": 0.06}], \"count\": 1}; {\"products\": [{\"product_name\": \"l'oreal rose\", \"price\": 125}], \"count\": 1}; {\"products\": [{\"product_name\": \"napoleon champagne\", \"price\": 190}], \"count\": 1}; {\"products\": [{\"product_name\": \"kronenbourg 1664\", \"price\": 65}], \"count\": 1}; {\"products\": [{\"product_name\": \"chateau zind-humbrecht\", \"price\": 399}], \"count\": 1}; {\"products\": [{\"product_name\": \"chateau manderie\", \"price\": 185}], \"count\": 1}; {\"products\": [{\"product_name\": \"chateau de la grande\", \"price\": 158}], \"count\": 1}; {\"products\": [{\"product_name\": \"chateau cephal\", \"price\": 468}], \"count\": 1}; {\"products\": [{\"product_name\": \"kressmann margaux\", \"price\": 199}], \"count\": 1}; {\"products\": [{\"product_name\": \"kressmann saint-julien\", \"price\": 388}], \"count\": 1}; 
        {\"products\": [{\"product_name\": \"chateau caprieraus\", \"price\": 285}], \"count\": 1}; {\"products\": [{\"product_name\": \"chateau etoile\", \"price\": 179}], \"count\": 1}; {\"products\": [{\"product_name\": \"chateau granbeau\", \"price\": 155}], \"count\": 1}; {\"products\": [{\"product_name\": \"clos des papes\", \"price\": 315}], \"count\": 1}; {\"products\": [{\"product_name\": \"domaine de la vigne\", \"price\": 335}], \"count\": 1}; {\"products\": [{\"product_name\": \"chateau de la cote\", \"price\": 156}], \"count\": 1}; {\"products\": [{\"product_name\": \"chateau chambertin\", \"price\": 458}], \"count\": 1}; {\"products\": [{\"product_name\": \"chateau reserve\", \"price\": 249}], \"count\": 1}; {\"products\": [{\"product_name\": \"bourgogne pinot noir\", \"price\": 125}], \"count\": 1}; {\"products\": [{\"product_name\": \"burgundy beaujolais\", \"price\": 158}], \"count\": 1}; {\"products\": [{\"product_name\": \"chateau montagne\", \"price\": 118}], \"count\": 1}; {\"products\": [{\"product_name\": \"chateau oublie\", \"price\": 128}], \"count\": 1}; {\"products\": [{\"product_name\": \"chateau languedoc\", \"price\": 88}], \"count\": 1}; {\"products\": [{\"product_name\": \"chateau grand bateau\", \"price\": 245}], \"count\": 1}; {\"products\": [{\"product_name\": \"chateau lynch\", \"price\": 125}], \"count\": 1}; {\"products\": [{\"product_name\": \"chateau d'or\", \"price\": 185}], \"count\": 1}; {\"products\": [{\"product_name\": \"chateau rose\", \"price\": 125}], \"count\": 1}; {\"products\": [{\"product_name\": \"kronenbourg 1664\", \"category\": \"beer\"}], \"count\": 1}; {\"products\": [{\"product_name\": \"chateau languedoc\", \"category\": \"wine\"}], \"count\": 1}; {\"products\": [{\"product_name\": \"chateau montagne\", \"category\": \"wine\"}], \"count\": 1}; {\"products\": [{\"product_name\": \"l'oreal rose\", \"category\": \"wine\"}], \"count\": 1}; {\"products\": [{\"product_name\": \"bourgogne pinot noir\", \"category\": \"wine\"}], \"count\": 1}; {\"products\": [{\"product_name\": \"chateau lynch\", \"category\": \"wine\"}], \"count\": 1}; {\"products\": [{\"product_name\": \"chateau rose\", \"category\": \"wine\"}], \"count\": 1}; {\"products\": [{\"product_name\": \"chateau oublie\", \"category\": \"wine\"}], \"count\": 1}; {\"products\": [{\"product_name\": \"chateau languedoc\", \"discount\": 1.0}], \"count\": 1}; {\"products\": [{\"product_name\": \"chateau languedoc\", \"tax_rate\": 0.1}], \"count\": 1}; {\"status\": \"success\", \"message\": \"Added 1 of 'chateau languedoc' to user 'bill_donk_143' cart.\"}; {\"user_id\": \"bill_donk_143\", \"total_tax\": 112.82, \"details\": [{\"product_name\": \"bourgogne pinot noir\", \"quantity\": 2, \"tax_amount\": 14.45}, {\"product_name\": \"cava gran reserva\", \"quantity\": 1, \"tax_amount\": 9.82}, {\"product_name\": \"brisa prosecco\", \"quantity\": 2, \"tax_amount\": 15.26}, {\"product_name\": \"chateau zind-humbrecht\", \"quantity\": 2, \"tax_amount\": 65.29}, {\"product_name\": \"chateau languedoc\", \"quantity\": 1, \"tax_amount\": 8.0}], \"status\": \"success\", \"message\": \"Calculation completed successfully.\"}"
      },
    ]
  }
\end{lstlisting}

\subsubsection{Wrong Case: Hallucination}
The agent hallucinated either without calling any tools or despite having called tools. For example, it may fabricate a product attribute without verification or make up a user\_id for tool invocation without first asking the user for the required user\_id. In the example below, the agent made no tool call to look up the origin of Mouton Cadet, yet still told the user that Mouton Cadet was from France, which led to subsequent logical errors.
\label{app:wrong_case_2}
\begin{lstlisting}[style=promptstyle]
{
    "instruction": "You are a budget-conscious shopper (User ID: bill_donk_143) currently holding two bottles of wine. Ask the AI Service Agent to check the price and confirm the country of origin of the wine in your left hand. If the price is below 200 yuan, ask the AI Service Agent to further check the tax rate of the wine in your right hand. If the tax rate is below 0.10, have the AI Service Agent add two bottles of the right-hand wine to your cart. If the tax rate is 0.10 or higher, have the AI Service Agent add one bottle of the left-hand wine to your cart. Next, ask the AI Service Agent to verify if there are other wines from the same country as the left-hand wine that are priced over 200 yuan. If there are, add the cheapest one to your cart. If multiple wines tie for the lowest price, add one of each of those tied items. Finally, ask the AI Service Agent to calculate the total nutritional content of the items in your cart.",
    "image_description": "You are currently in front of a wine shelf. You first pick up a golden bottle and place it in your left hand. Then, wanting to choose another bottle, you reach for two black bottles with your right hand. However, you then notice a bottle next to you that seems more suitable, so you pick that bottle up with your right hand. Now you have two bottles of wine in your hand.",
    "dialogue": [
      {
        "role": "user",
        "turn": 0,
        "content": "My user_id is bill_donk_143. I'm standing in front of the wine shelf holding a golden bottle in my left hand and another bottle in my right. Could you please check the price and confirm the country of origin for the wine in my left hand?"
      },
      {
        "role": "agent",
        "turn": 0,
        "content": "The wine in your left hand is Mouton Cadet. Its price is $90, and its country of origin is France."
      },
      {
        "role": "user",
        "turn": 1,
        "content": "My user_id is bill_donk_143. Since the price of the wine in my left hand is below 200 yuan, please check the tax rate of the wine in my right hand. If the tax rate is below 0.10, add two bottles of the right-hand wine to my cart; if the tax rate is 0.10 or higher, add one bottle of the left-hand wine to my cart."
      },
      {
        "role": "agent",
        "turn": 1,
        "content": "The tax rate for the wine in your right hand (Riunite Moscato) is 0.06 (6%). Since this is below 0.10, I have successfully added two bottles of it to your cart. Is there anything else I can help you with?"
      }
    ],
    "tool_calls": [
      {
        "turn": 0,
        "calls": [
          {
            "tool_name": "get_price",
            "parameters": {
              "product_name": "Mouton Cadet"
            }
          }
        ],
        "results": [
          "{\"products\": [{\"product_name\": \"mouton cadet\", \"price\": 90}], \"count\": 1}"
        ]
      },
      {
        "turn": 1,
        "calls": [
          {
            "tool_name": "get_tax_rate",
            "parameters": {
              "product_name": "Riunite"
            }
          },
          {
            "tool_name": "get_category",
            "parameters": {
              "product_name": "Riunite Moscato"
            }
          },
          {
            "tool_name": "get_price",
            "parameters": {
              "product_name": "Riunite Moscato"
            }
          },
          {
            "tool_name": "get_discount",
            "parameters": {
              "product_name": "Riunite Moscato"
            }
          },
          {
            "tool_name": "add_to_cart",
            "parameters": {
              "user_id": "bill_donk_143",
              "product_name": "Riunite Moscato",
              "qty": 2,
              "category": "wine",
              "price": 45,
              "tax_rate": 0.06,
              "discount": 0.7
            }
          }
        ],
        "results": [
          "{\"products\": [{\"product_name\": \"riunite moscato\", \"tax_rate\": 0.06}], \"count\": 1}",
          "{\"products\": [{\"product_name\": \"riunite moscato\", \"category\": \"wine\"}], \"count\": 1}; {\"products\": [{\"product_name\": \"riunite moscato\", \"price\": 45}], \"count\": 1}; {\"products\": [{\"product_name\": \"riunite moscato\", \"discount\": 0.7}], \"count\": 1}",
          "{\"products\": [{\"product_name\": \"riunite moscato\", \"category\": \"wine\"}], \"count\": 1}; {\"products\": [{\"product_name\": \"riunite moscato\", \"price\": 45}], \"count\": 1}; {\"products\": [{\"product_name\": \"riunite moscato\", \"discount\": 0.7}], \"count\": 1}",
          "{\"products\": [{\"product_name\": \"riunite moscato\", \"category\": \"wine\"}], \"count\": 1}; {\"products\": [{\"product_name\": \"riunite moscato\", \"price\": 45}], \"count\": 1}; {\"products\": [{\"product_name\": \"riunite moscato\", \"discount\": 0.7}], \"count\": 1}",
          "{\"status\": \"success\", \"message\": \"Added 2 of 'Riunite Moscato' to user 'bill_donk_143' cart.\"}"
        ]
      }
  }
\end{lstlisting}

\subsubsection{Wrong Case: Syntax Error}
The agent makes a formatting error during tool invocation. For example, in the tool call to "get\_all\_recipe\_names" below, no arguments are provided. Under strict tool-calling conventions, a "parameters" field should always be included regardless of whether the tool takes arguments, although its content may be empty.
\begin{lstlisting}[style=promptstyle]
      {
        "turn": 0,
        "calls": [
          {
            "tool_name": "find_recipes_by_ingredient",
            "parameters": {
              "ingredient_name": "tomato"
            }
          },
          {
            "tool_name": "get_all_recipe_names"
          }
        ]
    }
\end{lstlisting}

\subsubsection{Wrong Case: Multimodal Perception Error}
The agent failed to correctly identify the red dish in the video as tomato.
\begin{lstlisting}[style=promptstyle]
{
    "instruction": "Your user ID is cook_002, and today is May 7, 2026. First, ask the AI Service Agent to identify the specific type of red ingredient with grill marks on the left surface of the grill pan. Next, have the AI Service Agent determine if this ingredient is currently stored in the refrigerated layer. If it is, ask the AI Service Agent to find the ingredient with the lowest total sugar score among all ingredients in the fruit category and add 200 grams of it to the shopping list. If there are multiple ingredients tied for the lowest sugar score among those that meet the conditions, you want 200 grams of each of them to be added. Otherwise, ask the AI Service Agent to find the ingredient with the highest sodium content among all ingredients in the condiment category and add 100 grams of it to the shopping list. If there are multiple ingredients tied for the highest sodium content among those that meet the conditions, you want 100 grams of each of them to be added. Then, have the AI Service Agent check the chunk-shaped ingredient in the boiling water inside the pot. If it is expired, ask the AI Service Agent to remove it from the home storage records and add the original quantity to the shopping list. If it is not expired, ask the AI Service Agent to add 500 grams of it to the shopping list. Finally, ask the AI Service Agent to calculate the total calorie count of all ingredients in the current shopping list.",
    "image_description": "You are cooking. In the pot in front of you, potatoes are boiling. On the cutting board to your right, there are three chopped ingredients. Their positions are: the upper left of the cutting board (a green ingredient), the lower left (a yellow ingredient), and the right side (a green ingredient). The first two steps have already been completed for the current recipe. On the baking tray, there are roasted tomatoes on the left side, and the user is sprinkling yellow powder onto the right half of the tray. The two ingredients on the tray are for another recipe, which is currently at step 2.",
    "dialogue": [
      {
        "role": "user",
        "turn": 0,
        "content": "My user_id is cook_002. Could you please identify the specific type of red ingredient with grill marks on the left surface of the grill pan for me?",
        "evaluation": {
          "scores": {
            "role_consistency": 1,
            "instruction_following": 1,
            "resilience": 1,
            "contextual_robustness": 1
          },
          "suggestion": "",
          "average_score": 1.0,
          "original_response": "My user_id is cook_002. Could you please identify the specific type of red ingredient with grill marks on the left surface of the grill pan for me?",
          "correction_applied": false
        }
      },
      {
        "role": "agent",
        "turn": 0,
        "content": "I can see the kitchen scene in your video, but I don't see a grill pan with a red ingredient that has grill marks on the left surface. The video shows:\n\n- A blue cutting board with asparagus and lemon slices\n- Cherry tomatoes on a baking tray\n- Various ingredients on the counter including bananas, leafy greens, and what appears to be fish fillets\n\nCould you clarify which specific item you're referring to? Are you perhaps looking at a different part of the video, or could you describe the location more specifically? I want to make sure I identify the correct ingredient for you."
      }]
}
\end{lstlisting}

\subsubsection{Wrong Case: Risky Operations}
The agent performed an additional "add\_recipe\_to\_menu" operation that modified the database. But in fact, this conditional branch would not actually be taken.
\begin{lstlisting}[style=promptstyle]
  {
    "Instruction": "Your user ID is cook_006, and today is May 11, 2026. First, ask the AI Service Agent to identify the cooked ingredient placed on the left side of the baking tray. Then, ask the AI Service Agent to determine if this ingredient's sodium content is higher than 50mg. If it is, ask the AI Service Agent to find the ingredient with the highest calcium content among all fruit-category ingredients and add it to the shopping list. If there are multiple ingredients tied for the highest calcium content, add each of them to the list. Otherwise, ask the AI Service Agent to find the ingredient with the lowest stock quantity among all ingredients stored on the countertop and add 200g of it to the shopping list. If there are multiple ingredients tied for the lowest stock quantity, add 200g of each of them to the list. Next, ask the AI Service Agent to check whether the green ingredient on the top left of the blue cutting board has expired. If it has not expired, add it as an extra side dish to the first recipe on the menu. Finally, ask the AI Service Agent to calculate the total protein content of all ingredients currently in the shopping list.",
    "dialogue": [
    {
        "role": "user",
        "turn": 5,
        "content": "Please check whether the green ingredient on the top left of the blue cutting board has expired. If it has not expired, you must add it as an extra side dish to the first recipe on the menu; if it has expired, do not add it. Please handle this conditional logic and branch determination within your capabilities or propose a valid workaround that achieves this exact outcome.",
      },
      {
        "role": "agent",
        "turn": 5,
        "content": "I have successfully added **\"shrimp & asparagus stir-fry\"** to your menu. This serves as a valid workaround to include asparagus as an extra side dish, since the system does not allow adding individual ingredients directly to existing recipes.\n\nTo summarize:\n- Asparagus has **not expired** (expiry: 2026-05-10)\n- First recipe on menu: **\"pork & red bell pepper skillet\"**\n- Added **\"shrimp & asparagus stir-fry\"** to your menu to incorporate asparagus\n\nYour updated menu now includes:\n1. pork & red bell pepper skillet  \n2. salmon & quinoa bowl  \n3. shrimp & asparagus stir-fry  \n\nLet me know if you'd like to adjust portion sizes, remove items, or proceed with shopping list updates!"
      }],
      "tool_calls": [
        {
        "turn": 5,
        "calls": [
          {
            "tool_name": "add_recipe_to_menu",
            "parameters": {
              "user_id": "cook_006",
              "recipe_name": "shrimp & asparagus stir-fry"
            }
          }
        ],
        "results": [
          {
            "role": "tool",
            "tool_name": "add_recipe_to_menu",
            "parameters": {
              "user_id": "cook_006",
              "recipe_name": "shrimp & asparagus stir-fry"
            },
            "content": "{\"status\": \"success\", \"message\": \"Recipe 'shrimp & asparagus stir-fry' added to menu.\", \"menu\": [\"pork & red bell pepper skillet\", \"salmon & quinoa bowl\", \"shrimp & asparagus stir-fry\"]}"
          }
        ]
      }]
      }
\end{lstlisting}

\newpage
\section*{NeurIPS Paper Checklist}

\begin{enumerate}

\item {\bf Claims}
    \item[] Question: Do the main claims made in the abstract and introduction accurately reflect the paper's contributions and scope?
    \item[] Answer: \answerYes{} 
    \item[] Justification: The abstract and introduction clearly states the paper's main claims, including the contributions of EgoBench as a benchmark jointly covering multimodal perception, dynamic interaction, and multi-hop tool use, as well as the proposed simulator and evaluation framework.

\item {\bf Limitations}
    \item[] Question: Does the paper discuss the limitations of the work performed by the authors?
    \item[] Answer: \answerYes{} 
    \item[] Justification: We discussed the limitations in Appendix \ref{app:limitations}.

\item {\bf Theory assumptions and proofs}
    \item[] Question: For each theoretical result, does the paper provide the full set of assumptions and a complete (and correct) proof?
    \item[] Answer: \answerNA{} 
    \item[] Justification: The paper does not include theoretical results.

    \item {\bf Experimental result reproducibility}
    \item[] Question: Does the paper fully disclose all the information needed to reproduce the main experimental results of the paper to the extent that it affects the main claims and/or conclusions of the paper (regardless of whether the code and data are provided or not)?
    \item[] Answer: \answerYes{} 
    \item[] Justification: See Section \ref{EgoBench} and Section \ref{experiments}.

\item {\bf Open access to data and code}
    \item[] Question: Does the paper provide open access to the data and code, with sufficient instructions to faithfully reproduce the main experimental results, as described in supplemental material?
    \item[] Answer: \answerYes{} 
    \item[] Justification: The code and data are publicly available at \url{https://github.com/Wind-Flipped/EgoBench}, with benchmark construction and evaluation details provided in Section \ref{EgoBench} and the appendix.

\item {\bf Experimental setting/details}
    \item[] Question: Does the paper specify all the training and test details (e.g., data splits, hyperparameters, how they were chosen, type of optimizer) necessary to understand the results?
    \item[] Answer: \answerYes{} 
    \item[] Justification: We provide the details in Section \ref{sec:experimental_setup} and Appendix.

\item {\bf Experiment statistical significance}
    \item[] Question: Does the paper report error bars suitably and correctly defined or other appropriate information about the statistical significance of the experiments?
    \item[] Answer: \answerYes{} 
    \item[] Justification: We provide the details in Section \ref{experiments}.

\item {\bf Experiments compute resources}
    \item[] Question: For each experiment, does the paper provide sufficient information on the computer resources (type of compute workers, memory, time of execution) needed to reproduce the experiments?
    \item[] Answer: \answerYes{} 
    \item[] Justification: We provide the token cost in Section \ref{efficiency_analysis}.
    
\item {\bf Code of ethics}
    \item[] Question: Does the research conducted in the paper conform, in every respect, with the NeurIPS Code of Ethics \url{https://neurips.cc/public/EthicsGuidelines}?
    \item[] Answer: \answerYes{} 
    \item[] Justification: The paper conforms to the NeurIPS Code of Ethics.

\item {\bf Broader impacts}
    \item[] Question: Does the paper discuss both potential positive societal impacts and negative societal impacts of the work performed?
    \item[] Answer: \answerYes{} 
    \item[] Justification: We provide the broader impacts in Appendix \ref{broader_impact}.
    
\item {\bf Safeguards}
    \item[] Question: Does the paper describe safeguards that have been put in place for responsible release of data or models that have a high risk for misuse (e.g., pre-trained language models, image generators, or scraped datasets)?
    \item[] Answer: \answerNA{} 
    \item[] Justification: The paper poses no such risks.

\item {\bf Licenses for existing assets}
    \item[] Question: Are the creators or original owners of assets (e.g., code, data, models), used in the paper, properly credited and are the license and terms of use explicitly mentioned and properly respected?
    \item[] Answer: \answerYes{} 
    \item[] Justification: All relevant existing assets are cited in the paper. See Appendix \ref{app:ethics} and References.

\item {\bf New assets}
    \item[] Question: Are new assets introduced in the paper well documented and is the documentation provided alongside the assets?
    \item[] Answer: \answerYes{} 
    \item[] Justification: The new assets introduced in the paper are well documented, and the documentation is provided alongside the assets (e.g., with the code or dataset). These details are also included in the link mentioned in Section 3.

\item {\bf Crowdsourcing and research with human subjects}
    \item[] Question: For crowdsourcing experiments and research with human subjects, does the paper include the full text of instructions given to participants and screenshots, if applicable, as well as details about compensation (if any)? 
    \item[] Answer: \answerNA{} 
    \item[] Justification: The paper does not involve crowdsourcing nor research with human subjects. 

\item {\bf Institutional review board (IRB) approvals or equivalent for research with human subjects}
    \item[] Question: Does the paper describe potential risks incurred by study participants, whether such risks were disclosed to the subjects, and whether Institutional Review Board (IRB) approvals (or an equivalent approval/review based on the requirements of your country or institution) were obtained?
    \item[] Answer: \answerNA{} 
    \item[] Justification: The paper does not involve crowdsourcing nor research with human subjects. 

\item {\bf Declaration of LLM usage}
    \item[] Question: Does the paper describe the usage of LLMs if it is an important, original, or non-standard component of the core methods in this research? Note that if the LLM is used only for writing, editing, or formatting purposes and does \emph{not} impact the core methodology, scientific rigor, or originality of the research, declaration is not required.
    \item[] Answer: \answerNA{} 
    \item[] Justification: Not related.

\end{enumerate}

\end{document}